%% file: main.tex
\documentclass[10pt,twocolumn,letterpaper]{article}

\usepackage{cvpr}              

\input{preamble}

\definecolor{cvprblue}{rgb}{0.21,0.49,0.74}
\usepackage[pagebackref,breaklinks,colorlinks,citecolor=cvprblue]{hyperref}
\usepackage[accsupp]{axessibility}

\usepackage{amsfonts}
\usepackage{amsmath}
\usepackage{pifont}
\usepackage{subcaption}
\usepackage{booktabs}
\usepackage{tabularx}
\newcolumntype{Y}{>{\centering\arraybackslash}X}

\def\paperID{2904} 
\def\confName{CVPR}
\def\confYear{2024}

\title{Single Domain Generalization for Crowd Counting}

\author{Zhuoxuan Peng,  S.-H. Gary Chan\\
The Hong Kong University of Science and Technology\\
{\tt\small \{zpengac, gchan\}@cse.ust.hk}
}

\begin{document}
\maketitle
\input{sec/0_abs}    
\input{sec/1_intro}
\input{sec/2_related}
\input{sec/3_method}
\input{sec/4_exp}
\input{sec/5_conclu}
{
    \small
    \bibliographystyle{ieeenat_fullname}
    \bibliography{main}
}

\input{sec/X_suppl}

\end{document}

%% file: preamble.tex
\usepackage[dvipsnames]{xcolor}


%% file: sec/0_abs.tex
\begin{abstract}
Due to its promising results, density map regression has been widely employed for image-based crowd counting.  The approach, however,
 often suffers from severe performance degradation when tested on data from unseen scenarios, the so-called ``domain shift'' problem.  To address the problem, we investigate in this work single domain generalization (SDG) for crowd counting.  The existing SDG approaches are mainly for image classification and segmentation, and can hardly be extended to 
our case due to its regression nature and label ambiguity (i.e.,~ambiguous pixel-level ground truths). We propose MPCount, a novel effective SDG approach even for narrow source distribution. 
%
%
MPCount 
stores diverse density values for density map regression and reconstructs domain-invariant features by means of only one memory bank, a content error mask and attention consistency loss.
By partitioning the image into grids,
it 
employs patch-wise classification as an auxiliary task to 
mitigate label ambiguity.
%
%
Through extensive experiments on different datasets,  MPCount is shown to significantly improve counting accuracy compared to the state of the art under diverse scenarios unobserved in the training data characterized by narrow source distribution. Code is available at~\url{https://github.com/Shimmer93/MPCount}.
\end{abstract}

%% file: sec/1_intro.tex
\section{Introduction}
\label{sec:intro}

\begin{figure}[t]
    \centering
    \begin{subfigure}{.64\linewidth}
        \centering
        \includegraphics[width=\linewidth]{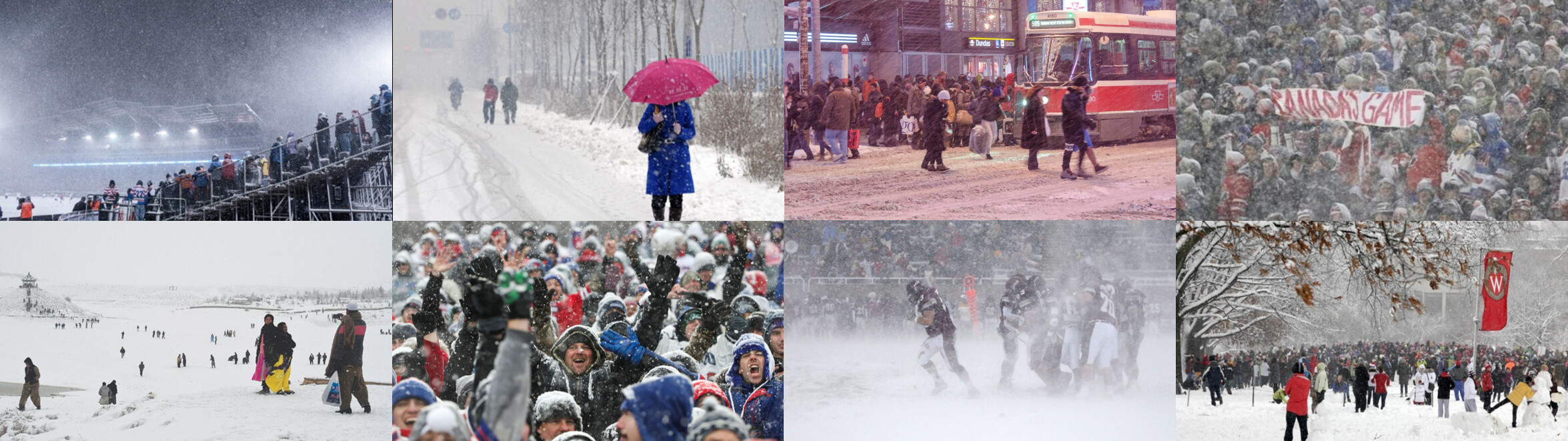}
        \caption{Source domain}
        \label{sfig:src}
    \end{subfigure}
    \begin{subfigure}{.32\linewidth}
        \centering
        \includegraphics[width=\linewidth]{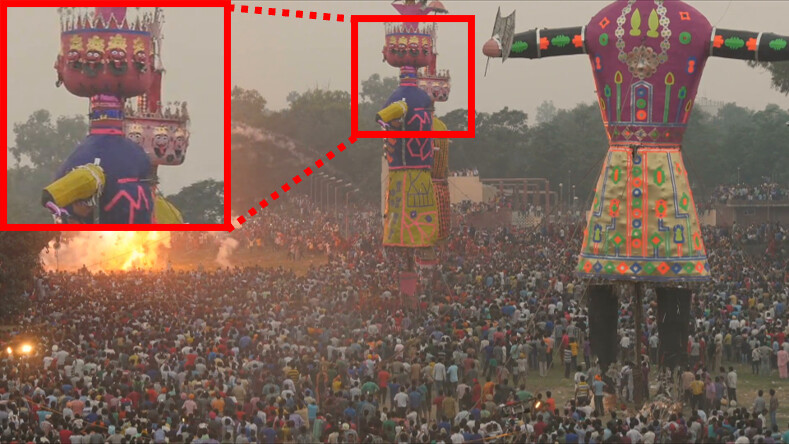}
        \caption{Target image}
        \label{sfig:tgt}
    \end{subfigure}
    \begin{subfigure}{.32\linewidth}
        \centering
        \includegraphics[width=\linewidth]{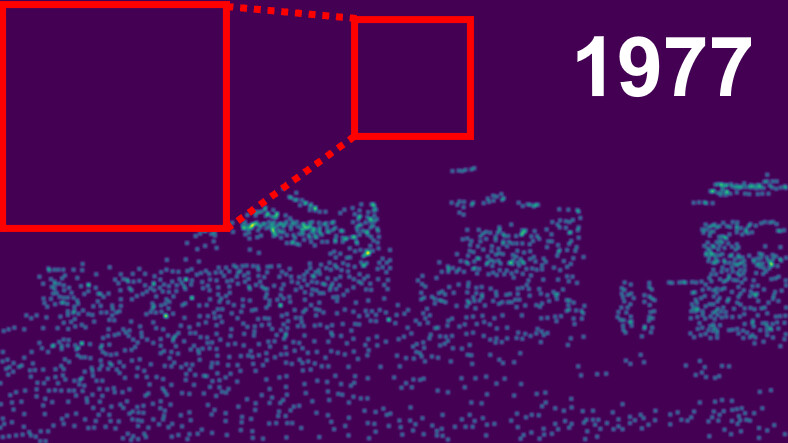}
        \caption{Ground-truth}
        \label{sfig:show_gt}
    \end{subfigure}
    \begin{subfigure}{.32\linewidth}
        \centering
        \includegraphics[width=\linewidth]{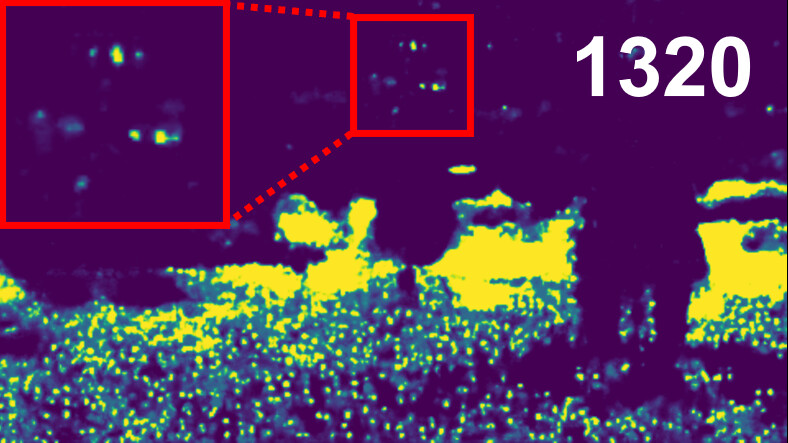}
        \caption{DCCUS}
        \label{sfig:show_dccus}
    \end{subfigure}
    \begin{subfigure}{.32\linewidth}
        \centering
        \includegraphics[width=\linewidth]{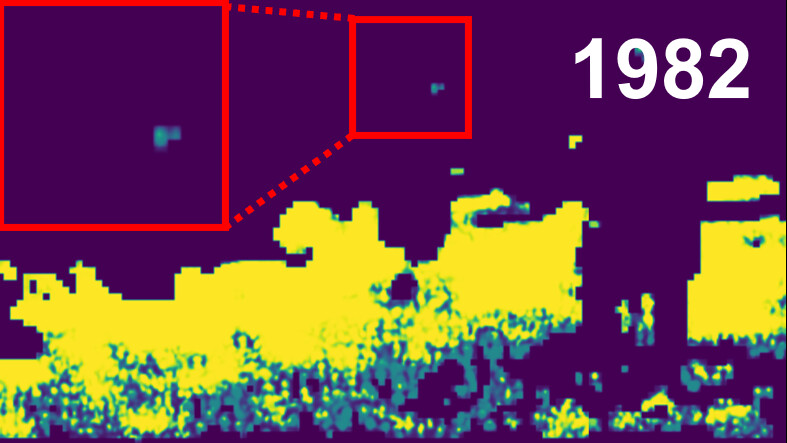}
        \caption{MPCount (Ours)}
        \label{sfig:show_ours}
    \end{subfigure}
    \caption{Single domain generalization for crowd counting. (a) Sample images from a single source domain $\mathcal{S}$ with narrow distribution. (b) A challenging target image from an unseen scenario. (c) The ground-truth density map. (d) The predicted density map from the previous work DCCUS~\cite{du2023domaingeneral} trained on $\mathcal{S}$. (e) The predicted density map from our MPCount trained on $\mathcal{S}$. MPCount achieves much lower counting error than DCCUS and makes better predictions in the cropped area against the domain shift.}
    \label{fig:show}
\end{figure}

Image-based crowd counting is to estimate the number of people (or objects) in an image. The recent mainstream counting approach is density map regression, where a density map with continuous pixel value is predicted from the input image, and the total crowd count is obtained by summing all these pixel values. During training, the ground-truth density map is generated by smoothing point annotations given by the locations of human heads. Density map regression has been demonstrated to achieve encouraging results in crowd counting.

Existing training models predominantly assume that the testing data share the same source distribution, i.e.,~fully supervised learning. Such assumption, however, often breaks in practice due to changes in camera position, weather condition, etc. Such deviation from the training data (i.e.~source domain) to unseen testing data (i.e.~target domain) is known as {\em domain shift}, which severely degrades the performance of deep learning models in real-world deployment.

To mitigate the domain shift problem, domain adaptation (DA) approaches have been proposed to transfer knowledge from the source domain to the target one, typically by fine-tuning pre-trained models with 
target domain data~\cite{7298684, wang2019learning, Cai_Chen_Guan_Lin_Lu_Wang_He_2023, TAS24}. Despite
promising results, the fine-tuning process of DA is usually laborious.  Furthermore, target domain data may not be easily available in reality. 

In contrast to DA, domain generalization (DG) aims to generalize models to any unobserved target domain without the need for target data. As data and model fine-tuning are not necessary at the target domain, DG is more convenient to be deployed in unseen or unpredictable domain.  We consider in this work the challenging case of single domain generalization (SDG) for image-based crowd counting, where only one source domain, likely of narrow distribution~(\cref{sfig:src}), is used for training. 

A common general SDG technique is to construct domain-invariant features
from different augmented versions of an image.  Such approach has been shown promising for various tasks~\cite{Choi2021RobustNetID, Vidit_2023_CVPR, Kim_2023_CVPR}.  Independent of data augmentation, 
many other SDG approaches have been proposed for classification and segmentation to further boost performance.  Unfortunately, most of them cannot be extended to crowd counting, a regression task without category information. Moreover, while labels in classification and segmentation tasks are precise and domain-invariant, the labels for density-based crowd counting are usually ambiguous. Such ``label ambiguity'' stems from the fact that the density values are generated from point annotations, and hence human heads and backgrounds may be assigned to similar value under different scenarios.

The recent work of~\cite{du2023domaingeneral} achieves DG  for crowd counting
based on dividing the source into multiple sub-domains.  While encouraging, it requires a rather broad source domain distribution.
%
We propose, for the first time, an SDG approach for crowd counting called MPCount without sub-domain division, thereof effective and extensible to narrow source distribution. We highlight its performance as compared with a state-of-the-art scheme in Fig.~\ref{fig:show}. MPCount 
achieves its superior performance with the following two novel components: 
\begin{itemize}
    \item \textit{An attention memory bank (A\textbf{M}B) to tackle density regression:}
    MPCount employs only one attention memory bank (AMB) that takes a pair of features encoded from an crowd image and its data-augmented version as input. To cover continuous density values for regression with a finite memory size, AMB reconstructs each feature vector as an attention over memory vectors. These memory vectors 
    learn domain-invariant representations from the pair of features of different styles but similar content (i.e.~crowd densities). 
    We present the content error mask (CEM) to eliminate domain-related content from input features, as characterized by a large discrepancy between instance-normalized feature element pairs. Furthermore, the newly employed attention consistency loss (ACL) enforces the similarity of attention scores produced from input features, ensuring the consistency of memory vectors.
    
    \item \textit{Patch-wise classification (\textbf{P}C) to tackle label ambiguity}: To address the label ambiguity, MPCount employs a novel auxiliary task called patch-wise classification (PC). In this task, each crowd image is evenly divided into fixed-size square patches, say $16\times16$, classified into two classes, i.e., containing human heads or not. The density values in the areas classified as without head are filtered out during density regression, so that the pixel-level ambiguity is overcome by coarser but more accurate patch-wise binary labels. 
\end{itemize}

To validate MPCount, we conduct extensive experiments on various crowd counting datasets including ShanghaiTech A (SHA) \& B (SHB) and JHU-Crowd++. 
We introduce a new challenging setting of narrow source distribution, where only images of the same category, such as ``snow'' (SN) and ``fog/haze'' (FH) in JHU-Crowd++, are used in training. 
We demonstrate that MPCount achieves excellent performance not only on traditional inter-dataset benchmarks, but also under our newly introduced narrow source setting.
Let  $\mathcal{S} \rightarrow \mathcal{T}$  denote the case where the source domain is $\mathcal{S}$ and the target domain is $\mathcal{T}$. As compared to the state of the art, MPCount is shown to reduce significantly counting error  by 21.8\% on SN $\rightarrow$ FH, 18.6\% on FH $\rightarrow$ SN, 18.2\% on SHB $\rightarrow$ SHA and 
9.5\% on SHA $\rightarrow$ SHB.

%% file: sec/2_related.tex
\section{Related Works}
\label{sec:related}

\subsection{Fully Supervised Crowd Counting}\label{sec:fscc}

The mainstream approach to crowd counting is via density map regression, where each pixel of an image is assigned a count value, and the summation of these values yields an estimate of the total number of people. Most methods focus on improving counting accuracy under the fully supervised setting by employing novel network designs~\cite{7780439, li2018csrnet, sasnet, lin2022boosting, bai2022transformer} or loss functions~\cite{9009503, wang2020DMCount, Wan_2021_CVPR, shu2022crowd}. While encouraging results have been shown, their performance usually drop significantly when evaluated on out-of-distribution data, due to their limited generalization ability. Besides, several works~\cite{Wang2020STNetST, 9710789, 9347700} have proposed auxiliary tasks in crowd counting for certain purposes. However, these tasks still require pixel-level predictions and contribute little to tackling label ambiguity. 

\subsection{Domain Adaptation (DA) for Crowd Counting}

Domain adaptation (DA) has been widely studied in crowd counting to address domain shift by adapting source domain information to a particular target domain. While a few methods study supervised DA where labeled data from target domain are available~\cite{7298684, Wang_2022}, most approaches tackle unsupervised DA, utilizing only unlabeled target domain data~\cite{wang2019learning, Gao2019DomainAdaptiveCC, 10.1145/3394171.3413825, 10.1145/3474085.3475230, 10.1145/3503161.3548298, Cai_Chen_Guan_Lin_Lu_Wang_He_2023}. Despite the notable advancements, DA methods normally require data from the target domain, which may not be easily available in practice. 

\subsection{Single Domain Generalization (SDG)}\label{sec:sdg}

Domain Generalization (DG) approaches are designed to train a network with generalization ability solely utilizing source domain data, and single domain generalization (SDG) is the special case when only one source domain is available. There are various techniques for SDG, including 
1) adversarial data generation~\cite{qiaoCVPR20learning, Li2021ProgressiveDE, Zhang2022SingleSourceDE, su2023slaug},
2) feature normalization/whitening transformation
~\cite{Pan2018TwoAO, 9513542, pan2018switchable, Choi2021RobustNetID, Xu2022DIRLDR} and 3) domain-general network design, 
such as a convolution layer~\cite{Wan2022MetaCN}, vision transformer~\cite{Sultana_2022_ACCV} and memory bank~\cite{Chen2021ASA}.
Among various SDG methods, a common general technique is to simulate domain shift with data augmentation and extract domain-invariant information from different augmented versions of features~\cite{Choi2021RobustNetID, Xu2022DIRLDR, Kim_2023_CVPR}. While SDG has attracted widespread attention in classification and segmentation, it is still a nascent topic in crowd counting due to its regression nature and label ambiguity. 

DCCUS~\cite{du2023domaingeneral} studies SDG for crowd counting. In this method, the source domain is dynamically clustered into sub-domains simulated as meta-training and -test sets in a meta-learning strategy. Two types of memory modules and several losses are proposed to distinguish and record domain-invariant and -related information.
DCCUS designs the memory mechanism specifically for regression, but the problem of ambiguous labels is not considered. Moreover, when the source domain follows a narrow distribution, the sub-domain division process may experience reduced effectiveness and affect the generalization ability of the counting model. MPCount, without sub-domain division, employs novel patch-wise classification to address the label ambiguity problem. Additionally, a single memory bank for regression is introduced without the need for sub-domain partitioning, enabling MPCount to generalize well even in source domains of narrow distributions. 

%% file: sec/3_method.tex
\begin{figure*}[t]
    \centering
    \includegraphics[width=\textwidth]{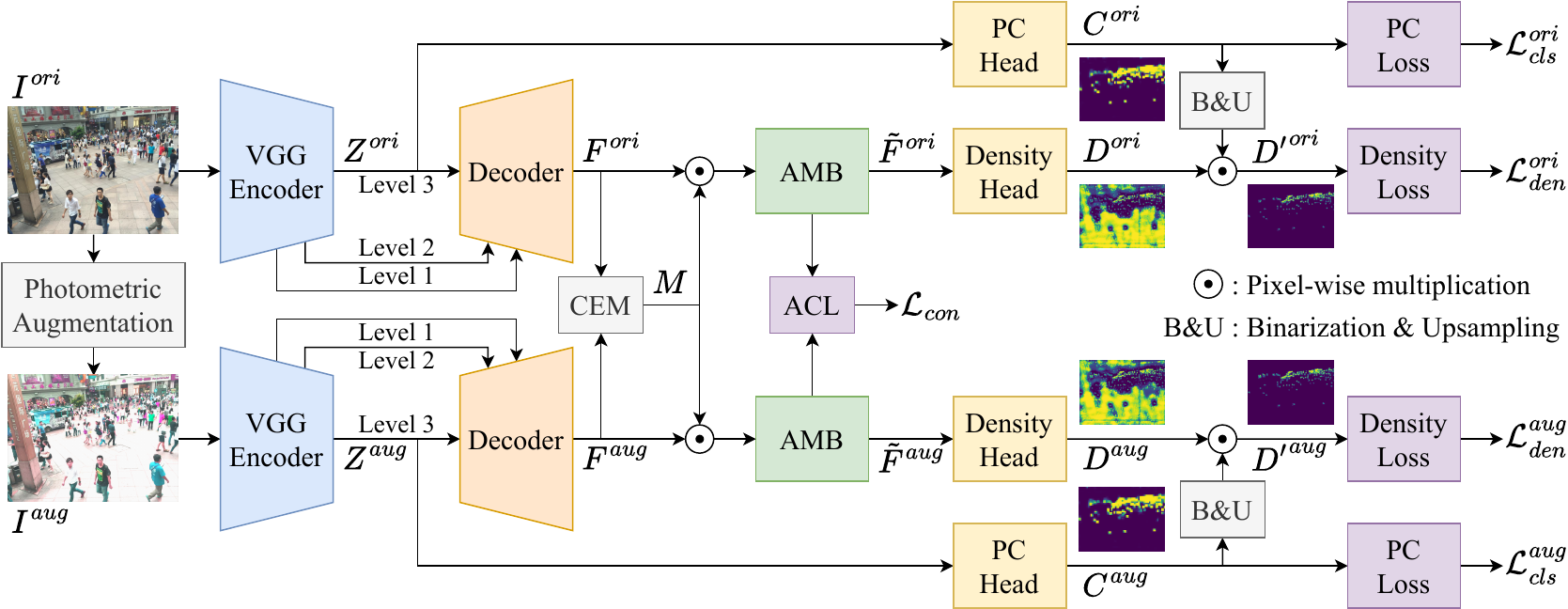}
    \caption{The overall training pipeline of our proposed MPCount. All identical modules in this diagram share the same weights. In the encoder-decoder structure, a higher level indicates a deeper feature. }
    \label{fig:mpcount}
\end{figure*}

\section{MPCount}

This section provides a detailed description of our MPCount scheme. We begin by reviewing the foundational concepts of crowd counting in ~\cref{sec:pre}. Next, in ~\cref{sec:over}, we present a comprehensive overview of the entire scheme. We then elaborate the attention memory bank (\cref{sec:dimb}) designed for density regression, followed by the content error mask (\cref{sec:cem}) and attention consistency loss (\cref{sec:acl}). After that, we discuss the patch-wise classification (\cref{sec:pcm}) proposed to address the challenge of ambiguous labels. Finally, we detail the overall training loss in ~\cref{sec:loss}.

\subsection{Crowd Counting Preliminaries} \label{sec:pre}
First, we review the preliminary knowledge of deep learning-based crowd counting via density map regression. Our objective is to train a neural network $\mathcal{N}$ with parameter $\theta$ that takes in an input image $\mathcal{I}\in\mathbb{R}^{H\times W\times 3}$ and outputs a density map $\hat{\mathcal{D}}=\mathcal{N}(\mathcal{I};\theta)$ of size $H\times W$. The estimated count $\hat{c}$ is the sum of all pixel-wise density values in $\hat{\mathcal{D}}$,  i.e.,

\begin{align}
    \hat{c} = \sum_{i=1}^H\sum_{j=1}^W \hat{\mathcal{D}}_{ij}.
\end{align}

Generation of ground-truth density maps usually follows the method in~\cite{Sindagi2017CNNBasedCM}. Given the human head point annotations $\mathcal{H}$ where each $h\in\mathcal{H}$ is the coordinate of a head location, the ground-truth density map $\mathcal{D}\in\mathbb{R}^{H\times W}$ is calculated as the sum of the 2D Gaussian filter applied to each head coordinate $h$, i.e.,

\begin{align}
    \mathcal{D} = \sum_{h\in \mathcal{H}}\delta(x-h)\times G_{\sigma}(x),
\end{align} where $\delta(\cdot)$ is the discrete delta function, and $G_{\sigma}(\cdot)$ is the Gaussian kernel with a fixed variance $\sigma$. 

A common objective function to supervise the network $\mathcal{N}$ can be written as

\begin{align}\label{eq:0}
    \mathbb{L}(\theta) = ||\mathcal{D}, \mathcal{\hat{D}}||_{2}^{2}.
\end{align}

\subsection{Scheme Overview}\label{sec:over}

The overall structure of our method is illustrated in ~\cref{fig:mpcount}. During training, photometric transformations are applied to the original image $\mathcal{I}^{ori}$ to produce an augmented version $\mathcal{I}^{aug}$. The feature extractor encodes them into features $\mathcal{F}^{ori}$ and $\mathcal{F}^{aug}$, and a content error mask (CEM) $\mathcal{M}$ is applied to them. The masked features are reconstructed into $\tilde{\mathcal{F}}^{ori}$ and $\tilde{\mathcal{F}}^{aug}$ via an attention memory bank (AMB) and passed to the density head for crowd density prediction. Meanwhile, the highest-level features encoded by the feature extractor, $\mathcal{Z}^{ori}$ and $\mathcal{Z}^{aug}$, are fed into the patch-wise classification (PC) head, and the predicted PC maps (PCMs) $\hat{\mathcal{C}}^{ori}$ and $\hat{\mathcal{C}}^{aug}$ are binarized and resized to serve as a mask to filter out areas without human heads in the estimated density maps $\hat{\mathcal{D}}^{ori}$ and $\hat{\mathcal{D}}^{aug}$. The final predictions $\mathcal{D}'^{ori}$ and $\mathcal{D}'^{aug}$ are two density maps with certain areas masked out by the PCMs. The entire training process is supervised under a combination of density losses, PC losses and the attention consistency loss (ACL). The inference process solely employs $\mathcal{I}^{ori}$ as input, and outputs a single density map $\mathcal{D}'^{ori}$.

\subsection{Attention Memory Bank (AMB)} \label{sec:dimb}
The attention memory bank (AMB) is designed to automatically learn domain-invariant features for density regression. 
~\cite{kim2022pin} proposes a memory bank where memory vectors are updated individually, each corresponding to a certain category. However, this design cannot be applied to regression tasks, since a finite number of memory vectors cannot cover continuous density values. 
Inspired by~\cite{du2023domaingeneral}, we reconstruct each feature vector as an attention over memory vectors, so that any density value may be represented as a linear combination of representative values in the memory bank. 

The AMB $\mathcal{V}\in \mathbb{R}^{M\times C}$ consists of $M$ memory vectors of dimension $C$. Given a flattened input feature map $\mathcal{F}\in\mathbb{R}^{HW\times C}$, we first compute the attention scores $\mathcal{A}$ between the feature $\mathcal{F}$ (query) and the memory $\mathcal{V}$ (key). Then the reconstructed feature map $\tilde{\mathcal{F}}$ is calculated as a linear combination of $\mathcal{V}$ (value) with $\mathcal{A}$ as weights. The reconstruction process can be summarized as follows: 

\begin{align}
\mathcal{A}(\mathcal{F}, \mathcal{V}) &= \text{Softmax}\left(\frac{\mathcal{F}\mathcal{V}^{T}}{\sqrt{C}}\right),\label{eq:3}
\end{align} and
\begin{align}
\tilde{\mathcal{F}} &= \mathcal{A}\mathcal{V}.\label{eq:4}
\end{align}


While~\cite{du2023domaingeneral} relies on multiple memory banks corresponding to different sub-domains to distinguish domain-invariant and -related information, we only employ a single AMB and propose the novel CEM~(\cref{sec:cem}) and ACL~(\cref{sec:acl}) to ensure that it stores domain-invariant representations. To this end, our feature reconstruction mechanism can work without sub-domain division, thereof is effective even with a narrow source distribution.

\subsection{Content Error Mask (CEM)} \label{sec:cem}

CEM is proposed to guarantee the resemblance of content contained in the pair of input features by excluding possible domain-related content information. A common technique to disentangle style and content in a feature is instance normalization (IN)~\cite{Ulyanov2016InstanceNT}, where the feature statistics (channel-wise mean and variance) are considered to contain style information, while the instance normalized features retain content information. Based on this idea, we assume that the discrepancy of instance normalized features reflect effects of domain shift on feature content and may mislead AMB to learn domain-related information. Therefore, we filter out elements in the input features with difference between their instance normalized versions above a certain threshold, resulting in a pair of features with similar content information for domain-invariant feature reconstruction.

More specifically, given the features extracted from the original and augmented images, $\mathcal{F}^{ori}$ and $\mathcal{F}^{aug}\in \mathbb{R}^{H\times W\times C}$, we define the content error mask $\mathcal{M}\in \mathbb{R}^{H\times W\times C}$ as follows:

\begin{align} \label{eq:1}
    \mathcal{M}_{ijk}=\left\{\begin{aligned}
&1\quad \text{if }|\text{IN}(\mathcal{F}^{ori})_{ijk}-\text{IN}(\mathcal{F}^{aug})_{ijk}| \leq \alpha\\
&0\quad \text{otherwise}
\end{aligned}\right.
\end{align} where $\alpha$ is the threshold indicating whether the error value reflects possible inconsistency of content information caused by domain shift. 

Given $\mathcal{F}$ to be one of $\mathcal{F}^{ori}$ and $\mathcal{F}^{aug}$, we filter out the domain-related content information by

\begin{align}
\mathcal{F}' = \text{Dropout2D}(\mathcal{F} \odot \mathcal{M}).
\end{align} The random 2D dropout is applied to $\mathcal{F}'$ to prevent the memory from relying solely on certain channels.

\subsection{Attention Consistency Loss (ACL)} \label{sec:acl}

The style information of $\mathcal{F}^{ori}$ and $\mathcal{F}^{aug}$ preserved in ~\cref{sec:cem} may produce a large discrepancy on their attention distributions over the memory vectors in AMB even if they contain identical content information. To ensure that each memory vector consistently store specific domain-invariant representations, we propose the attention consistency loss (ACL) to enforce the similarity of attention scores produced from $\mathcal{F}^{ori}$ and $\mathcal{F}^{aug}$. 

We consider $\mathcal{A}$ in Eq.~\ref{eq:3} as a distribution, and the ACL $\mathbb{L}_{con}$ is calculated as the distance between the distribution $\mathcal{A}^{ori}$ produced by $\mathcal{F}^{ori}$ and $\mathcal{A}^{aug}$ produced by $\mathcal{F}^{aug}$. 
Here the simple Euclidean distance is selected as the distance measure due to its computational stability during training: 

\begin{align}
\mathbb{L}_{con} = ||\mathcal{A}^{aug},\mathcal{A}^{ori}||_2^2.
\end{align}

\begin{figure}[t]
    \centering
    \begin{subfigure}{.3\linewidth}
        \centering
        \includegraphics[width=\linewidth]{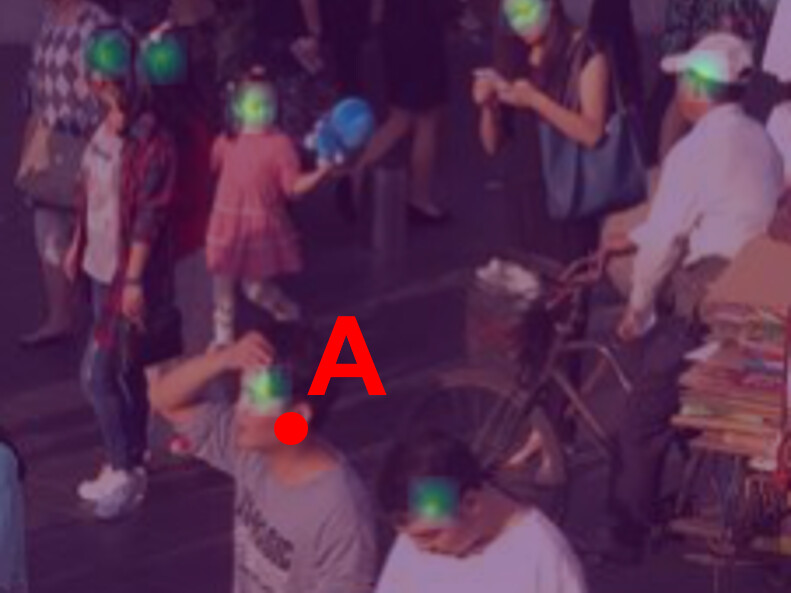}
        \caption{Density Map}
        \label{sfig:cmp_den}
    \end{subfigure}
    \begin{subfigure}{.3\linewidth}
        \centering
        \includegraphics[width=\linewidth]{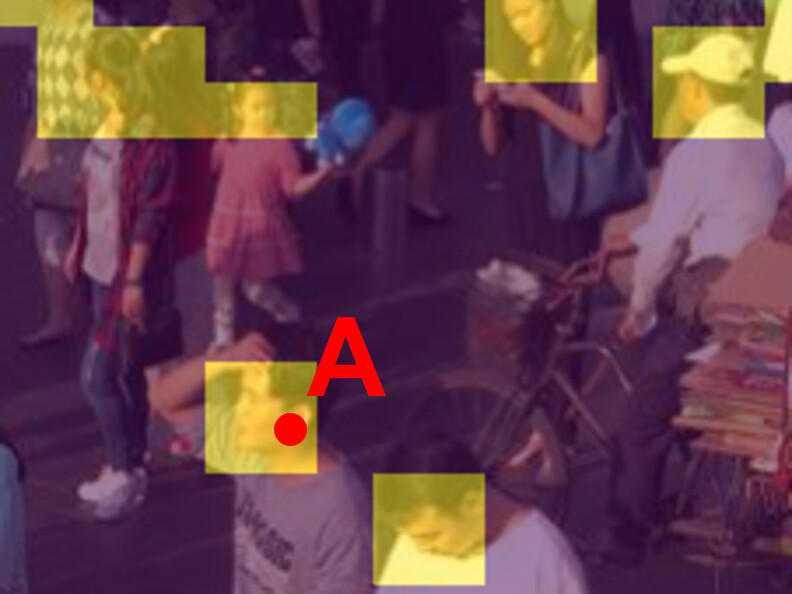}
        \caption{PCM}
        \label{sfig:cmp_pcm}
    \end{subfigure}
    \caption{Illustration of label ambiguity and how PCM tackles it. Point A is a typical example which belongs to a human head but is assigned the density value 0 as the background in the density map due to varying head sizes. In the PCM, the patch covering A is correctly classified as containing heads, thanks to the patch-level accuracy of the labels. }
    \label{fig:cmp}
\end{figure}

\subsection{Patch-wise Classification (PC)} \label{sec:pcm}

As discussed in ~\cref{sec:pre}, ground-truth density maps are usually calculated only based on point annotations. In different scenarios, pixels of a human head and the background might be assigned the same density value due to various head sizes (\cref{sfig:cmp_den}), violating the common assumption that labels are accurate and domain-invariant, thereof resulting in the ambiguous labels in crowd counting. 

To tackle the label ambiguity, we propose a novel auxiliary task named patch-wise classification (PC) and its corresponding supervision signal, patch-wise classification map (PCM), as illustrated in ~\cref{sfig:cmp_pcm}. A PCM divides an image evenly into $P\times P$ ($P=16$ empirically) patches and classifies each patch into two categories: containing human heads or not. Such patch-level predictions mitigate the uncertainty in pixel-level density maps by providing coarser but more accurate patch-level information.

In practice, a ground truth PCM $\mathcal{C}$ with patch size $P$ can be calculated as:
\begin{align}
\mathcal{C}_{ij} = 1 - \delta\left(\sum_{k=iP}^{(i+1)P}\sum_{l=jP}^{(j+1)P}\mathcal{D}_{kl}\right),
\end{align} where $\mathcal{D}$ is the ground truth density map and $\delta(\cdot)$ is the discrete delta function. 


Our model predicts a PCM $\mathcal{\hat{C}}$ supervised by the binary cross entropy (BCE) loss:
\begin{align}
\mathbb{L}_{cls} = \text{BCE}(\mathcal{C}, \mathcal{\hat{C}}).
\end{align}

Subsequently, $\mathcal{\hat{C}}$ is binarized and resized to obtain $\mathcal{C}'$ which matches the dimensions of the predicted density map $\mathcal{\hat{D}}$. $\mathcal{C}'$ is then used to mask the regions classified without crowds in $\mathcal{\hat{D}}$, resulting in the final density map $\mathcal{D}'$:
\begin{align}
\mathcal{D}'= \mathcal{\hat{D}} \odot \mathcal{C}'.
\end{align}

Finally, the density regression head is supervised with the ground truth density map $\mathcal{D}$, and the common Euclidean distance in~\cref{eq:0} is used as the objective function:
\begin{align}
    \mathbb{L}_{den} = ||\mathcal{D}, \mathcal{D}'||_{2}^{2}.
\end{align}

\subsection{Model Training}\label{sec:loss}

The overall training loss is a combination of density losses $\mathbb{L}_{den}^{ori}, \mathbb{L}_{den}^{aug}$, PC losses $\mathbb{L}_{cls}^{ori}, \mathbb{L}_{cls}^{aug}$ and the attention consistency loss $\mathbb{L}_{con}$, which can be written as
\begin{align}
\mathbb{L} = \mathbb{L}_{den}^{ori} + \mathbb{L}_{den}^{aug} + \lambda_{cls}\left(\mathbb{L}_{cls}^{ori} + \mathbb{L}_{cls}^{aug}\right) + \lambda_{con}\mathbb{L}_{con},
\end{align} 
where $\lambda_{cls}$ and $\lambda_{con}$ are weighting parameters to balance different loss terms.

%% file: sec/4_exp.tex
\section{Illustrative Experimental Results}

In this section, we first introduce the datasets used in experiments in ~\cref{sec:data}. Next, the implementation details are specified in ~\cref{sec:imp}, then evaluation metrics are described in ~\cref{sec:met}. We present the results compared with state of the art in ~\cref{sec:comp}. Finally, we conduct ablation studies and other additional analysis in ~\cref{sec:abl}.

\subsection{Datasets}\label{sec:data}

We evaluate our method on four mainstream crowd counting datasets: ShanghaiTech Part A \& B, UCF-QNRF and JHU-Crowd++. 

\begin{figure}[t]
    \centering
    \begin{subfigure}{.24\linewidth}
        \centering
        \includegraphics[width=\linewidth]{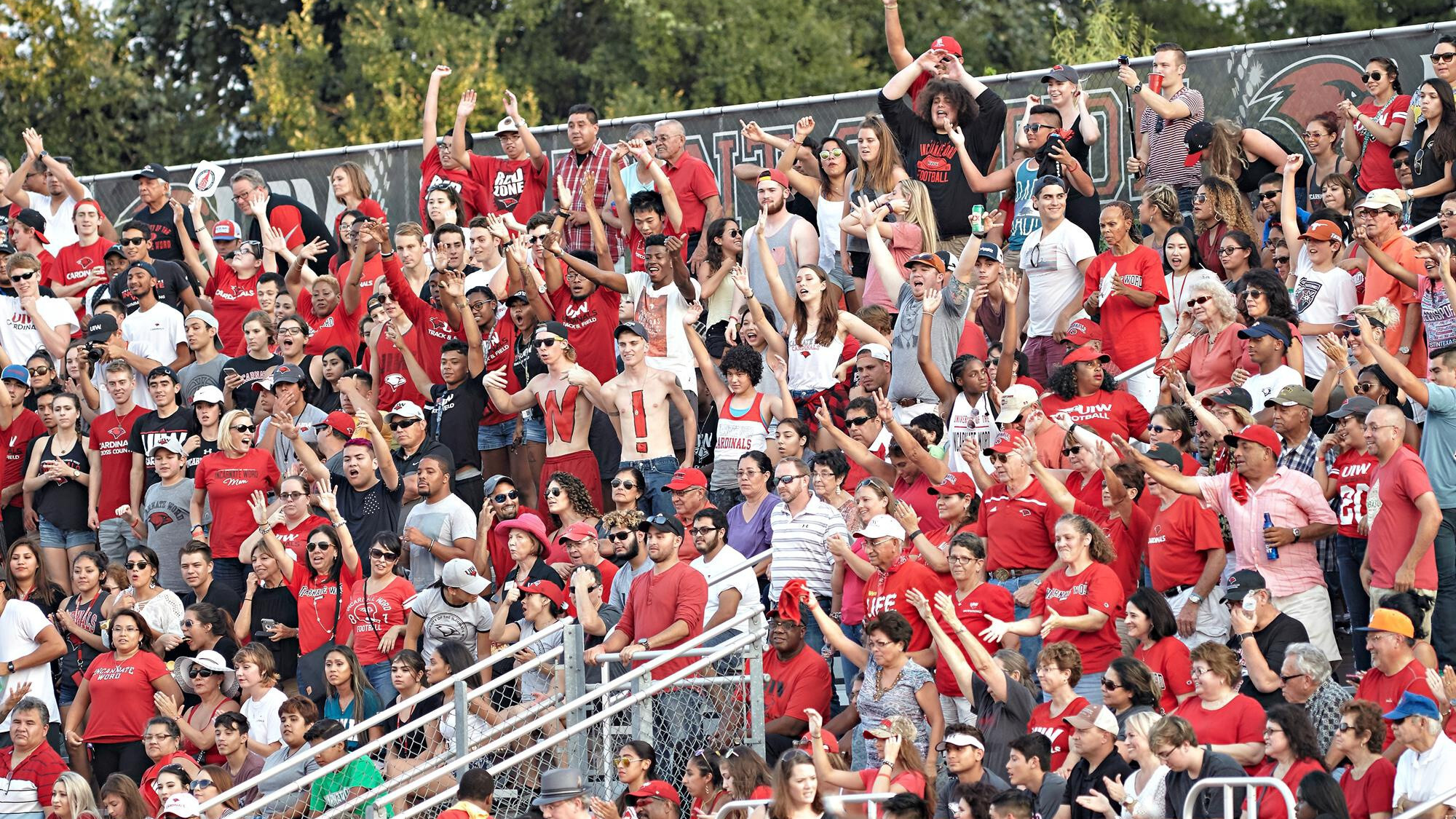}
    \end{subfigure}
    \begin{subfigure}{.24\linewidth}
        \centering
        \includegraphics[width=\linewidth]{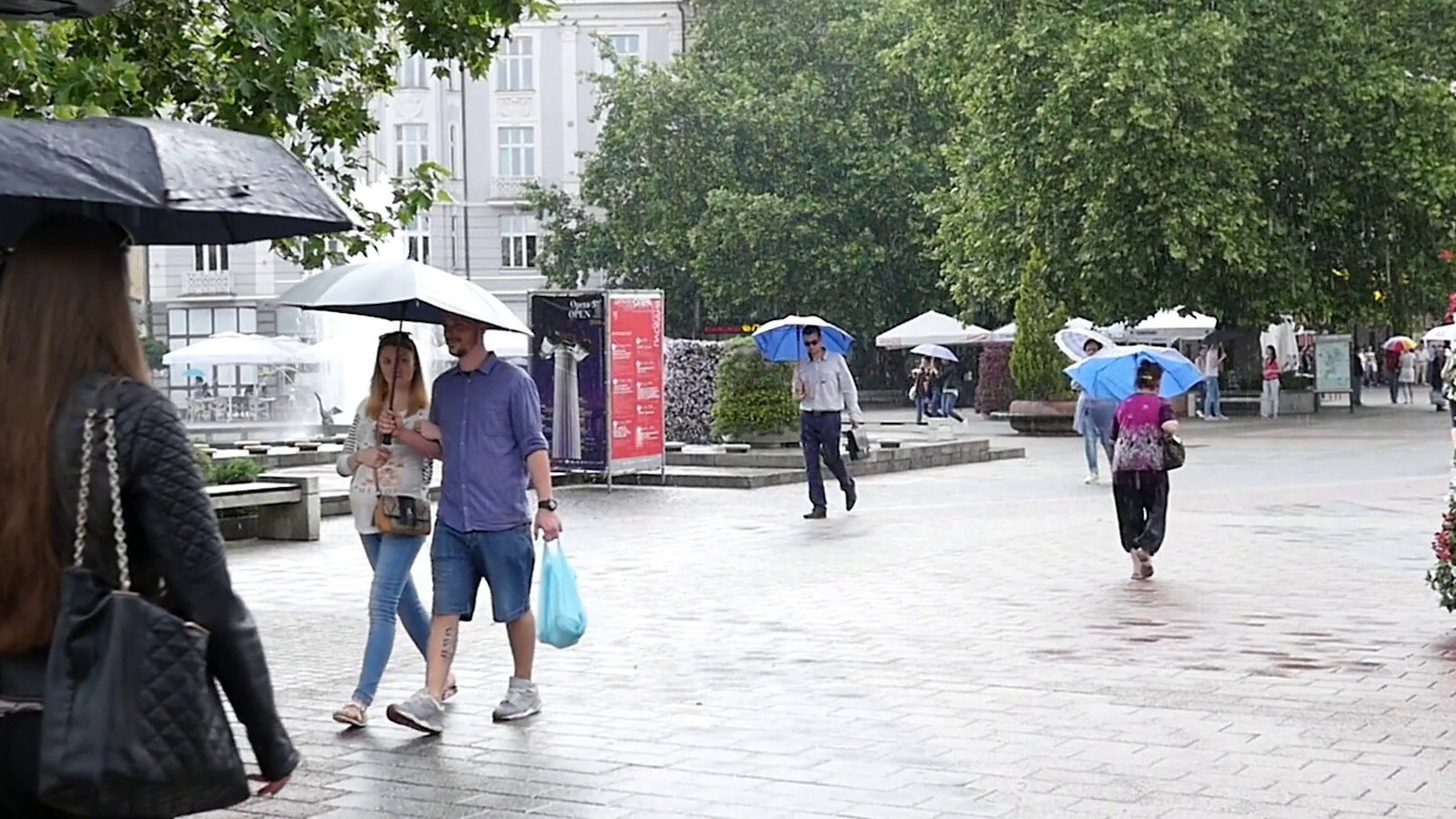}
    \end{subfigure}
    \begin{subfigure}{.24\linewidth}
        \centering
        \includegraphics[width=\linewidth]{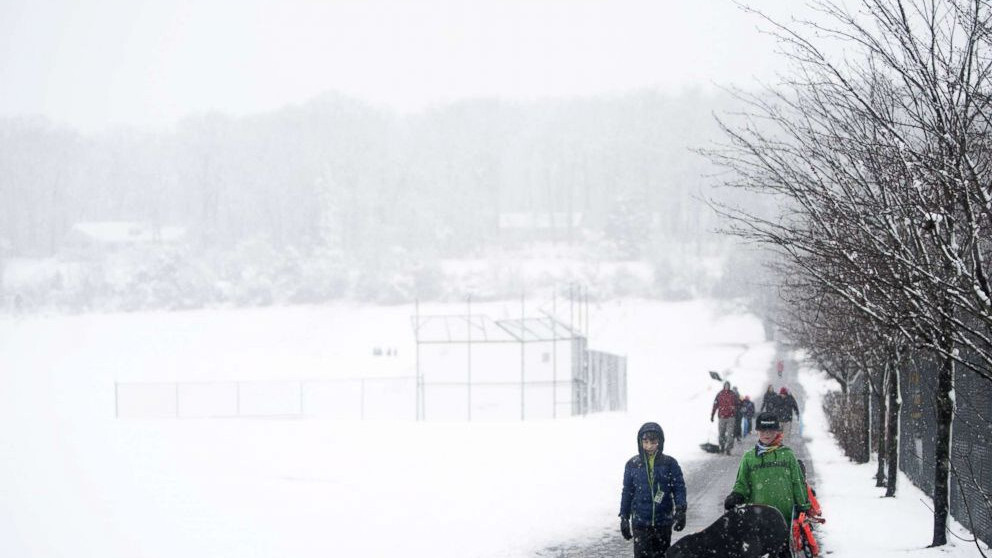}
    \end{subfigure}
    \begin{subfigure}{.24\linewidth}
        \centering
        \includegraphics[width=\linewidth]{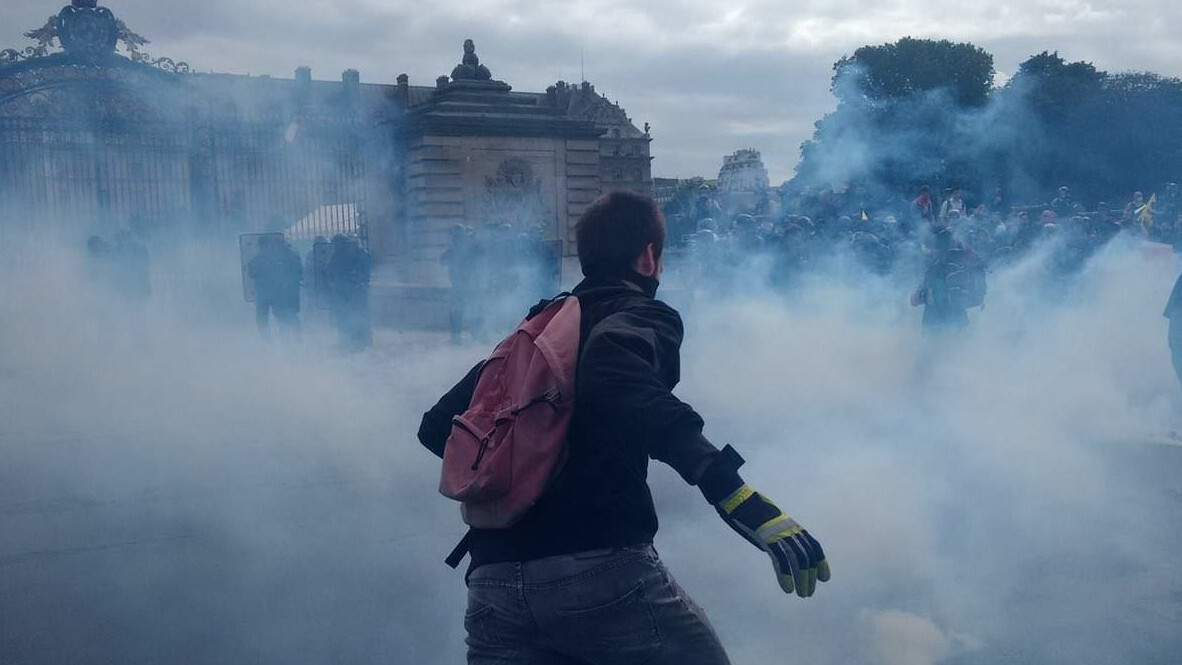}
    \end{subfigure}
    \begin{subfigure}{.24\linewidth}
        \centering
        \includegraphics[width=\linewidth]{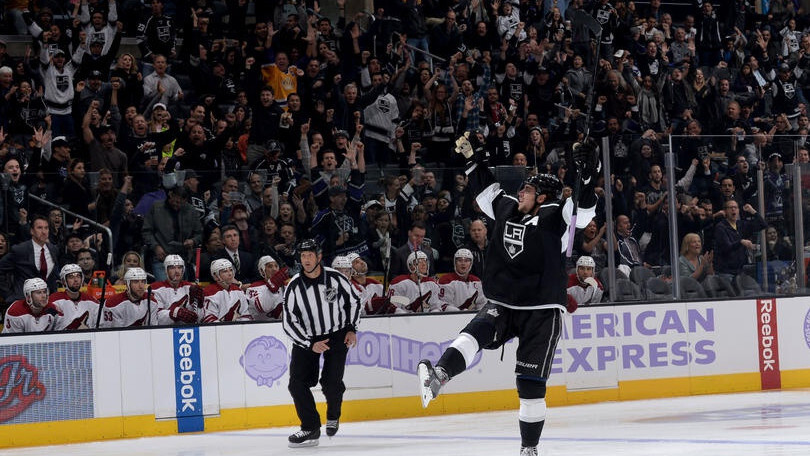}
    \end{subfigure}
    \begin{subfigure}{.24\linewidth}
        \centering
        \includegraphics[width=\linewidth]{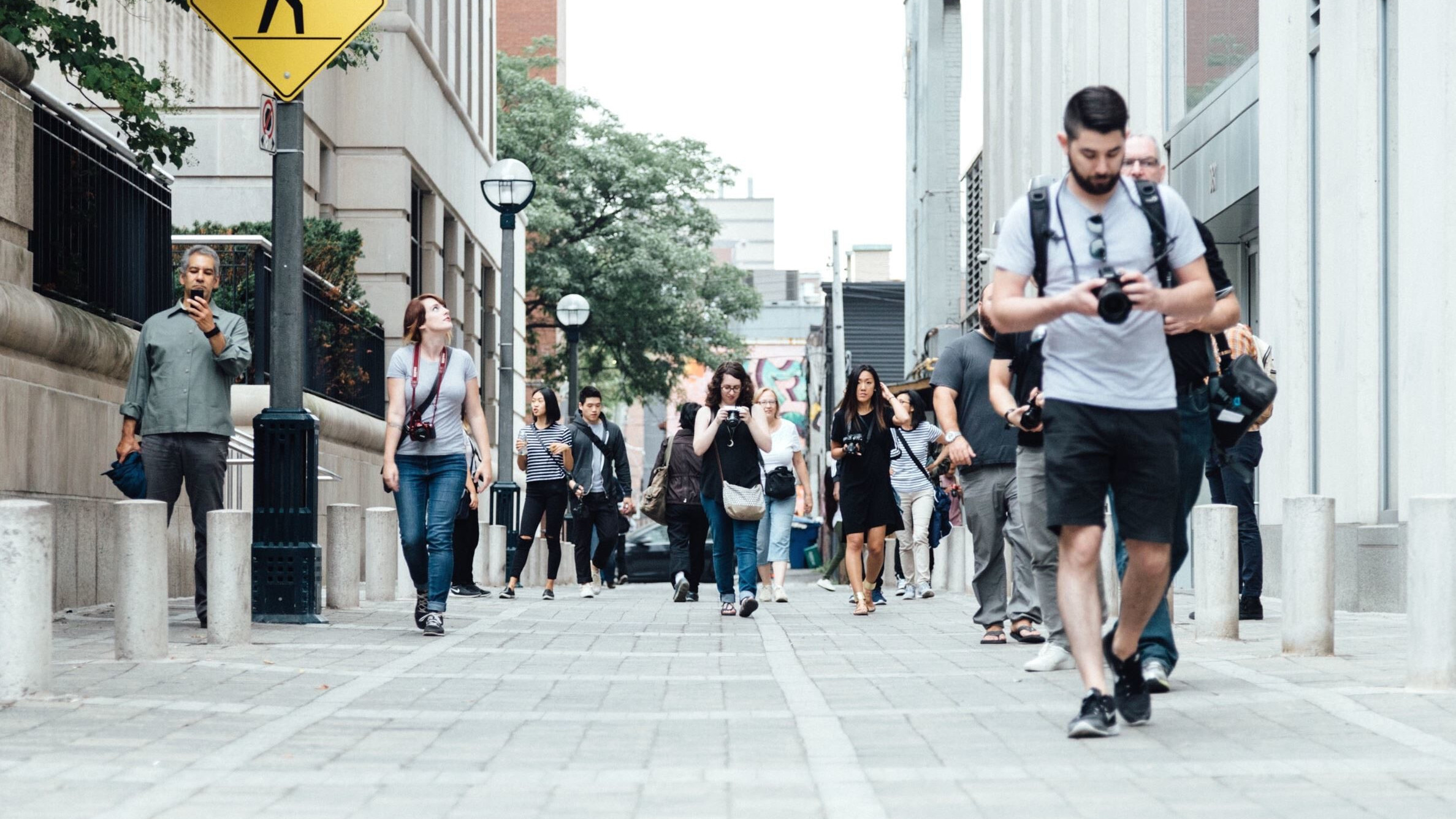}
    \end{subfigure}
    \begin{subfigure}{.24\linewidth}
        \centering
        \includegraphics[width=\linewidth]{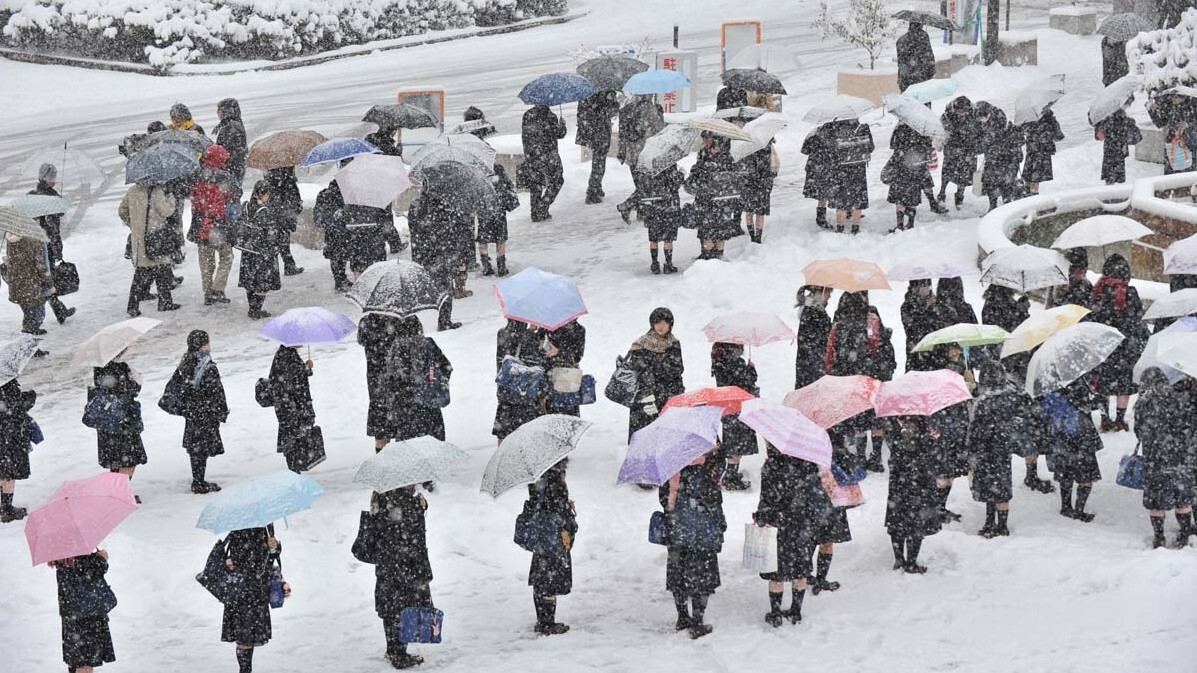}
    \end{subfigure}
    \begin{subfigure}{.24\linewidth}
        \centering
        \includegraphics[width=\linewidth]{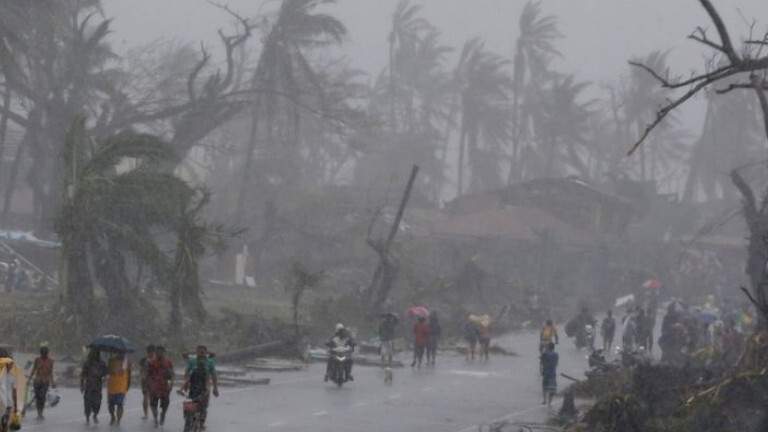}
    \end{subfigure}
    \begin{subfigure}{.24\linewidth}
        \centering
        \includegraphics[width=\linewidth]{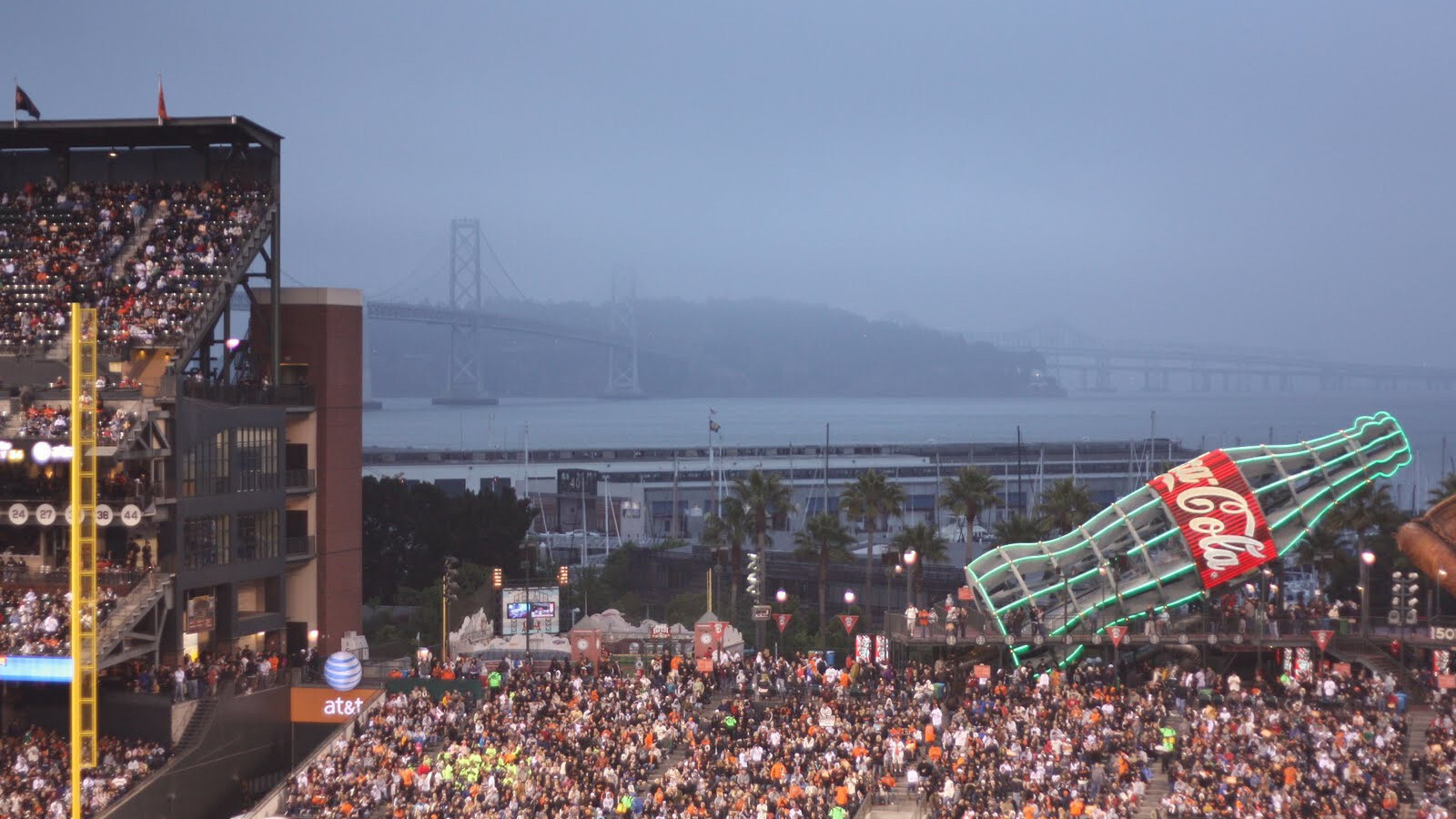}
        \caption{Stadium}
    \end{subfigure}
    \begin{subfigure}{.24\linewidth}
        \centering
        \includegraphics[width=\linewidth]{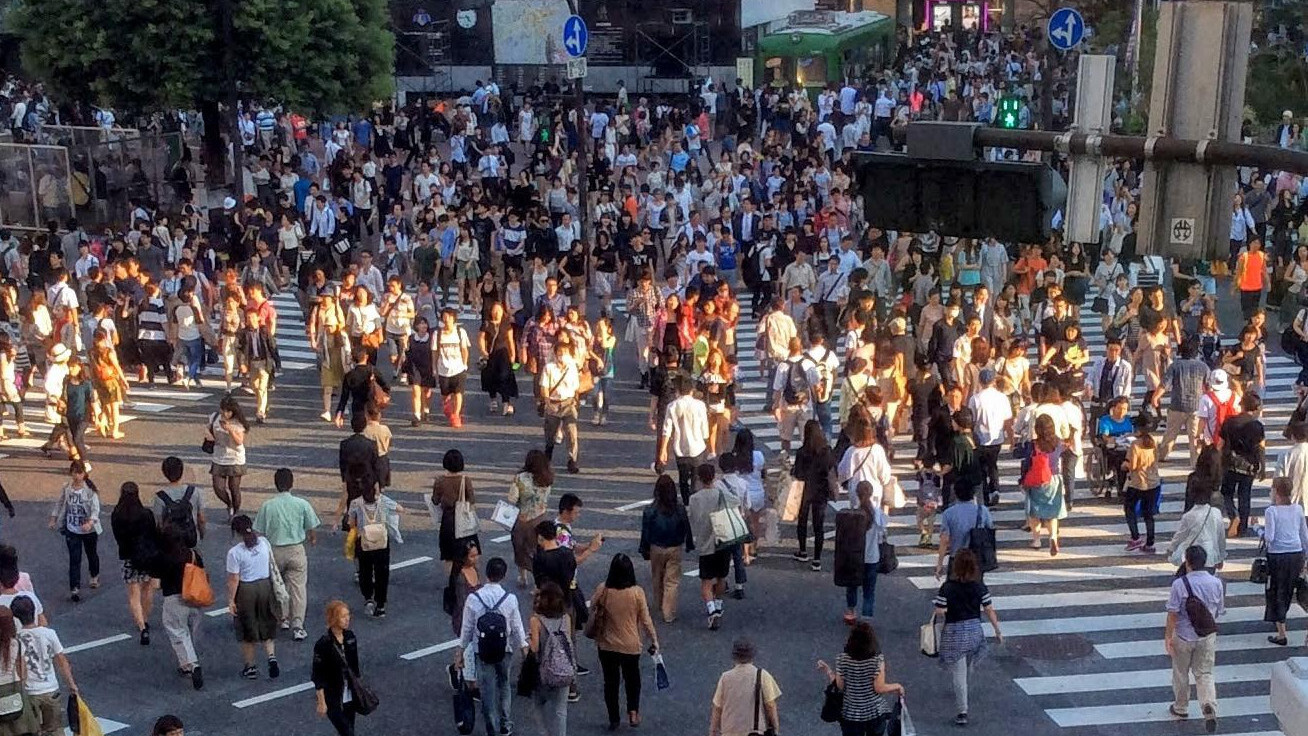}
        \caption{Street}
    \end{subfigure}
    \begin{subfigure}{.24\linewidth}
        \centering
        \includegraphics[width=\linewidth]{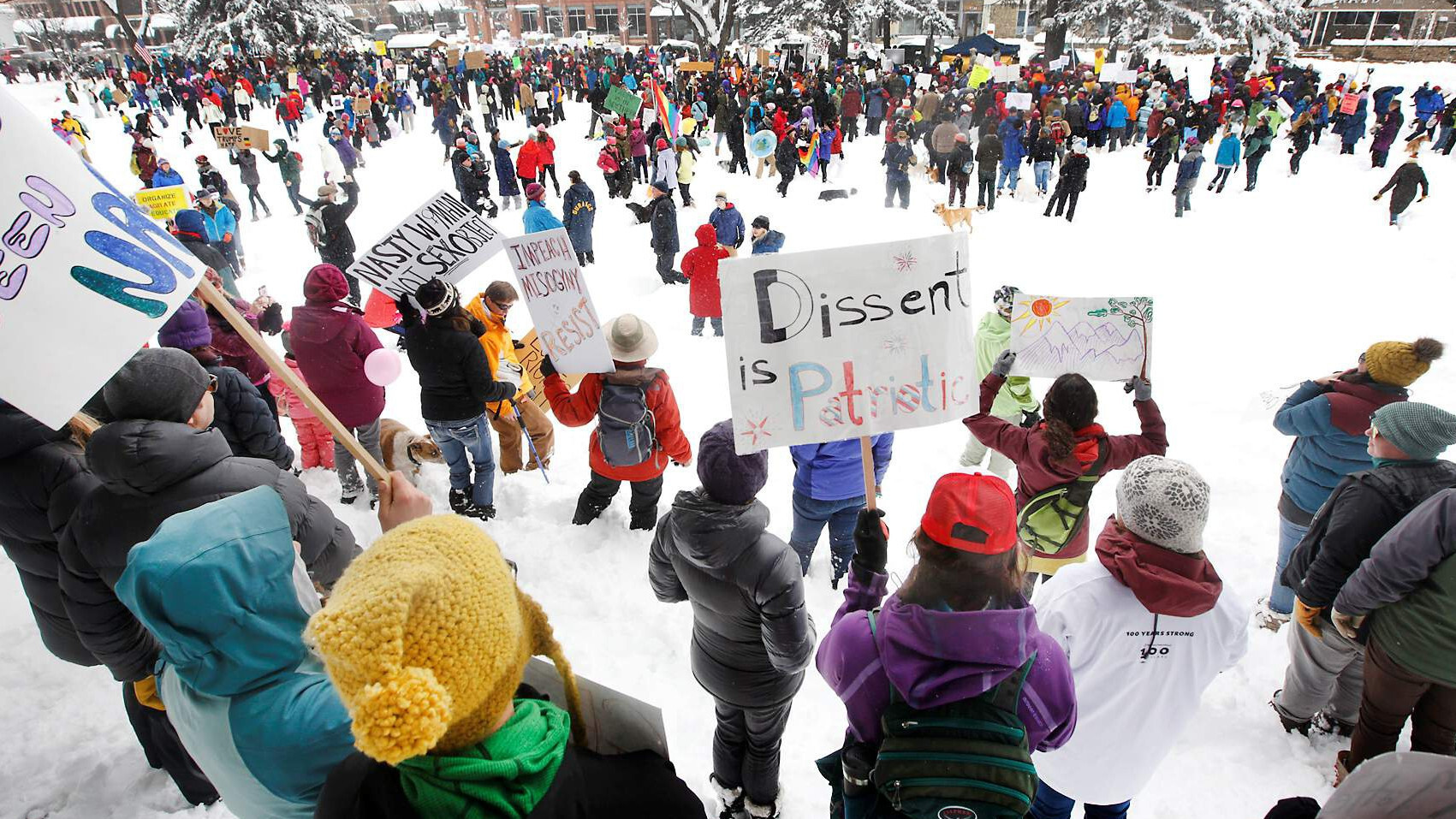}
        \caption{Snow}
    \end{subfigure}
    \begin{subfigure}{.24\linewidth}
        \centering
        \includegraphics[width=\linewidth]{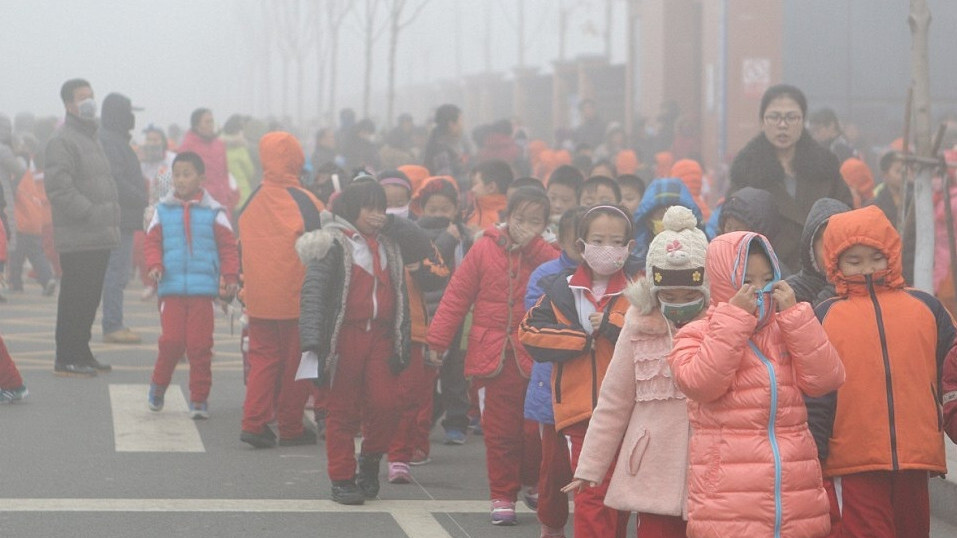}
        \caption{Fog/Haze}
    \end{subfigure}
    \caption{Sample images with different instance-level labels in JHU-Crowd++. }
    \label{fig:jhu}
\end{figure}

\begin{itemize}
    \item \textit{ShanghaiTech}~\cite{7780439} is composed of two parts, SHA (A) and SHB (B). SHA contains 300 training images and 182 testing images, while SHB contains 400 training images and 316 testing images. Images in SHA are collected from the Internet and feature highly crowded scenes. In contrast, SHB is captured from several streets in Shanghai, and its images exhibit crowd densities generally lower than those in SHA. 
    \item \textit{UCF-QNRF} (Q)~\cite{Idrees2018CompositionLF} consists of 1201 training images and 334 testing images. It is a challenging dataset with a large range of crowd densities, scenes, viewpoints and lighting conditions.
    \item \textit{JHU-Crowd++} ~\cite{Sindagi2020JHUCROWDLC} is a large-scale dataset containing 4372 images, with 2722, 500 and 1600 images in the training, validation and testing set, respectively. In addition, image-level labels are also provided in this dataset, including 16 types of scenes and 4 types of weather conditions. For scene annotations, there are 879 images labeled ``stadium'' (SD) and 573 labeled ``street'' (ST), which are utilized as two source domains. For weather annotations, we use 201 images labeled ``snow'' (SN) and 168 images labeled ``fog/haze'' (FH) as source domains. In each domain, 80\% of the images are selected as the training set while the rest 20\% are for testing. Data with the same label belong to a narrower distribution than mainstream datasets, thus are more challenging for SDG. Illustrative samples in these scene-specific and weather-specific domains are available in ~\cref{fig:jhu}.
\end{itemize}

Let $\mathcal{S}\rightarrow\mathcal{T}$ denote the case where $\mathcal{S}$ is the source domain and $\mathcal{T}$ is the target domain. The datasets are utilized in experiments under two distinct types of settings: 1) An entire dataset is regarded as a single domain: A $\rightarrow$ B / Q, B $\rightarrow$ A / Q and Q $\rightarrow$ A / B; 2) A dataset is partitioned into subsets according to image-level labels, and one subset constitutes a single domain: SD $\leftrightarrow$ SR and SN $\leftrightarrow$ FH. 

\begin{table*}[t]
    \centering
    \small
    \resizebox{\textwidth}{!}{
    \begin{tabular}{c|cc|cc|cc|cc|cc|cc|cc}
    \toprule
    \toprule
    \multicolumn{3}{c|}{Source $\rightarrow$ Target} & \multicolumn{2}{c|}{A $\rightarrow$ B} & \multicolumn{2}{c|}{A $\rightarrow$ Q} & \multicolumn{2}{c|}{B $\rightarrow$ A} & \multicolumn{2}{c|}{B $\rightarrow$ Q} & \multicolumn{2}{c|}{Q $\rightarrow$ A} & \multicolumn{2}{c}{Q $\rightarrow$ B}\\
    \midrule
        Method  & DA & DG & MAE & MSE & MAE & MSE & MAE & MSE & MAE & MSE & MAE & MSE & MAE & MSE\\
        \midrule
        BL \cite{9009503}      & \ding{55} & \ding{55} & 15.9 & 25.8 & 166.7 & 287.6 & 138.1 & 228.1 & 226.4 & 411.0 & - & - & - & -\\
        DMCount \cite{wang2020DMCount} & \ding{55} & \ding{55} & 23.1 & 34.9 & 134.4 & 252.1 & 143.9 & 239.6 & 203.0 & 386.1 & 73.4 & 135.1 & 14.3 & 27.5\\
        SASNet \cite{sasnet}  & \ding{55} & \ding{55} & 21.3 & 33.2 & 211.2 & 418.6 & 132.4 & 225.6 & 273.5 & 481.3 & 73.9 & 116.4 & 13.0 & 22.1\\
        ChfL \cite{shu2022crowd}  & \ding{55} & \ding{55} & 18.7 & 29.1 & 122.3 & 218.0 & 121.3 & 200.8 & 197.1 & 357.9 & 68.7 & 118.5 & 14.0 & 27.0\\
        MAN \cite{lin2022boosting}  & \ding{55} & \ding{55} & 22.1 & 32.8 & 138.8 & 266.3 & 133.6 & 255.6 & 209.4 & 378.8 & 67.1 & 122.1 & 12.5 & 22.2\\
        \midrule
        RBT \cite{10.1145/3394171.3413825} & \ding{51} & \ding{55} & 13.4 & 29.3 & 175.0 & 294.8 & 112.2 & 218.2 & 211.3 & 381.9 & - & - & - & -\\
        $\text{C}^2$MoT \cite{10.1145/3474085.3475230} & \ding{51} & \ding{55} & 12.4 & 21.1  & 125.7 & 218.3 & 120.7 & 192.0 & 198.9 & 368.0 & - & - & - & -\\
        FGFD \cite{10.1145/3503161.3548298} & \ding{51} & \ding{55} & 12.7 & 23.3 & 124.1 & 242.0 & 123.5 & 210.7 & 209.7 & 384.7 & 70.2 & 118.4 & 12.5 & 20.3\\
        DAOT \cite{10.1145/3581783.3611793} & \ding{51} & \ding{55} & 10.9 & 18.8 & 113.9 & 215.6 & - & - & - & - & 67.0 & 128.4 & 11.3 & 19.6\\
        FSIM \cite{10272678} & \ding{51} & \ding{55} & 11.1 & 19.3 & 105.3 & 191.1 & 120.3 & 202.6 & 194.9 & 324.5 & 66.8 & 111.5 & 11.0 & 19.7\\
        \midrule
        IBN \cite{Pan2018TwoAO} & \ding{55} & \ding{51} & 19.1 & 30.8 & 280.2 & 561.0 & 125.9 & 202.3 & 183.5 & 317.4 & 105.9 & 174.6 & 16.0 & 25.5\\
        SW \cite{pan2018switchable} & \ding{55} & \ding{51} & 20.2 & 30.4 & 285.5 & 431.0 & 126.7 & 193.8 & 200.7 & 333.2 & 102.4 & 168.8 & 19.0 & 32.9\\
        ISW \cite{Choi2021RobustNetID} & \ding{55} & \ding{51} & 23.9 & 37.7 & 215.6 & 399.7 & 156.2 & 291.5 & 263.1 & 442.2 & 83.4 & 136.0 & 22.1 & 34.7\\
        DG-MAN \cite{Mansilla2021DomainGV} & \ding{55} & \ding{51} & 17.3 & 28.7 & 129.1 & 238.2 & 130.7 & 225.1 & 182.4 & 325.8 & - & - & - & -\\
        DCCUS \cite{du2023domaingeneral} & \ding{55} & \ding{51} & 12.6 & 24.6 & 119.4 & 216.6 & 121.8 & 203.1 & 179.1 & 316.2 & 67.4 & 112.8 & \textbf{12.1} & \textbf{20.9}\\
        \textbf{MPCount (Ours)} & \ding{55} & \ding{51} & \textbf{11.4} & \textbf{19.7} & \textbf{115.7} & \textbf{199.8} & \textbf{99.6} & \textbf{182.9} & \textbf{165.6} & \textbf{290.4} & \textbf{65.5} & \textbf{110.1} & 12.3 & 24.1\\
        \bottomrule
        \bottomrule
        
    \end{tabular}}
    \caption{Comparison with the state-of-the-art methods on SHA (A), SHB (B) and UCF-QNRF (Q). Domain adaptation methods are only for reference since data in target domain might be used.}
    \label{tab:exp1}
\end{table*}

\begin{table*}[t]
    \centering
    \small
    \begin{tabular}{c|cc|cc|cc|cc|cc}
    \toprule
    \toprule
    \multicolumn{3}{c|}{Source $\rightarrow$ Target} & \multicolumn{2}{c|}{SD $\rightarrow$ SR} & \multicolumn{2}{c|}{SR $\rightarrow$ SD} & \multicolumn{2}{c|}{SN $\rightarrow$ FH} & \multicolumn{2}{c}{FH $\rightarrow$ SN}\\
    \midrule
        Method & DA & DG & MAE & MSE & MAE & MSE & MAE & MSE & MAE & MSE \\
        \midrule
        BL \cite{9009503} & \ding{55} & \ding{55} & 42.1 & 79.0 & 262.7 & 1,063.9 & 48.1 & 129.5 & 343.8 & 770.5\\
        MAN \cite{lin2022boosting} & \ding{55} & \ding{55} & 45.1 & 79.0 & 246.1 & 950.8 & 38.1 & 68.0 & 445.0 & 979.3\\
        \midrule
        DAOT \cite{10.1145/3581783.3611793} & \ding{51} & \ding{55} & 45.3 & 88.0 & 278.7 & 1,624.3 & 42.3 & 73.0 & 151.6 & 273.9\\
        \midrule
        IBN \cite{Pan2018TwoAO} & \ding{55} & \ding{51} & 92.2 & 178.0 & 318.1 & 1,420.4 & 109.7 & 267.7 & 491.8 & 1,110.4\\
        SW \cite{pan2018switchable} & \ding{55} & \ding{51} & 110.3 & 202.4 & 312.6 & 1,072.4 & 131.5 & 306.6 & 381.3 & 825.0\\
        ISW \cite{Choi2021RobustNetID} & \ding{55} & \ding{51} & 108.1 & 212.4 & 385.9 & 1,464.8 & 151.6 & 365.7 & 276.6 & 439.8\\
        DCCUS \cite{du2023domaingeneral} & \ding{55} & \ding{51} & 90.4 & 194.1 & 258.1 & 1,005.9 & 54.5 & 125.8 & 399.7 & 945.0\\
        \textbf{MPCount (Ours)} & \ding{55} & \ding{51} & \textbf{37.4} & \textbf{70.1} & \textbf{218.6} & \textbf{935.9} & \textbf{31.3} & \textbf{55.0} & \textbf{216.3} & \textbf{421.4} \\
        \bottomrule
        \bottomrule
        
    \end{tabular}
    \caption{Comparison with the state of the art on data in JHU-Crowd++ with labels ``Stadium''(SD), ``Street''(SR), ``Snow''(SN) and ``Fog/Haze''(FH). Domain adaptation methods are only for reference since data in target domain might be used.}
    \label{tab:exp2}
\end{table*}

\subsection{Implementation}\label{sec:imp}
We adopt VGG16-BN~\cite{Simonyan2014VeryDC} as the feature extractor in our model. 
The density regression head is a single convolution layer with $1\times 1$ filters, while the classification head consists of a $3\times 3$ and a $1\times 1$ convolution layers. Batch normalization is employed after all convolution layers except the last ones. 
For data augmentation, we apply three types of photometric transformations, color jittering, Gaussian blurring and sharpening. We also randomly crop image patches with a size of $320\times 320$ and adopt random horizontal flipping to both the original and augmented images simultaneously. We select AdamW~\cite{Loshchilov2017FixingWD} as the optimizer and OneCycleLR~\cite{Smith2017SuperconvergenceVF} as the learning rate scheduler with maximum learning rate set to 1e-3 and a maximum epoch of 300. The memory size $M$ is set to 1024 and the dimension $C$ is 256. We use a content error threshold $\alpha$ of 0.5, and the loss weights $\lambda_{cls}$ and $\lambda_{con}$ are both set to 10. 

\subsection{Evaluation Metrics}\label{sec:met}

We evaluate our method with mean absolute error (MAE) and mean squared error (MSE), defined as follows:

\begin{align}
    MAE &= \frac{1}{N}\sum_{i=1}^N| c_i - \hat{c_i}|, MSE &= \sqrt{\frac{1}{N}\sum_{i=1}^N(c_i - \hat{c_i})^2},
\end{align}
where $N$ is the number of testing images, $c_i$ is the ground truth count of the $i$-th image and $\hat{c_i}$ is the predicted count. Lower values of both metrics indicate better performance.

\begin{figure}[t]
    \centering
    \begin{subfigure}{.24\linewidth}
        \centering
        \includegraphics[width=\linewidth]{figs/cnt/img_1.jpg}
    \end{subfigure}
    \begin{subfigure}{.24\linewidth}
        \centering
        \includegraphics[width=\linewidth]{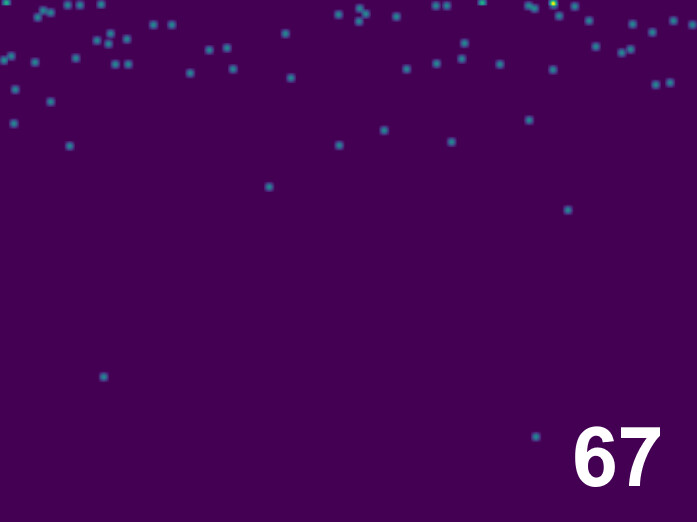}
    \end{subfigure}
    \begin{subfigure}{.24\linewidth}
        \centering
        \includegraphics[width=\linewidth]{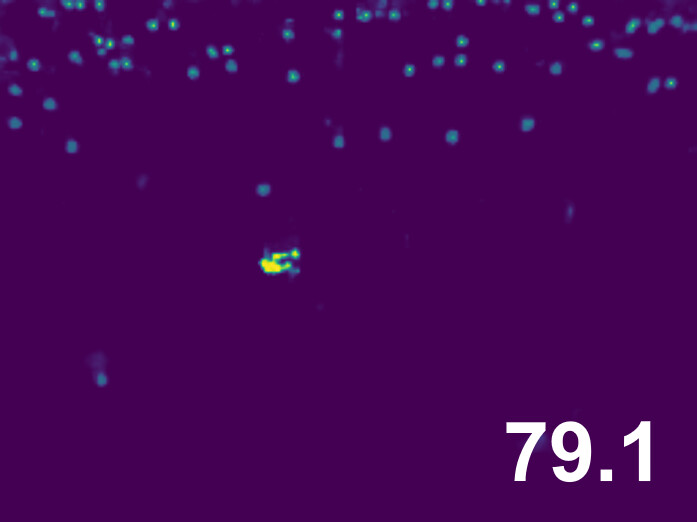}
    \end{subfigure}
    \begin{subfigure}{.24\linewidth}
        \centering
        \includegraphics[width=\linewidth]{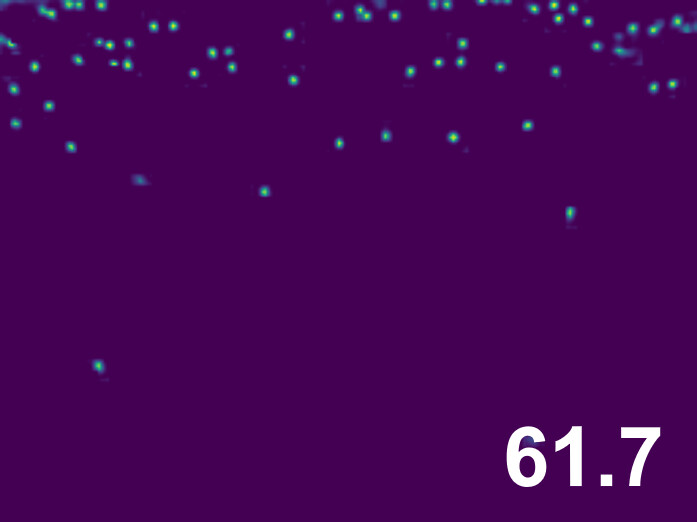}
    \end{subfigure}
    \begin{subfigure}{.24\linewidth}
        \centering
        \includegraphics[width=\linewidth]{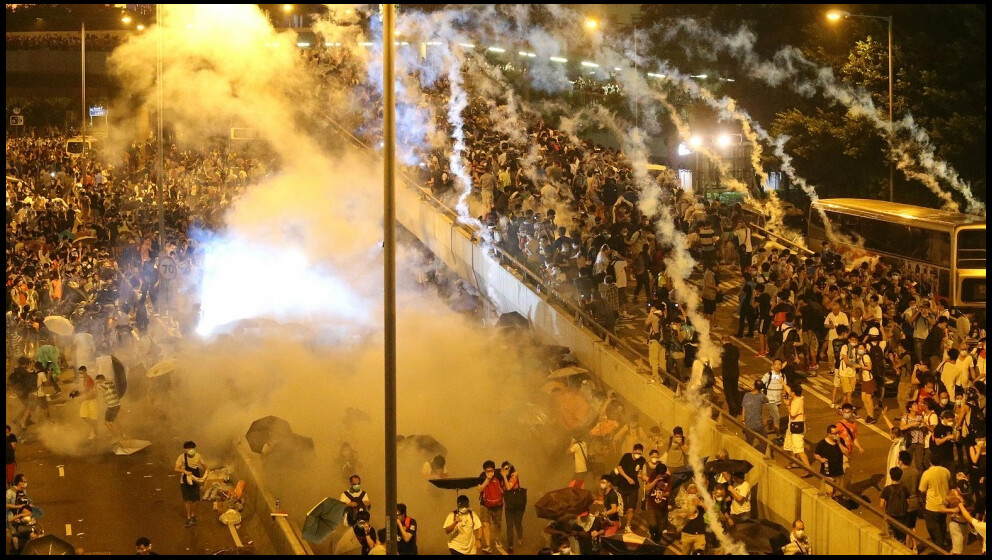}
    \end{subfigure}
    \begin{subfigure}{.24\linewidth}
        \centering
        \includegraphics[width=\linewidth]{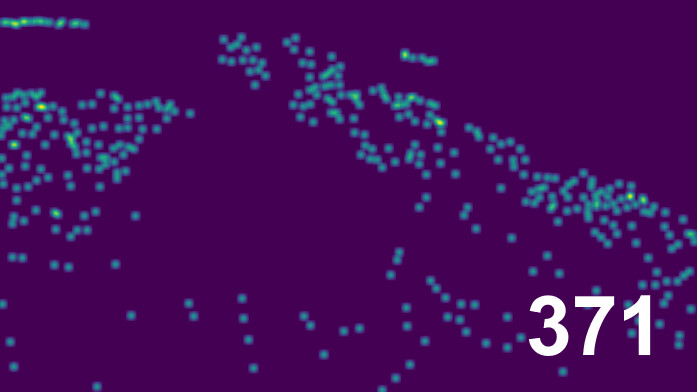}
    \end{subfigure}
    \begin{subfigure}{.24\linewidth}
        \centering
        \includegraphics[width=\linewidth]{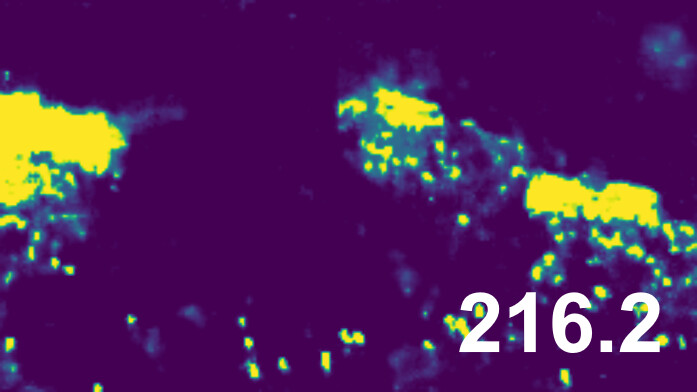}
    \end{subfigure}
    \begin{subfigure}{.24\linewidth}
        \centering
        \includegraphics[width=\linewidth]{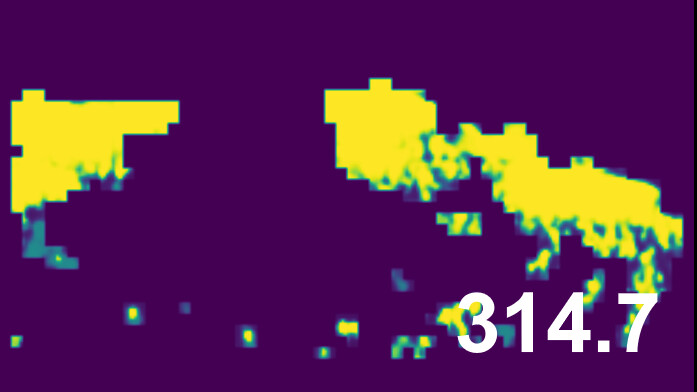}
    \end{subfigure}
    \begin{subfigure}{.24\linewidth}
        \centering
        \includegraphics[width=\linewidth]{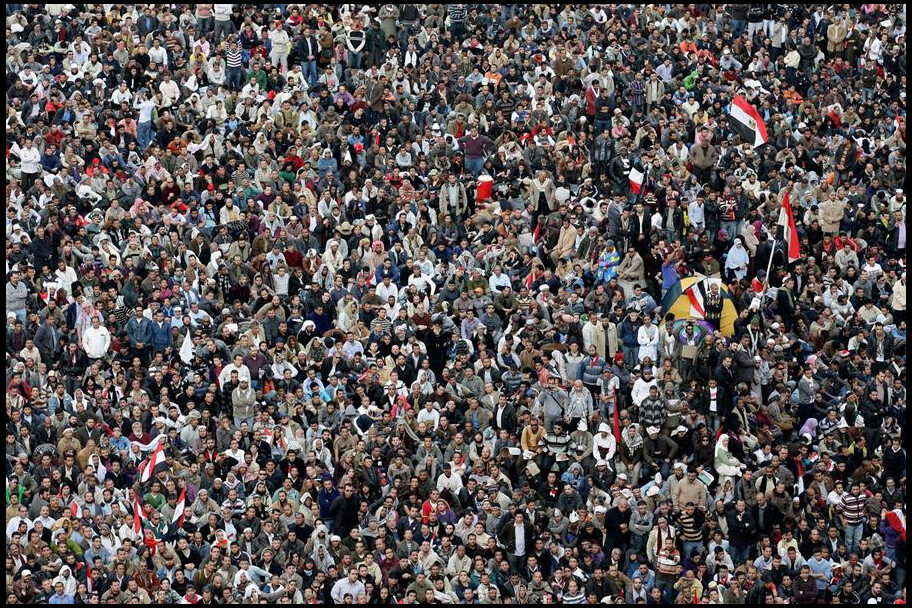}
        \caption{Image}
    \end{subfigure}
    \begin{subfigure}{.24\linewidth}
        \centering
        \includegraphics[width=\linewidth]{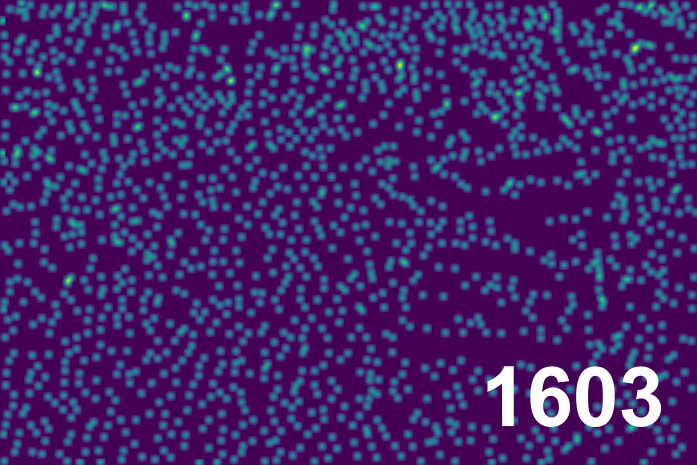}
        \caption{GT}
    \end{subfigure}
    \begin{subfigure}{.24\linewidth}
        \centering
        \includegraphics[width=\linewidth]{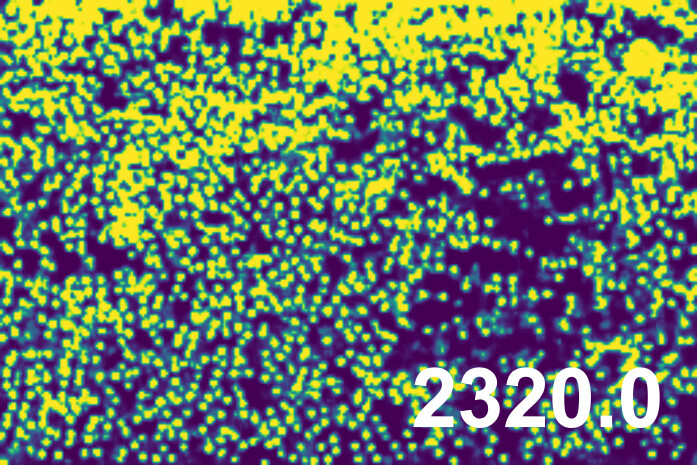}
        \caption{DCCUS}
    \end{subfigure}
    \begin{subfigure}{.24\linewidth}
        \centering
        \includegraphics[width=\linewidth]{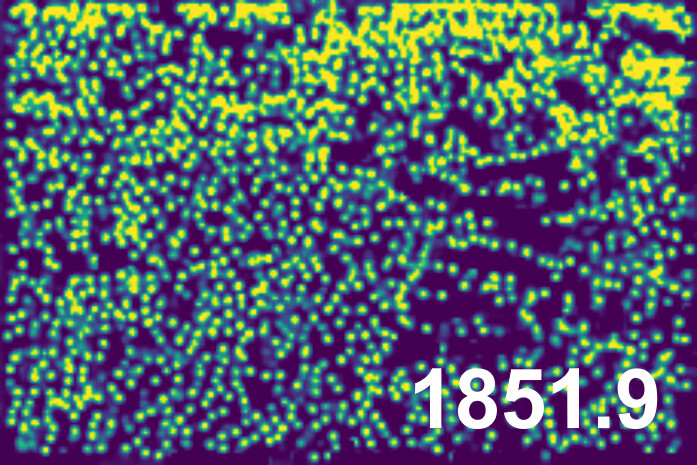}
        \caption{MPCount}
    \end{subfigure}
    \caption{Visualization results of DCCUS~\cite{du2023domaingeneral} and MPCount under different DG settings. First row: A $\rightarrow$ B; second row: SN $\rightarrow$ FH; third row: B $\rightarrow$ A.}
    \label{fig:pred}
\end{figure}

\subsection{Comparison with State of the Art}\label{sec:comp}

In this subsection, we compare our MPCount with state-of-the-art methods on different benchmarks, as illustrated in~\cref{fig:pred}. ~\cref{tab:exp1} shows the results on A $\rightarrow$ B / Q, B $\rightarrow$ A / Q and Q $\rightarrow$ A / B. The selected baselines can be divided into three categories: 
\begin{itemize}
    \item \textit{Fully supervised crowd counting}: BL~\cite{9009503}, DMCount~\cite{wang2020DMCount}, SASNet~\cite{sasnet}, ChfL~\cite{shu2022crowd} and MAN~\cite{lin2022boosting}.
    \item \textit{Domain adaptation for crowd counting}: RBT~\cite{10.1145/3394171.3413825}, $\text{C}^2$MoT~\cite{10.1145/3474085.3475230}, FGFD~\cite{10.1145/3503161.3548298}, DAOT~\cite{10.1145/3581783.3611793} and FSIM~\cite{10272678}.
    \item \textit{Single domain generalization}: IBN~\cite{Pan2018TwoAO}, SW~\cite{pan2018switchable}, ISW~\cite{Choi2021RobustNetID}, DG-MAN~\cite{Mansilla2021DomainGV} and DCCUS~\cite{du2023domaingeneral}.
\end{itemize}
All results are copied from previous papers except IBN, SW and ISW, which are feature normalization/whitening-based SDG methods originally designed for classification or segmentation and adapted to crowd counting by us. Domain adaptation methods are listed only for reference, as data from target domains are usually visible during adaptation. 

Our MPCount outperforms all the DG methods under most of these settings, including an significant error reduction by 18.2\% on B $\rightarrow$ A. It is notable that MPCount exhibits outstanding performance in some cases even compared to DA methods, providing compelling evidence for the effectiveness of our design. 

In ~\cref{tab:exp2}, we further conduct experiments on scene-specific domains SD $\leftrightarrow$ SR and weather-specific domains SN $\leftrightarrow$ FH. BL~\cite{9009503}, MAN~\cite{lin2022boosting}, DAOT~\cite{10.1145/3581783.3611793} and DCCUS~\cite{du2023domaingeneral} are trained using their released code, while IBN~\cite{Pan2018TwoAO}, SW~\cite{pan2018switchable} and ISW~\cite{Choi2021RobustNetID} are adapted by ourselves. We observe that the normalization/whitening-based methods cannot achieve satisfactory results, possibly because useful information is eliminated from feature statistics. DCCUS also obtains worse results than fully supervised methods, reflecting that the clustering procedure may fail to produce meaningful sub-domains when the source domain distribution is narrow. In contrast, our MPCount still performs the best among all tested methods, demonstrating its superiority in challenging conditions with narrow domain distributions. 

\subsection{Ablation Studies}\label{sec:abl}
In this section, we conduct ablation studies and other analysis under the setting A $\rightarrow$ B / Q. 

\noindent \textit{Effects of model components.} We start from a simple encoder-decoder structured baseline model and add or remove each component individually to validate its effectiveness. The experimental results are displayed in ~\cref{tab:ablation}. 

\begin{table}[t]
    \centering
    \begin{tabularx}{\linewidth}{Y Y Y Y|Y Y|Y Y}
        \toprule
        \toprule
        \multicolumn{4}{c|}{Components} & \multicolumn{2}{c|}{A $\rightarrow$ B} & \multicolumn{2}{c}{A $\rightarrow$ Q} \\
        \midrule
        AMB & CEM & ACL & PC & MAE & MSE & MAE & MSE\\
        \midrule
        & & & & 21.6 & 39.1 & 200.9 & 325.5\\
        \ding{51} & & & & 18.9 & 32.0 & 176.3 & 292.8\\
        \ding{51} & \ding{51} & & & 15.3 & 25.5 & 156.8 & 232.7\\
        \ding{51} & & \ding{51} & & 16.9 & 28.7 & 172.5 & 281.6\\
        \ding{51} & \ding{51} & \ding{51} & & 14.1 & 24.3 & 170.6 & 286.8\\
        \midrule
        & & & \ding{51} & 14.4 & 30.5 & 144.2 & 262.6\\
        \ding{51} & & & \ding{51} & 13.4 & 25.2 & 133.4 & 224.0\\
        \ding{51} & \ding{51} & & \ding{51} & 13.3 & 23.4 & 126.9 & 222.3\\
        \ding{51} & \ding{51} & \ding{51} & \ding{51} & \textbf{11.4} & \textbf{19.7} & \textbf{115.7} & \textbf{199.8} \\
        \bottomrule
        \bottomrule
    \end{tabularx}
    \captionof{table}{Ablation studies on the model components. }
    \label{tab:ablation}
\end{table}

\begin{table}[t]
    \centering
    \begin{tabular}{c c|c c|c c}
        \toprule
        \toprule
        Parameter & Property & \multicolumn{2}{c|}{A $\rightarrow$ B} & \multicolumn{2}{c}{A $\rightarrow$ Q} \\
        \midrule
        $\alpha$ & PDE(\%) & MAE & MSE & MAE & MSE\\
        \midrule
        0.1 & 28.6 & 16.7 & 25.9 & 157.3 & 338.5\\
        0.3 & 14.6 & 12.4 & 20.8 & 132.1 & 223.6\\
        0.5 & 10.5 & \textbf{11.4} & \textbf{19.7} & \textbf{115.7} & \textbf{199.8}\\
        0.7 & 5.9 & 11.9 & 23.8 & 123.8 & 214.1\\
        0.9 & 3.6 & 13.8 & 27.2 & 124.7 & 209.6\\
        \bottomrule
        \bottomrule
    \end{tabular}
    \captionof{table}{Ablation studies on the error threshold $\alpha$ in~\cref{eq:1}.}
    \label{tab:theta}
\end{table}

\begin{itemize}
    \item \textit{Attention Memory Bank}: After adding AMB, the performance consistently improves on both target domains, as shown in ~\cref{tab:ablation}. This validates that AMB can effectively help the model generalize by reconstructing domain-invariant features for density regression.
    \item \textit{Content Error Mask \& Attention Consistency Loss}: Next, we test the effects of CEM and ACL on AMB. The results in ~\cref{tab:ablation} show that both components generally improves the performance, especially when PC is present. 
    \item \textit{Patch-wise Classification}: Finally, models with PC outperforms their counterparts without PC, as visible in ~\cref{tab:ablation}. This suggests that such an auxiliary task is useful for DG by compensating for ambiguous pixel-level labels with accurate patch-level labels. 
\end{itemize}

\noindent \textit{Effects of $\alpha$.} The threshold of CEM $\alpha$ mentioned in ~\cref{eq:1} controls the Portion of Diminished Elements (PDE) in features. According to the results in ~\cref{tab:theta}, $\alpha=0.5$ produces the best performance under all settings, with a PDE of 10.5\%. It is also indicated that using a too large value of $\alpha$ (0.9) may be inadequate to filter out domain-related information, while a too small value (0.1) would let useful information eliminated. 

\noindent \textit{Effects of $M$ and $C$.} The number of memory vectors $M$ and dimension of each vector $C$ are two important parameters of AMB. To demonstrate their effects independently, we adjust one parameter while keeping the other fixed. The results in ~\cref{tab:size} shows that among all variations of parameters, $M=1024,C=256$ is the optimal choice. 

\begin{figure}[t]
    \centering
    \begin{subfigure}{.155\linewidth}
        \centering
        \includegraphics[width=\linewidth]{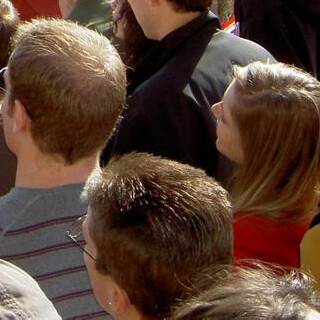}
    \end{subfigure}
    \begin{subfigure}{.155\linewidth}
        \centering
        \includegraphics[width=\linewidth]{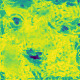}
    \end{subfigure}
    \begin{subfigure}{.155\linewidth}
        \centering
        \includegraphics[width=\linewidth]{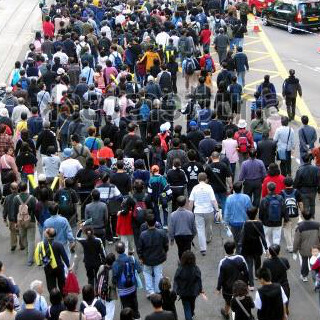}
    \end{subfigure}
    \begin{subfigure}{.155\linewidth}
        \centering
        \includegraphics[width=\linewidth]{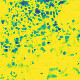}
    \end{subfigure}
    \begin{subfigure}{.155\linewidth}
        \centering
        \includegraphics[width=\linewidth]{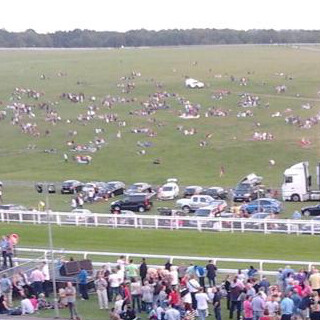}
    \end{subfigure}
    \begin{subfigure}{.155\linewidth}
        \centering
        \includegraphics[width=\linewidth]{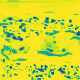}
    \end{subfigure}

    \caption{Visualized CEMs. Yellower means less filtered.}
    \label{fig:cem}
\end{figure}

\begin{table}[t]
    \centering
    \begin{tabular}{c c|c c|c c}
        \toprule
        \toprule
        \multicolumn{2}{c|}{Parameters} & \multicolumn{2}{c|}{A $\rightarrow$ B} & \multicolumn{2}{c}{A $\rightarrow$ Q} \\
        \midrule
        $M$ & $C$ &MAE & MSE & MAE & MSE\\
        \midrule
        256  & 256 & 15.9 & 24.0 & 138.4 & 226.1\\
        512  & 256 & 13.7 & 28.1 & 124.4 & 217.2\\
        1024 & 256 & \textbf{11.4} & \textbf{19.7} & \textbf{115.7} & \textbf{199.8}\\
        2048 & 256 & 12.0 & 23.3 & 128.1 & 218.1\\
        \midrule
        1024 & 128  & 16.3 & 28.6 & 127.4 & 223.3\\
        1024 & 256  & \textbf{11.4} & \textbf{19.7} & \textbf{115.7} & \textbf{199.8}\\
        1024 & 512  & 14.3 & 27.7 & 130.1 & 250.1\\
        1024 & 1024 & 14.2 & 24.8 & 128.4 & 220.2\\
        \bottomrule
        \bottomrule
    \end{tabular}
    \captionof{table}{Ablation studies on the number of memory vectors $M$ and vector size $C$.}
    \label{tab:size}
\end{table}
\noindent \textit{Visualizations of CEMs.} We present visualizations of CEMs in~\cref{fig:cem}, each summed along the channel dimension. The results indicate that crowd information is generally more sensitive to domain shift than environments.

\begin{figure}[t]
    \centering
    \begin{subfigure}{.24\linewidth}
        \centering
        \includegraphics[width=\linewidth]{figs/pcm/img1.jpg}
    \end{subfigure}
    \begin{subfigure}{.24\linewidth}
        \centering
        \includegraphics[width=\linewidth]{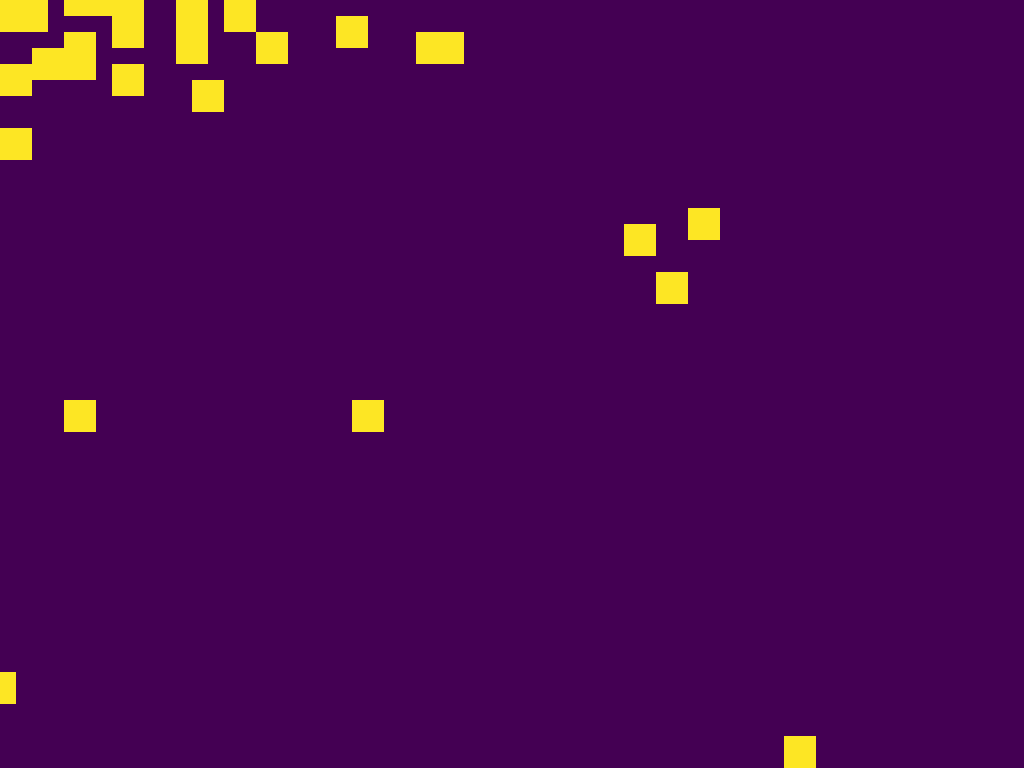}
    \end{subfigure}
    \begin{subfigure}{.24\linewidth}
        \centering
        \includegraphics[width=\linewidth]{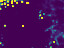}
    \end{subfigure}
    \begin{subfigure}{.24\linewidth}
        \centering
        \includegraphics[width=\linewidth]{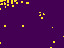}
    \end{subfigure}
    \begin{subfigure}{.24\linewidth}
        \centering
        \includegraphics[width=\linewidth]{figs/pcm/img2.jpg}
    \end{subfigure}
    \begin{subfigure}{.24\linewidth}
        \centering
        \includegraphics[width=\linewidth]{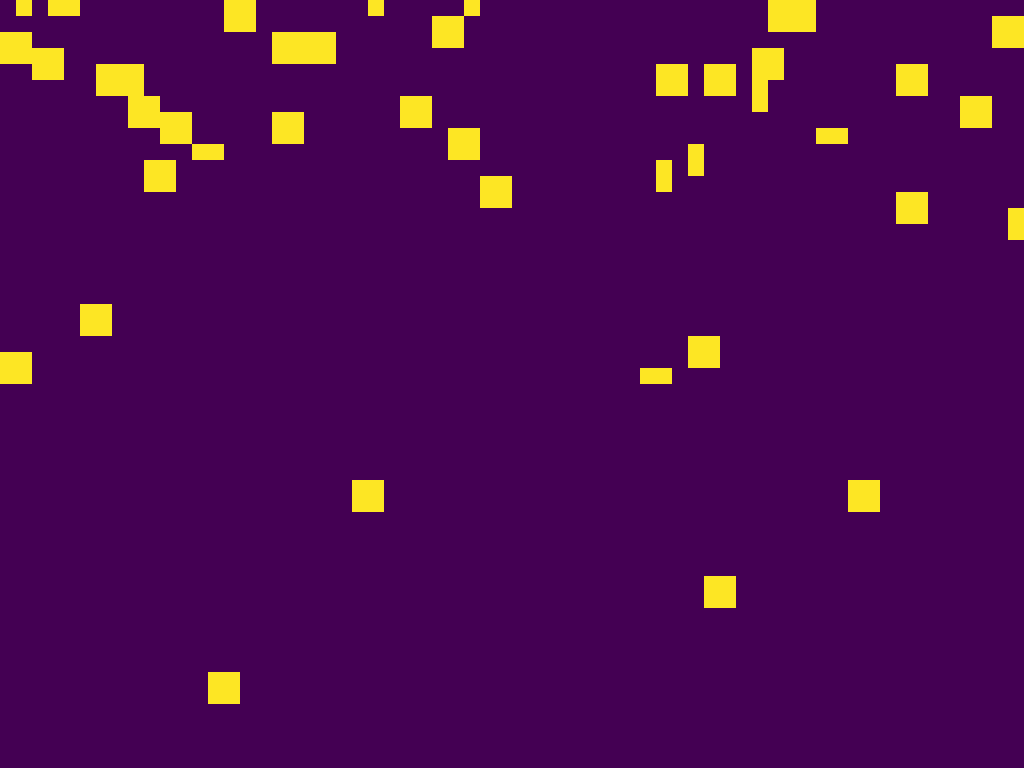}
    \end{subfigure}
    \begin{subfigure}{.24\linewidth}
        \centering
        \includegraphics[width=\linewidth]{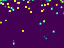}
    \end{subfigure}
    \begin{subfigure}{.24\linewidth}
        \centering
        \includegraphics[width=\linewidth]{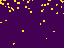}
    \end{subfigure}
    \begin{subfigure}{.24\linewidth}
        \centering
        \includegraphics[width=\linewidth]{figs/pcm/img3.jpg}
        \caption{Image}
    \end{subfigure}
    \begin{subfigure}{.24\linewidth}
        \centering
        \includegraphics[width=\linewidth]{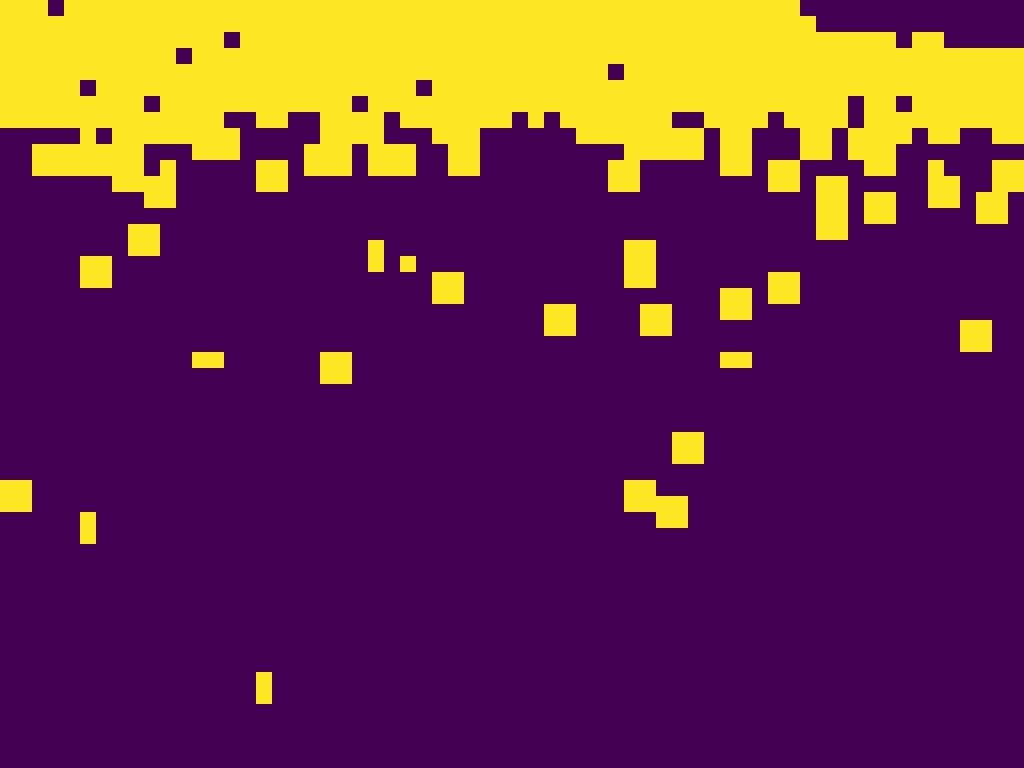}
        \caption{GT}
    \end{subfigure}
    \begin{subfigure}{.24\linewidth}
        \centering
        \includegraphics[width=\linewidth]{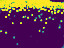}
        \caption{Predicted}
    \end{subfigure}
    \begin{subfigure}{.24\linewidth}
        \centering
        \includegraphics[width=\linewidth]{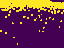}
        \caption{Binarized}
    \end{subfigure}
    
    \caption{Visualization of (a) sample images from SHB, (b) the ground-truth PCMs, (c) PCMs predicted by MPCount trained on SHA and (d) predicted PCMs after binarization.}
    \label{fig:pcm0}
\end{figure}

\noindent \textit{Visualizations of PCMs.} We visualize the predicted and binarized PCMs from MPCount for qualitative analysis. As shown in ~\cref{fig:pcm0}, MPCount predicts accurate PCMs that provide reliable information of crowd locations. In the raw predicted PCMs, there are unconfident predictions with low classification scores, which are filtered out in the binarized PCMs. The results verify that PC can effectively mitigate the label ambiguity problem and contribute to a more robust crowd counting model against domain shift.

%% file: sec/5_conclu.tex
\section{Conclusion}
This paper proposes MPCount to address single domain generalization for crowd counting with possibly narrow source distribution. MPCount tackles two unique challenges, namely, density regression and label ambiguity. We propose an attention memory bank to reconstruct domain-invariant features for regression, with the content error mask eliminating domain-related content information and the attention consistency loss ensuring the consistency of memory vectors. Patch-wise classification is a novel auxiliary task to mitigate the ambiguous pixel-level labels with accurate information, enhancing the robustness of density predictions. Extensive experiments on well-known datasets show that MPCount achieves significantly the best results on various benchmarks by outperforming the state of the art under different unseen scenarios and narrow source domain. 

\noindent \textbf{Acknowledgement.} We would like to thank the Turing AI Computing Cloud (TACC)~\cite{tacc} and HKUST iSING Lab for providing us computation resources on their platform.

%% file: sec/X_suppl.tex
\clearpage
\setcounter{page}{1}
\maketitlesupplementary

\section{More Implementation Details}
This section presents more details about our implementations of MPCount and other DG methods adapted to crowd counting.

\subsection{MPCount}
\begin{figure}[h]
    \centering
    \includegraphics[width=\linewidth]{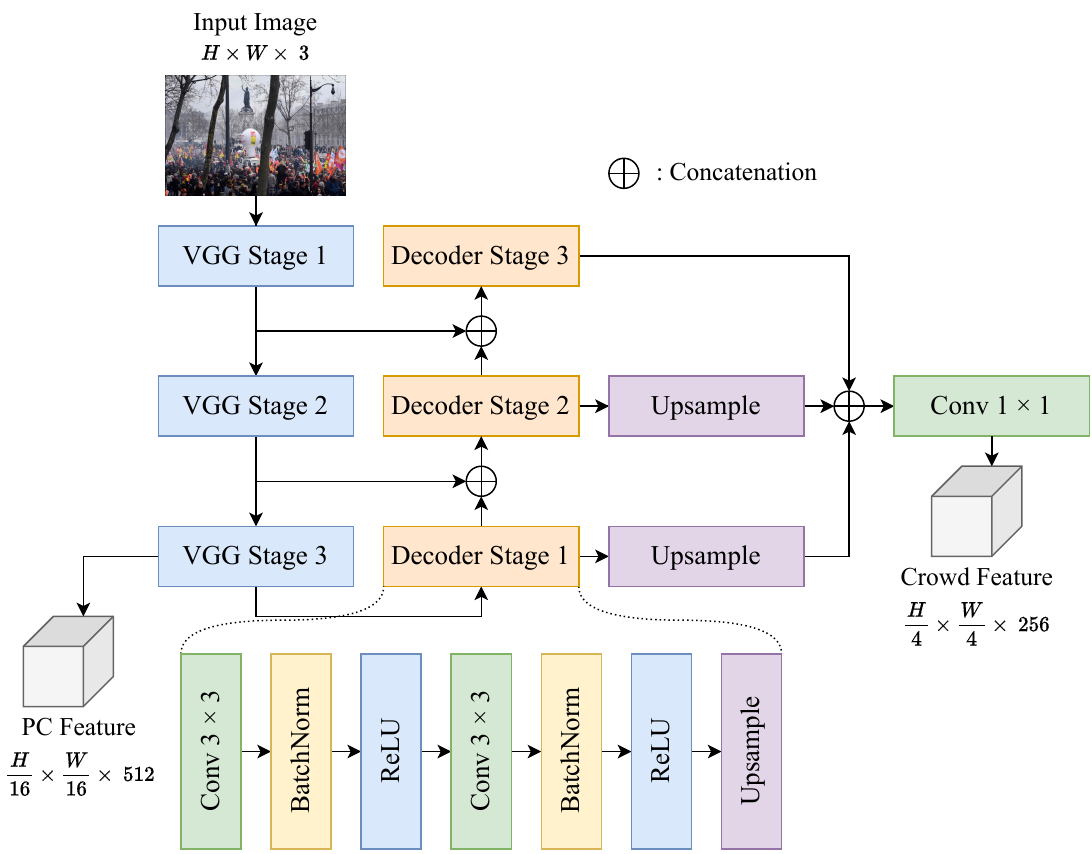}
    \caption{Structure of the model backbone of MPCount}
    \label{fig:backbone}
\end{figure}

\cref{fig:backbone} illustrates the detailed structure of our model backbone adapted from VGG-16BN. Skip connections are added to integrate multi-level information, enhancing the expressive power of crowd features. Note that we only utilize the deepest level of features for PC prediction, as shallow features usually encode more style information~\cite{Pan2018TwoAO} and thus are more vulnerable to domain shift. 

For data augmentation, we use three types of photometric transformations: (1) \textit{Color jittering} with probability 0.8 (brightness of 0.5, contrast of 0.2, saturation of 0.2, and hue of 0.1); (2) \textit{Gaussian Blurring} with probability 0.5 (kernel size of 3 and sigma of 1); (3) \textit{Sharpening} with probability 0.5 (sharpness factor of 5). 

All experiments on MPCount are conducted on four NVIDIA GeForce 3090Ti GPUs, and implementations are based on Python 3.10 and PyTorch 2.0. During training, the batch size and number of workers are both set to 16. The random seed is set to 2023. Following~\cite{gao2019c}, the ground-truth density maps are multiplied by 1,000.

\subsection{Adapted Methods}
\begin{figure}[h]
    \centering
    \includegraphics[width=\linewidth]{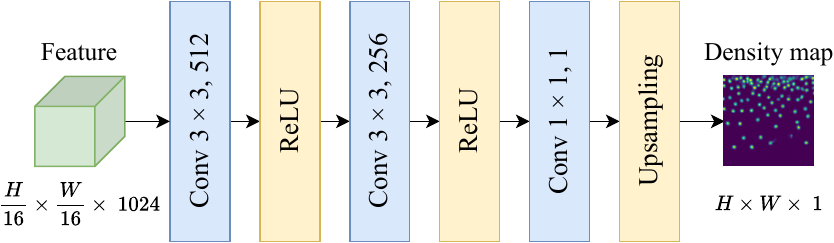}
    \caption{Regression head used for the adapted methods}
    \label{fig:head}
\end{figure}

IBN~\cite{Pan2018TwoAO}, SW~\cite{pan2018switchable} and ISW~\cite{Choi2021RobustNetID} are three DG methods originally designed for image classification or semantic segmentation. We select ResNet-50~\cite{He2015DeepRL} truncated after conv4\_6 as their common backbone and connect it with a density regression head consisting of three convolutional layers, as illustrated in~\cref{fig:head}.

\section{More Quantitative Results}
This section includes more quantitative results of MPCount. 

\subsection{Quantitative Results of PCM prediction}
\cref{tab:pcm_acc} shows the quantitative results of PCM prediction from MPCount on SHA and SHB. Three common metrics for image segmentation are used, namely mAcc, mIoU and mDice. Since PC is only an auxiliary task to assist crowd density prediction, and no attention is specifically paid to improve the accuracy of PCMs, we do not require and expect the performance to be perfect. However, as shown in~\cref{tab:pcm_acc}, satisfactory results are still achieved to support the function of PCMs to filter out areas without human head.

\begin{table}[h]
    \centering
    \begin{tabular}{c|c c c c}
        \toprule
        \toprule
        Metric & mAcc(\%) & mIoU(\%) & mDice(\%)\\
        \midrule
        A $\rightarrow$ B & 96.9 & 82.3 & 90.3\\
        B $\rightarrow$ A & 87.7 & 73.7 & 85.1\\
        \bottomrule
        \bottomrule
    \end{tabular}
    \captionof{table}{Quantitative Results of PCM prediction on SHA (A) and SHB (B)}
    \label{tab:pcm_acc}
\end{table}

\section{More Visualization Results}
\begin{figure}[h]
    \centering
    \begin{subfigure}{.24\linewidth}
        \centering
        \includegraphics[width=\linewidth]{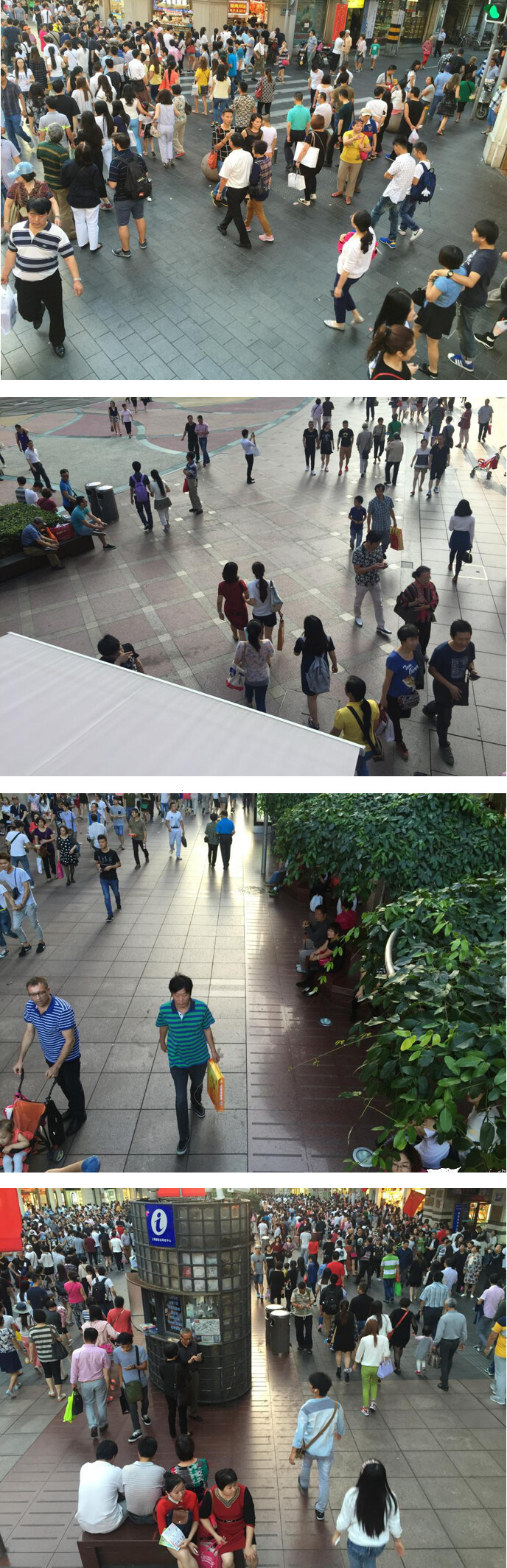}
        \caption{Image}
    \end{subfigure}
    \begin{subfigure}{.24\linewidth}
        \centering
        \includegraphics[width=\linewidth]{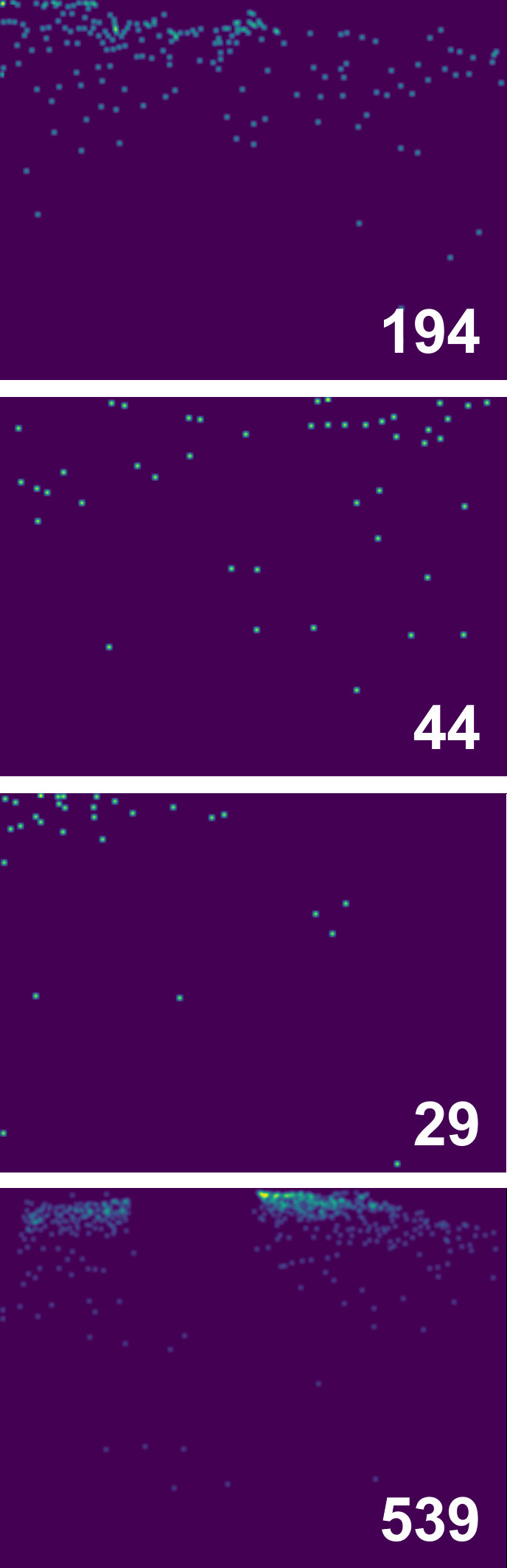}
        \caption{GT}
    \end{subfigure}
    \begin{subfigure}{.24\linewidth}
        \centering
        \includegraphics[width=\linewidth]{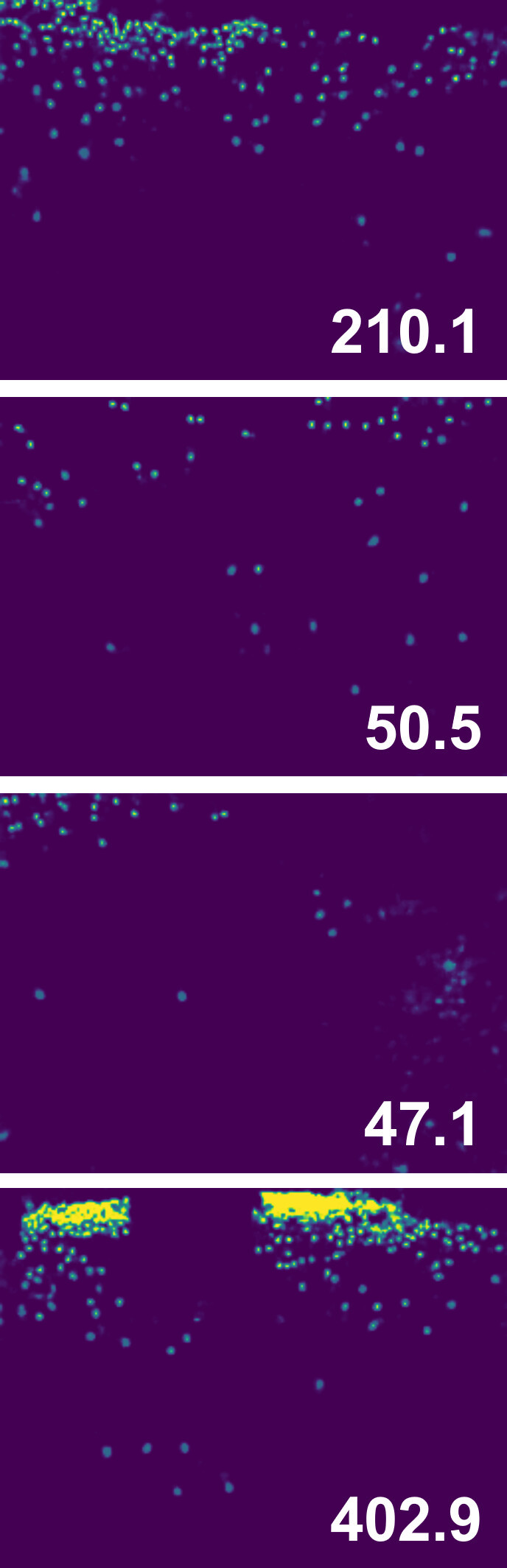}
        \caption{DCCUS}
    \end{subfigure}
    \begin{subfigure}{.24\linewidth}
        \centering
        \includegraphics[width=\linewidth]{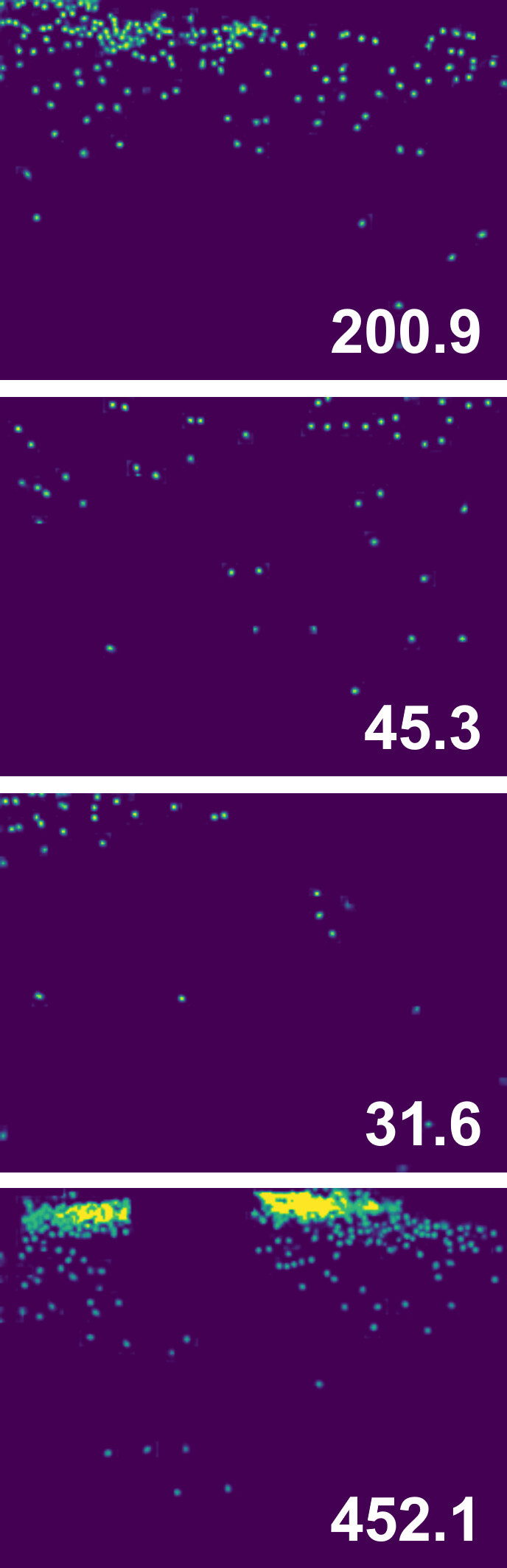}
        \caption{MPCount}
    \end{subfigure}
    \caption{Visualization results on A $\rightarrow$ B}
    \label{fig:a2b}
\end{figure}

This section shows more visualization results of predicted density maps and PCMs from our MPCount. We use DCCUS~\cite{du2023domaingeneral} as the baseline for comparison. 
Visualization results of density maps are shown in~\cref{fig:a2b}~$\sim$~\cref{fig:fog}, and PCMs are displayed in~\cref{fig:pcm}. 

\subsection{Visualization of Density Maps}
\begin{figure}[h]
    \centering
    \begin{subfigure}{.24\linewidth}
        \centering
        \includegraphics[width=\linewidth]{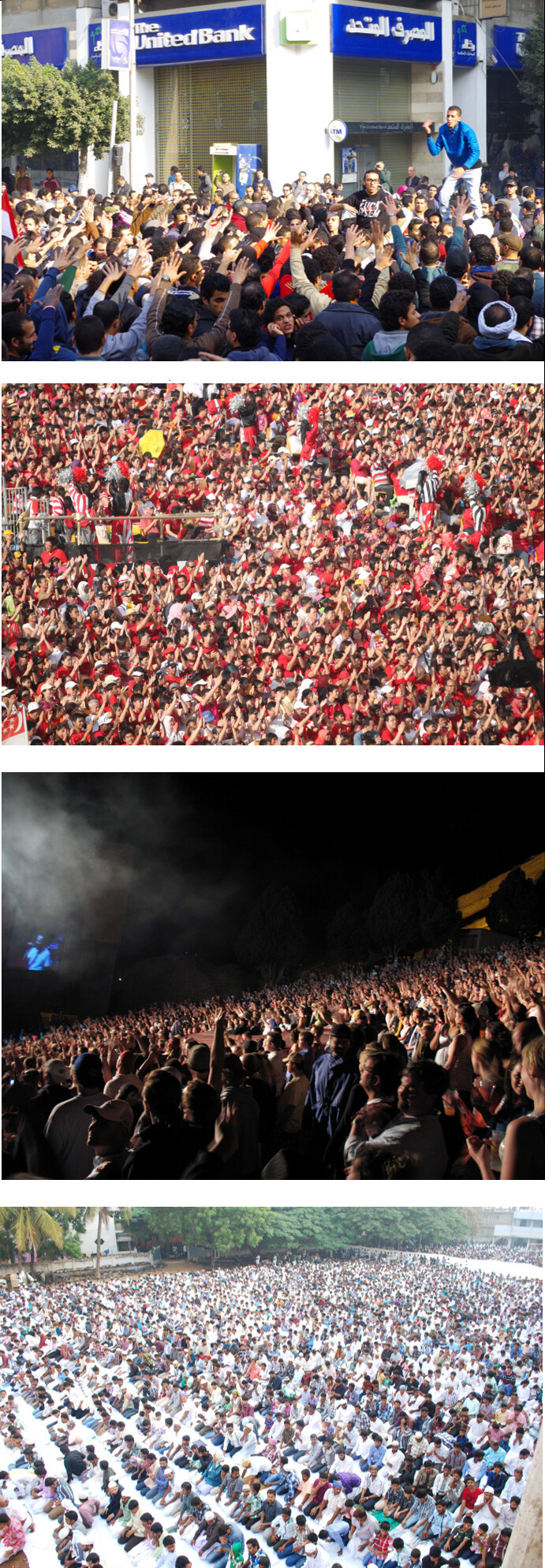}
        \caption{Image}
    \end{subfigure}
    \begin{subfigure}{.24\linewidth}
        \centering
        \includegraphics[width=\linewidth]{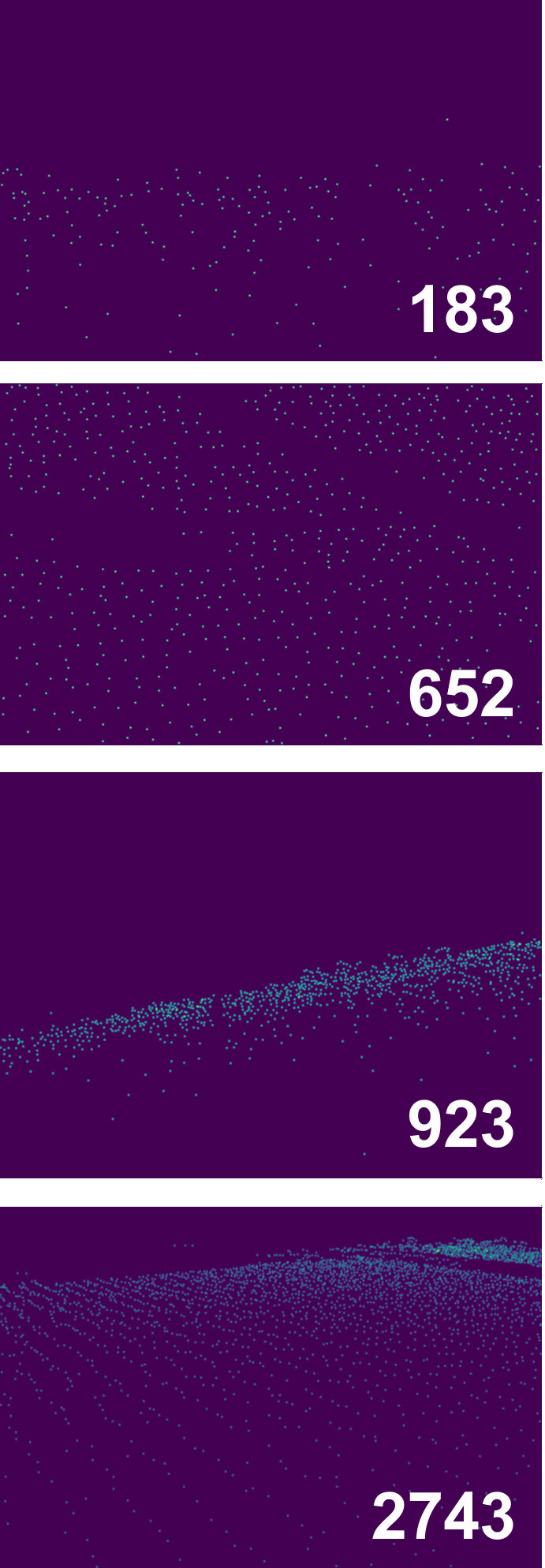}
        \caption{GT}
    \end{subfigure}
    \begin{subfigure}{.24\linewidth}
        \centering
        \includegraphics[width=\linewidth]{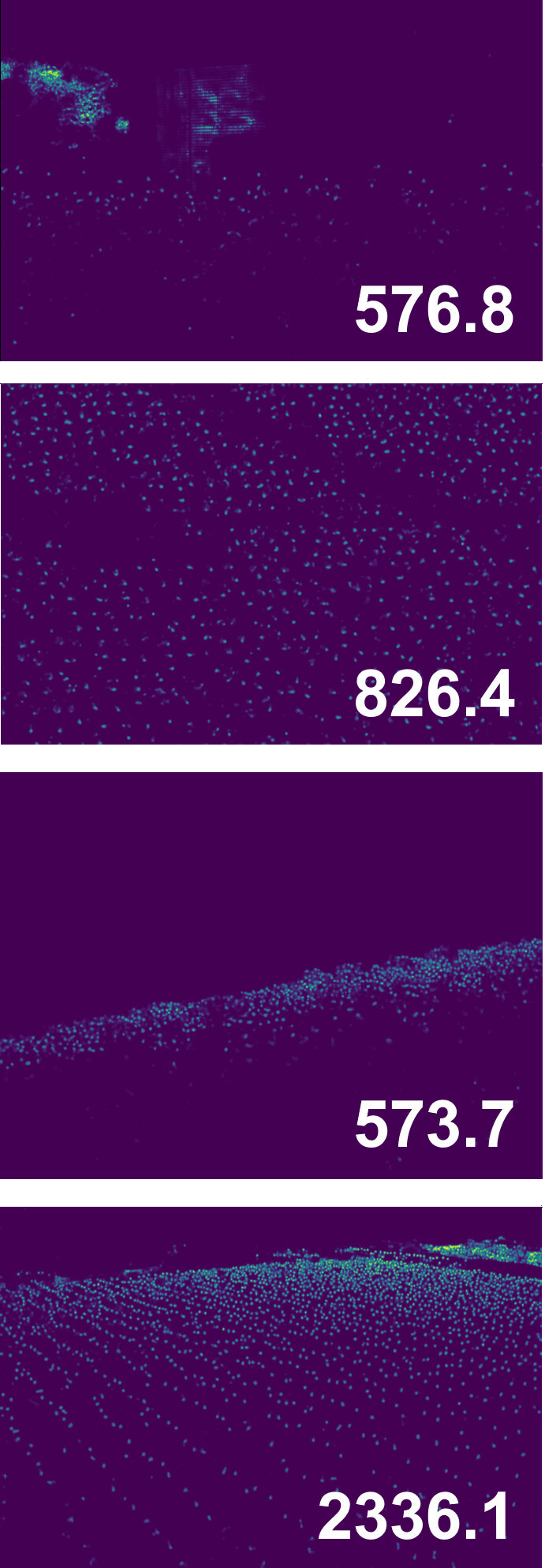}
        \caption{DCCUS}
    \end{subfigure}
    \begin{subfigure}{.24\linewidth}
        \centering
        \includegraphics[width=\linewidth]{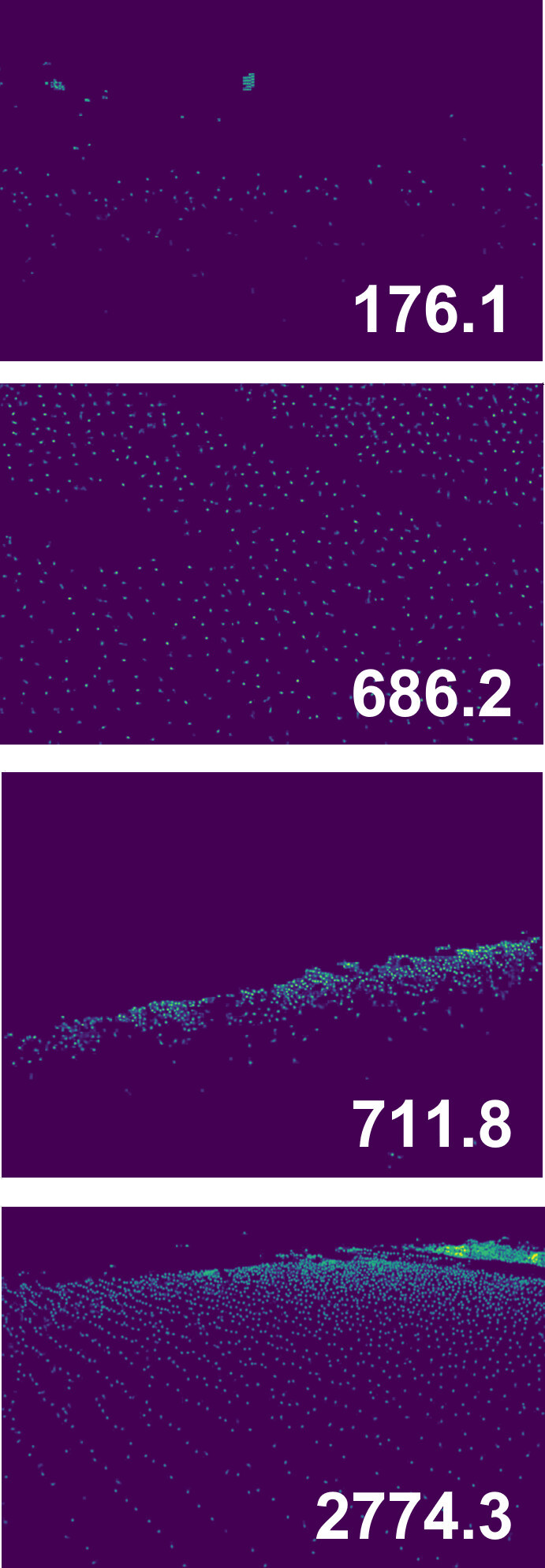}
        \caption{MPCount}
    \end{subfigure}
    \caption{Visualization results on A $\rightarrow$ Q}
    \label{fig:a2q}
\end{figure}

\begin{figure}[h]
    \centering
    \begin{subfigure}{.24\linewidth}
        \centering
        \includegraphics[width=\linewidth]{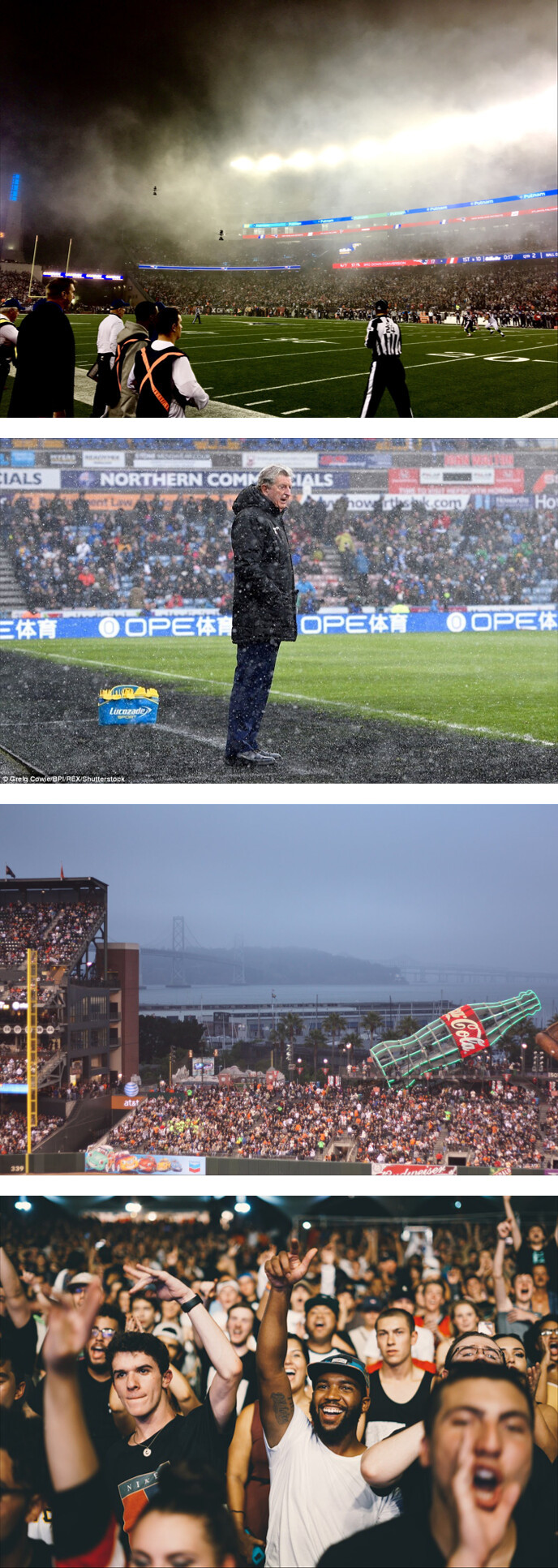}
        \caption{Image}
    \end{subfigure}
    \begin{subfigure}{.24\linewidth}
        \centering
        \includegraphics[width=\linewidth]{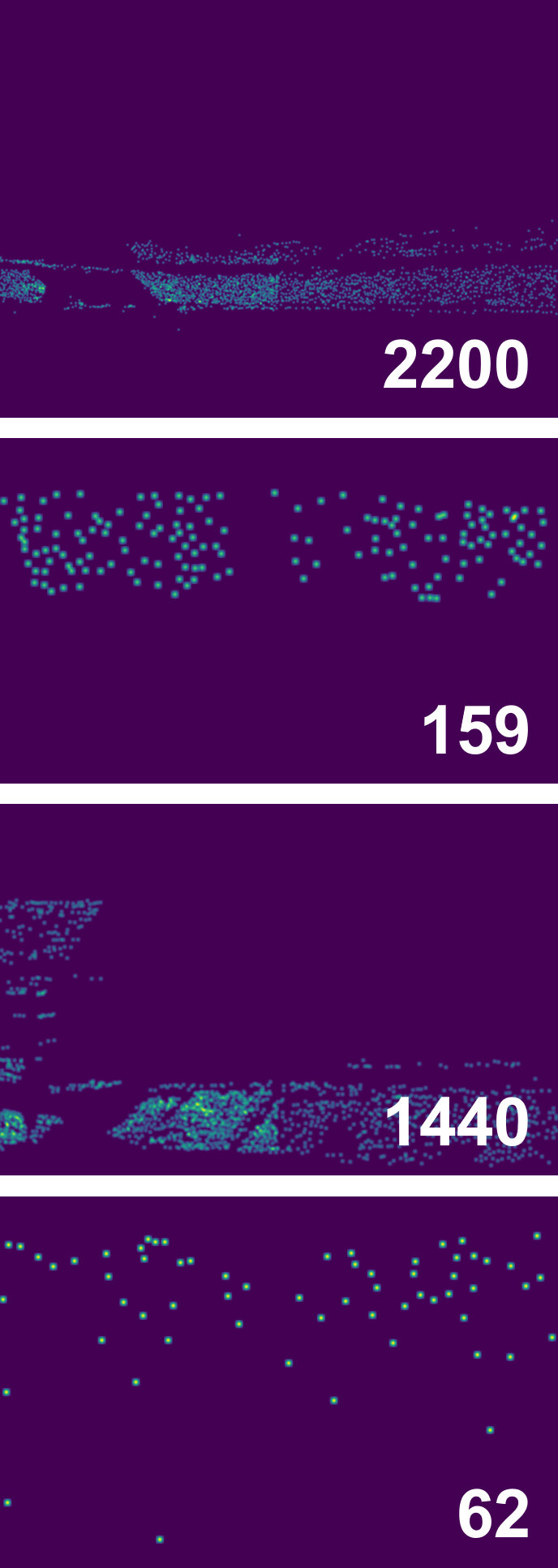}
        \caption{GT}
    \end{subfigure}
    \begin{subfigure}{.24\linewidth}
        \centering
        \includegraphics[width=\linewidth]{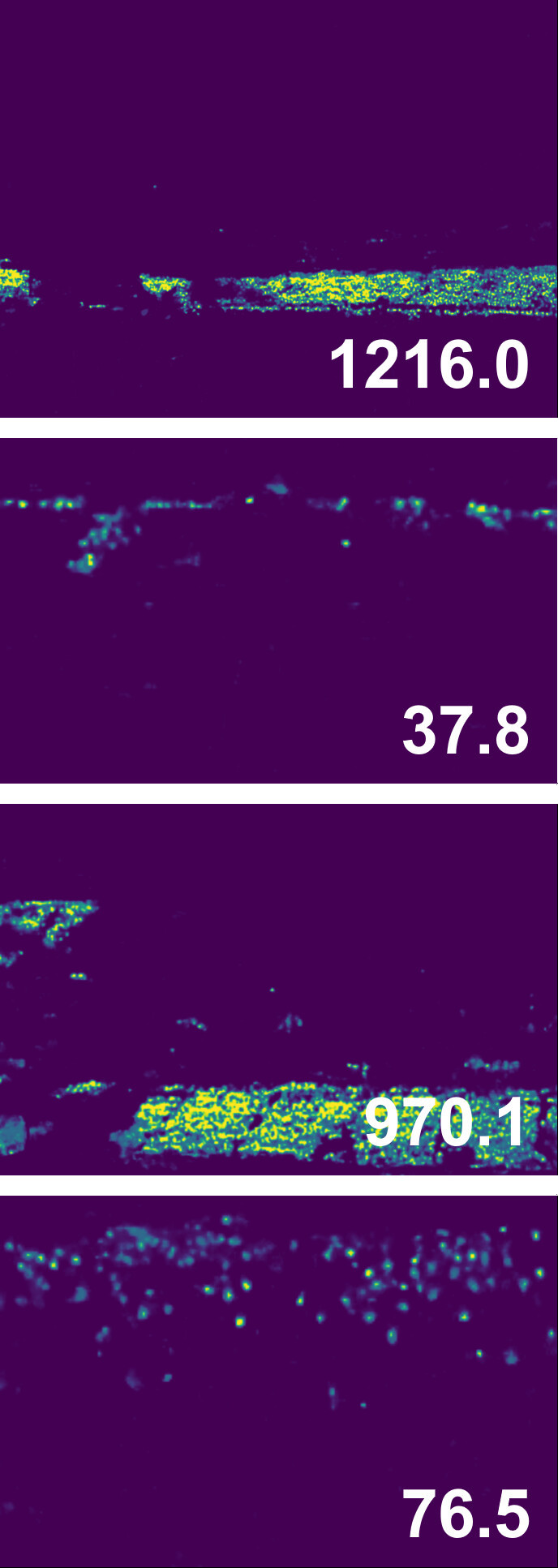}
        \caption{DCCUS}
    \end{subfigure}
    \begin{subfigure}{.24\linewidth}
        \centering
        \includegraphics[width=\linewidth]{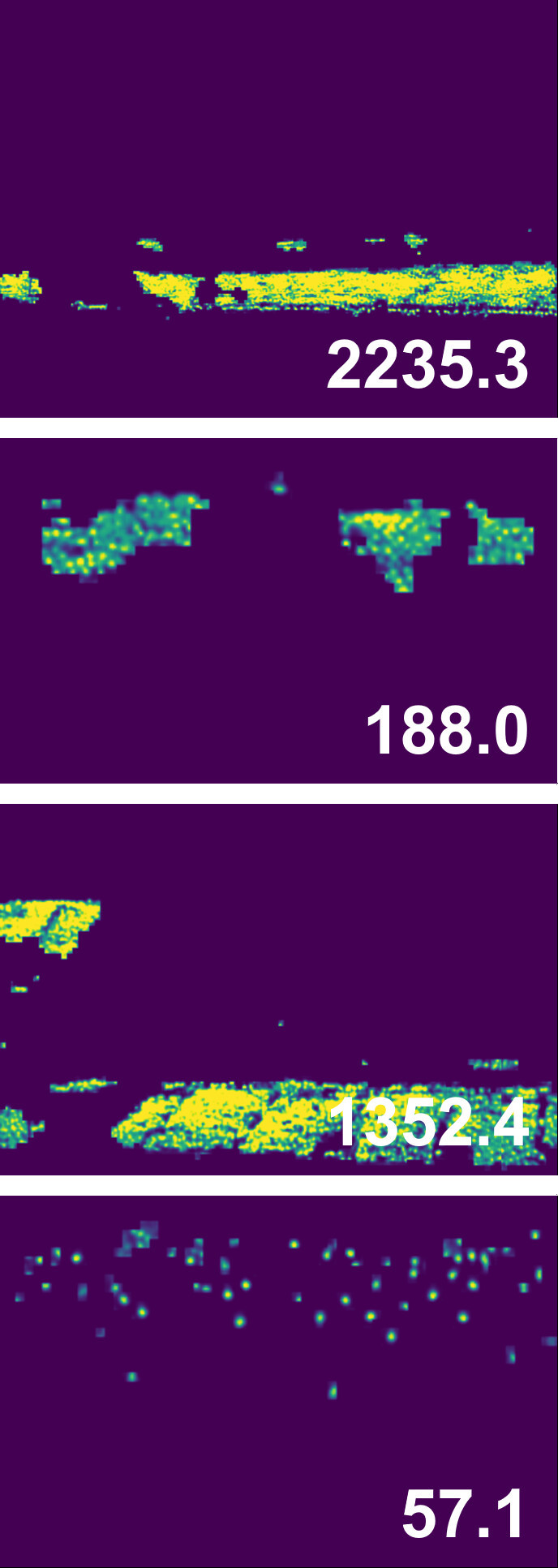}
        \caption{MPCount}
    \end{subfigure}
    \caption{Visualization results on SR $\rightarrow$ SD}
    \label{fig:stadium}
\end{figure}

\begin{figure}[h]
    \centering
    \begin{subfigure}{.24\linewidth}
        \centering
        \includegraphics[width=\linewidth]{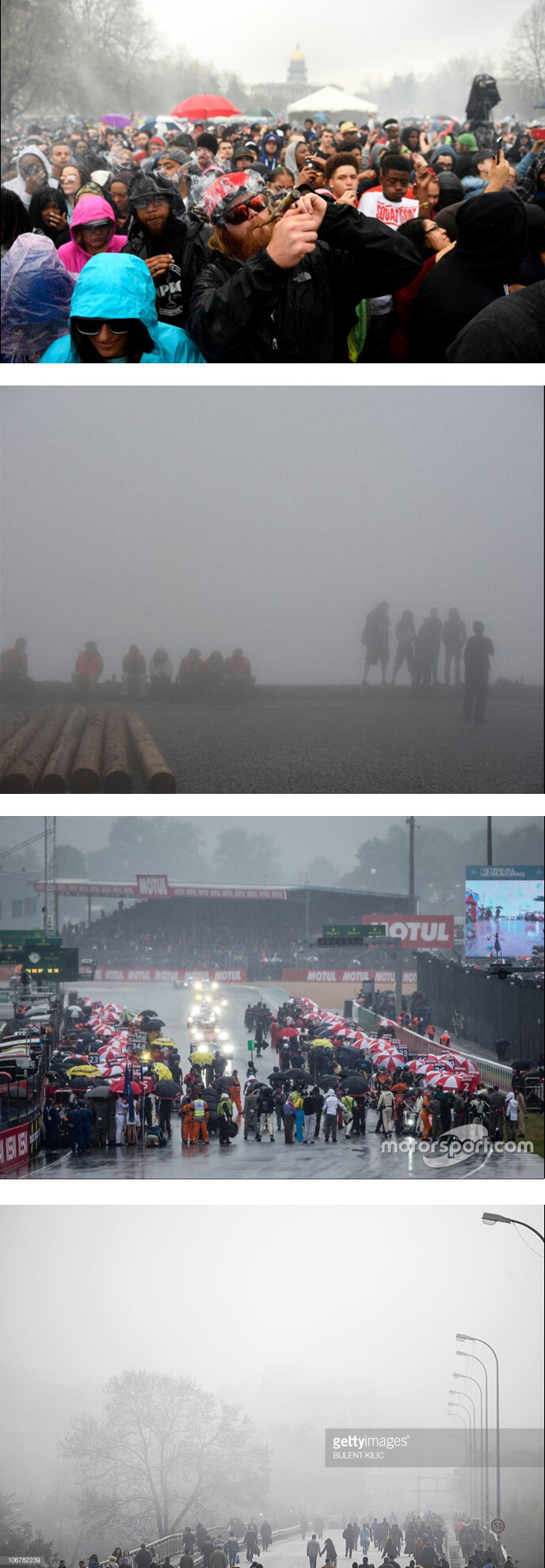}
        \caption{Image}
    \end{subfigure}
    \begin{subfigure}{.24\linewidth}
        \centering
        \includegraphics[width=\linewidth]{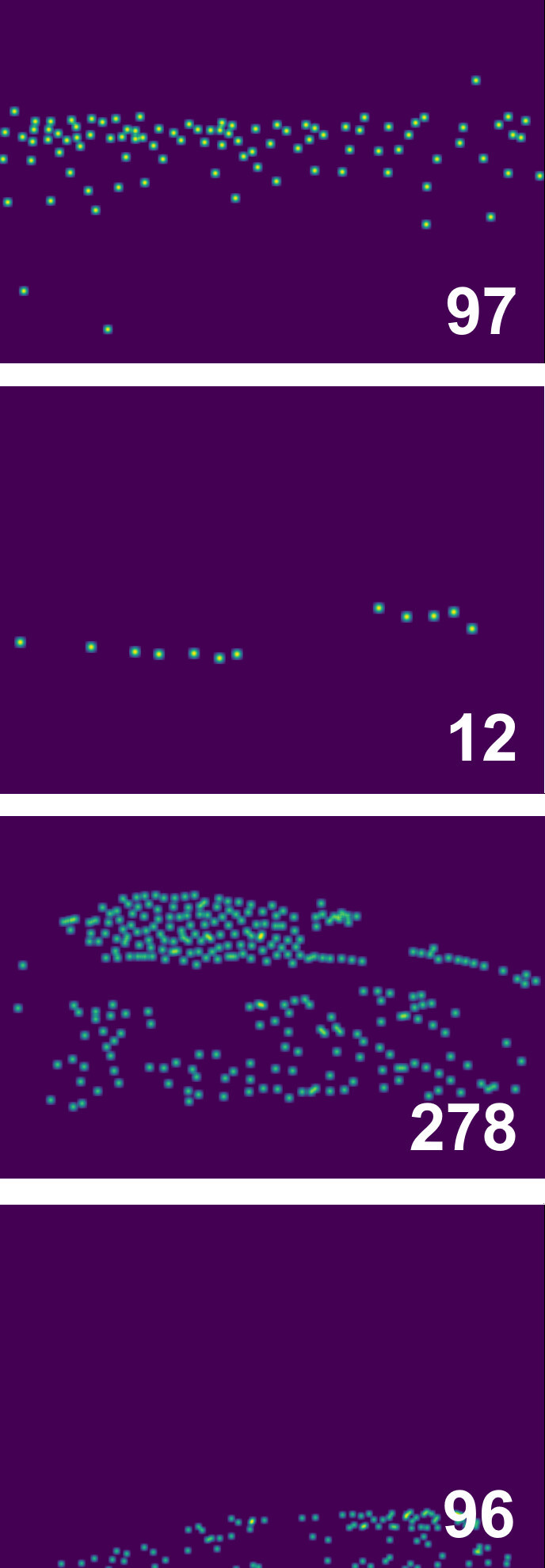}
        \caption{GT}
    \end{subfigure}
    \begin{subfigure}{.24\linewidth}
        \centering
        \includegraphics[width=\linewidth]{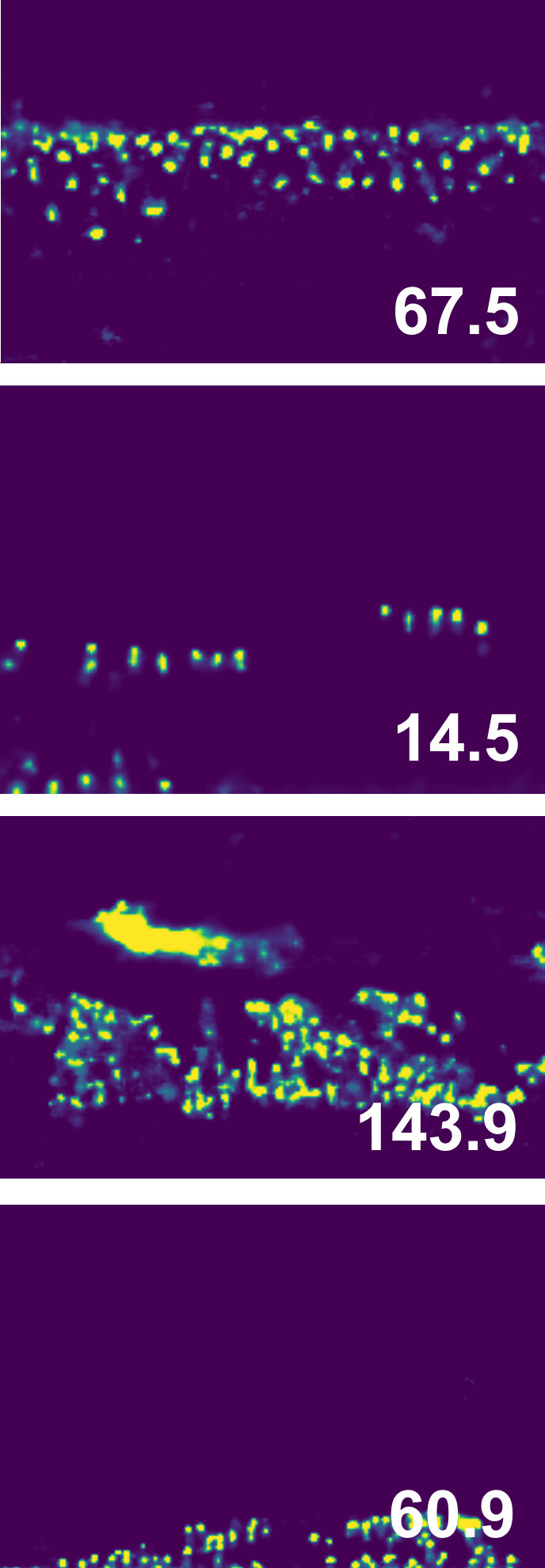}
        \caption{DCCUS}
    \end{subfigure}
    \begin{subfigure}{.24\linewidth}
        \centering
        \includegraphics[width=\linewidth]{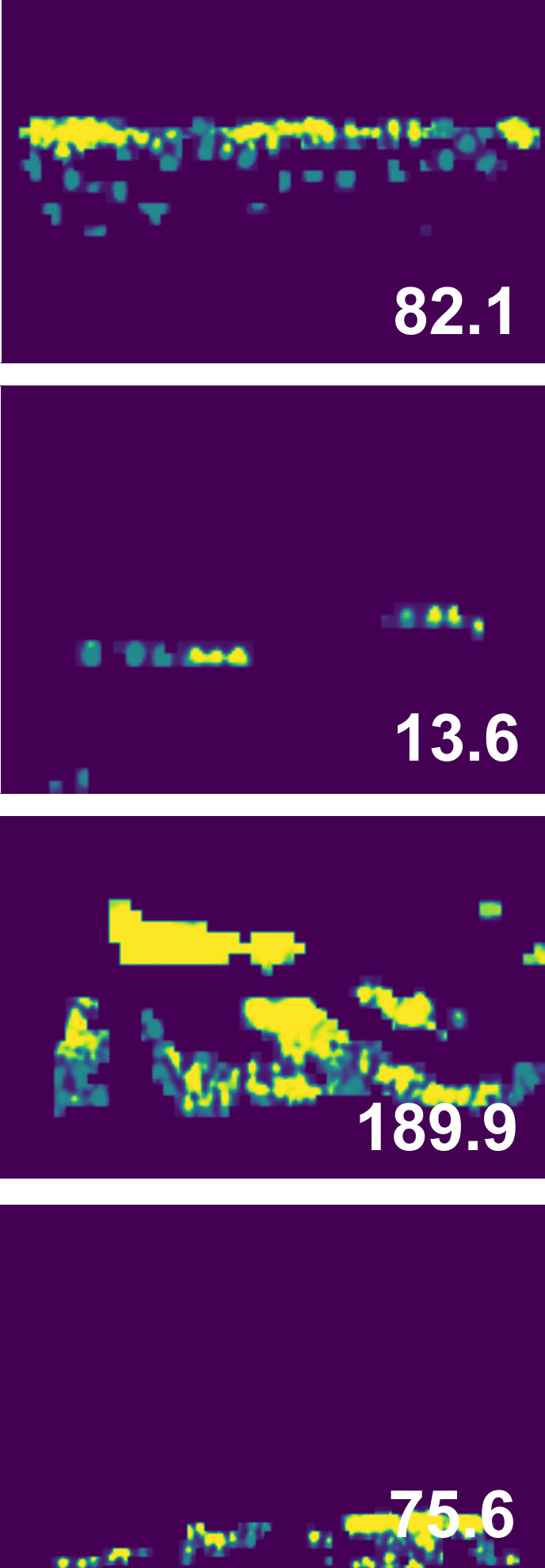}
        \caption{MPCount}
    \end{subfigure}
    \caption{Visualization results on SN $\rightarrow$ FH}
    \label{fig:fog}
\end{figure}

The visualization result on various datasets demonstrate that MPCount predict more accurate density maps compared with DCCUS under different scenarios. The performance gap is even more significant on data with narrow distribution, as shown in~\cref{fig:stadium} and~\cref{fig:fog}. In many cases like the first row in ~\cref{fig:a2q}, DCCUS wrongly outputs high density values from areas containing no human head. MPCount alleviates this problem with PC maps effectively filtering out areas without head based on accurate patch-level information.

\subsection{Visualization of PCMs}
\begin{figure}[h]
    \centering
    \begin{subfigure}{.24\linewidth}
        \centering
        \includegraphics[width=\linewidth]{figs_suppl/pcm/IMG_52_IMG_52.jpg}
    \end{subfigure}
    \begin{subfigure}{.24\linewidth}
        \centering
        \includegraphics[width=\linewidth]{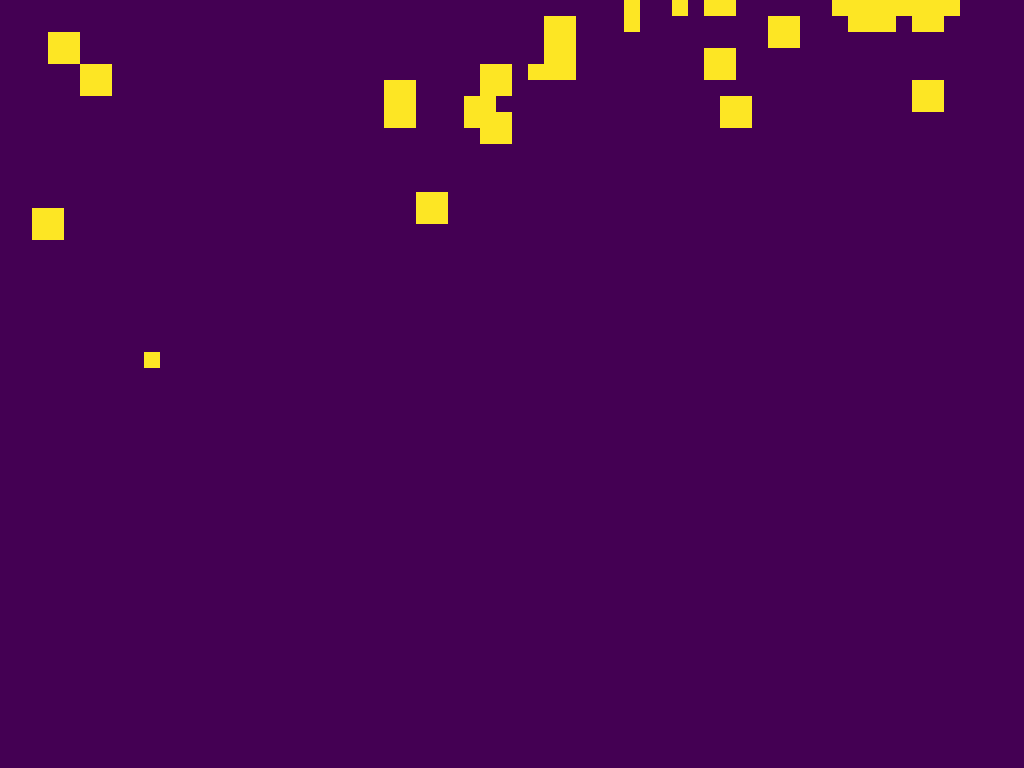}
    \end{subfigure}
    \begin{subfigure}{.24\linewidth}
        \centering
        \includegraphics[width=\linewidth]{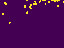}
    \end{subfigure}
    \begin{subfigure}{.24\linewidth}
        \centering
        \includegraphics[width=\linewidth]{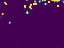}
    \end{subfigure}
    \begin{subfigure}{.24\linewidth}
        \centering
        \includegraphics[width=\linewidth]{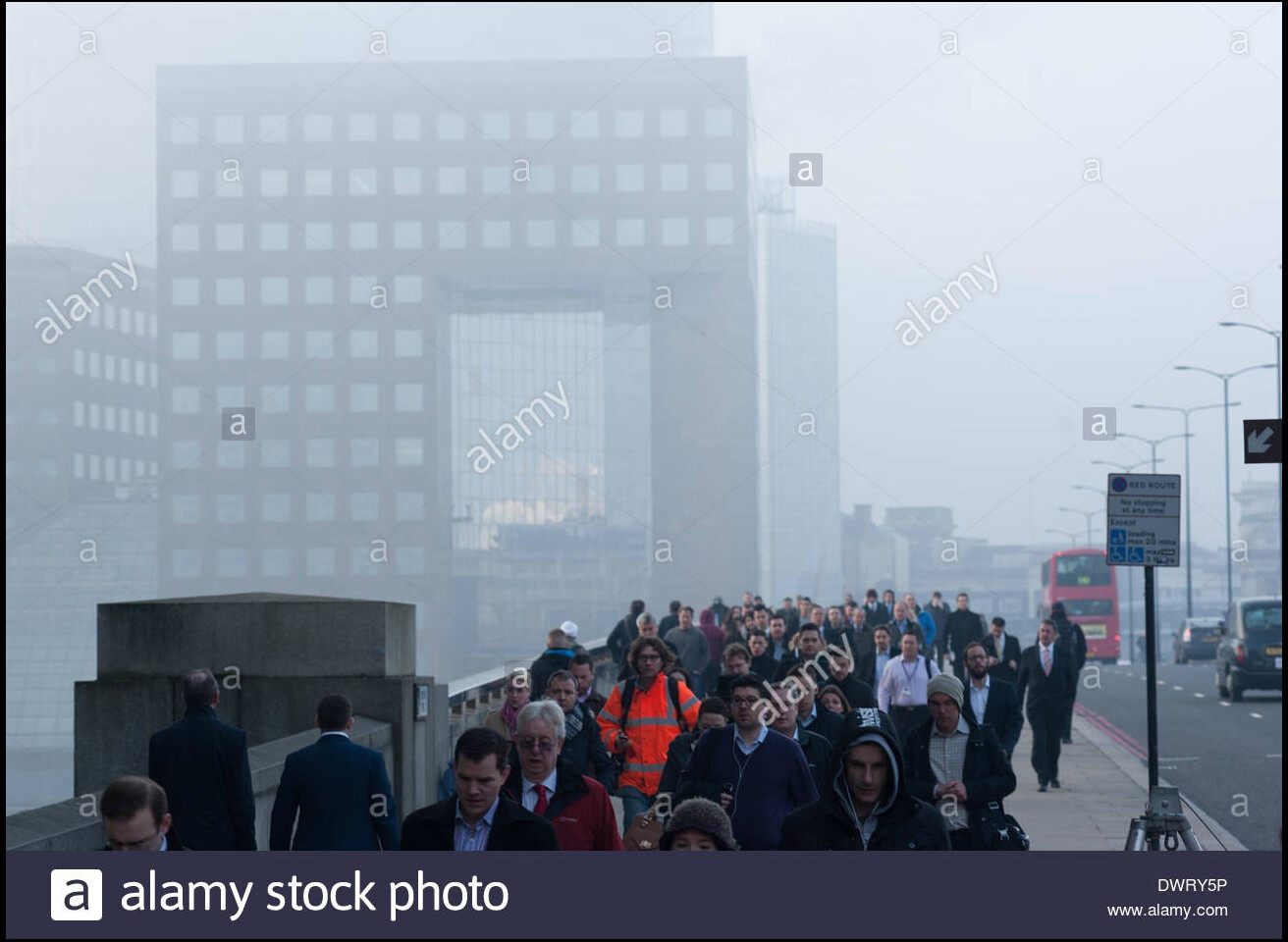}
    \end{subfigure}
    \begin{subfigure}{.24\linewidth}
        \centering
        \includegraphics[width=\linewidth]{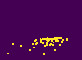}
    \end{subfigure}
    \begin{subfigure}{.24\linewidth}
        \centering
        \includegraphics[width=\linewidth]{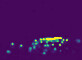}
    \end{subfigure}
    \begin{subfigure}{.24\linewidth}
        \centering
        \includegraphics[width=\linewidth]{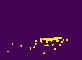}
    \end{subfigure}
    \begin{subfigure}{.24\linewidth}
        \centering
        \includegraphics[width=\linewidth]{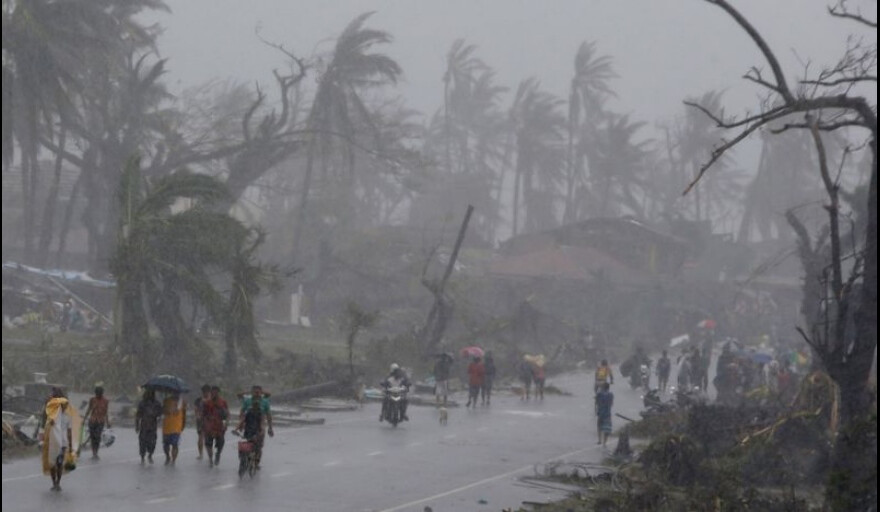}
    \end{subfigure}
    \begin{subfigure}{.24\linewidth}
        \centering
        \includegraphics[width=\linewidth]{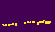}
    \end{subfigure}
    \begin{subfigure}{.24\linewidth}
        \centering
        \includegraphics[width=\linewidth]{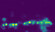}
    \end{subfigure}
    \begin{subfigure}{.24\linewidth}
        \centering
        \includegraphics[width=\linewidth]{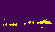}
    \end{subfigure}
    \begin{subfigure}{.24\linewidth}
        \centering
        \includegraphics[width=\linewidth]{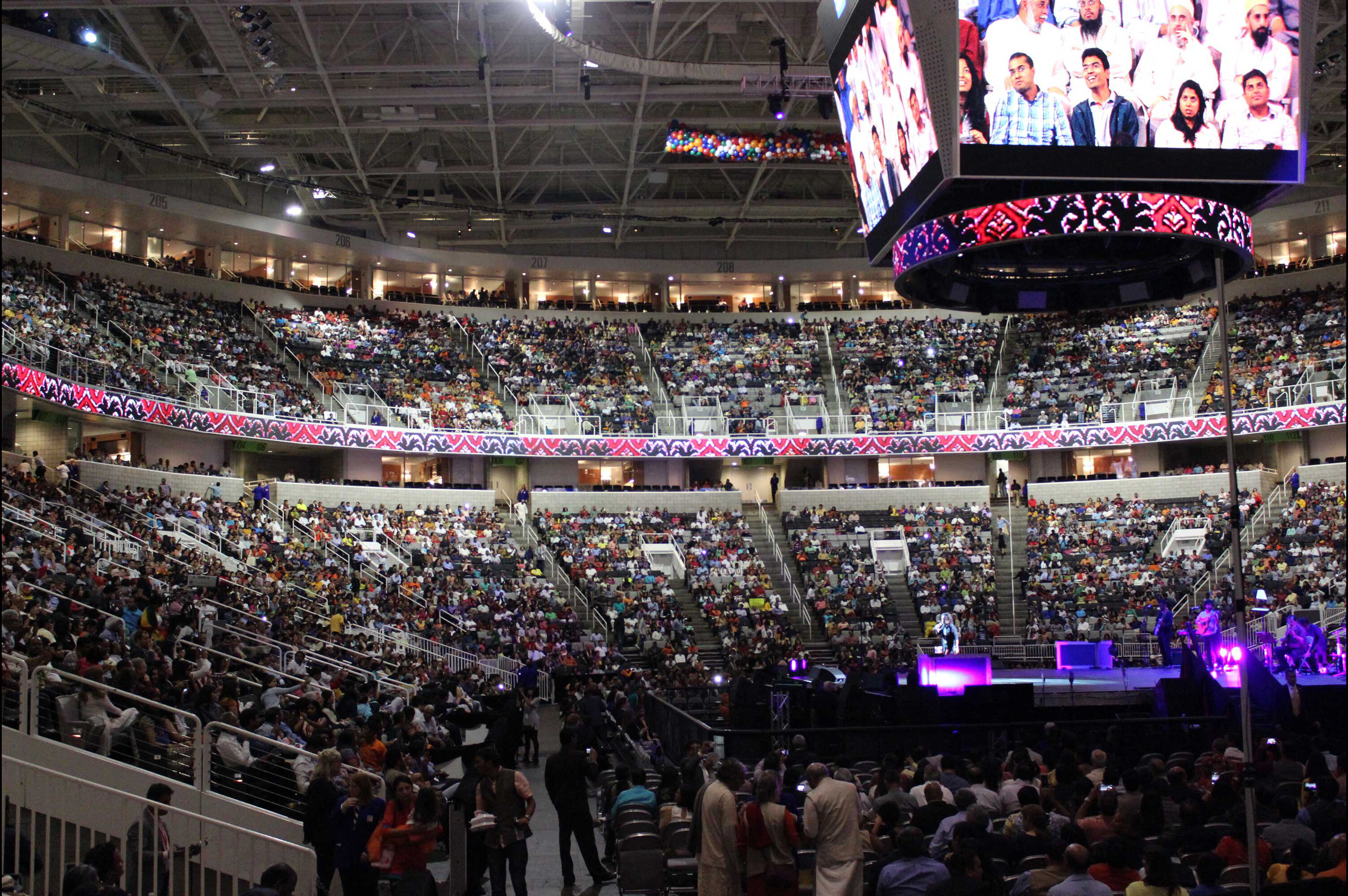}
    \end{subfigure}
    \begin{subfigure}{.24\linewidth}
        \centering
        \includegraphics[width=\linewidth]{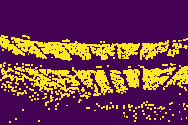}
    \end{subfigure}
    \begin{subfigure}{.24\linewidth}
        \centering
        \includegraphics[width=\linewidth]{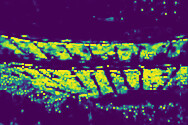}
    \end{subfigure}
    \begin{subfigure}{.24\linewidth}
        \centering
        \includegraphics[width=\linewidth]{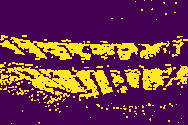}
    \end{subfigure}
    \begin{subfigure}{.24\linewidth}
        \centering
        \includegraphics[width=\linewidth]{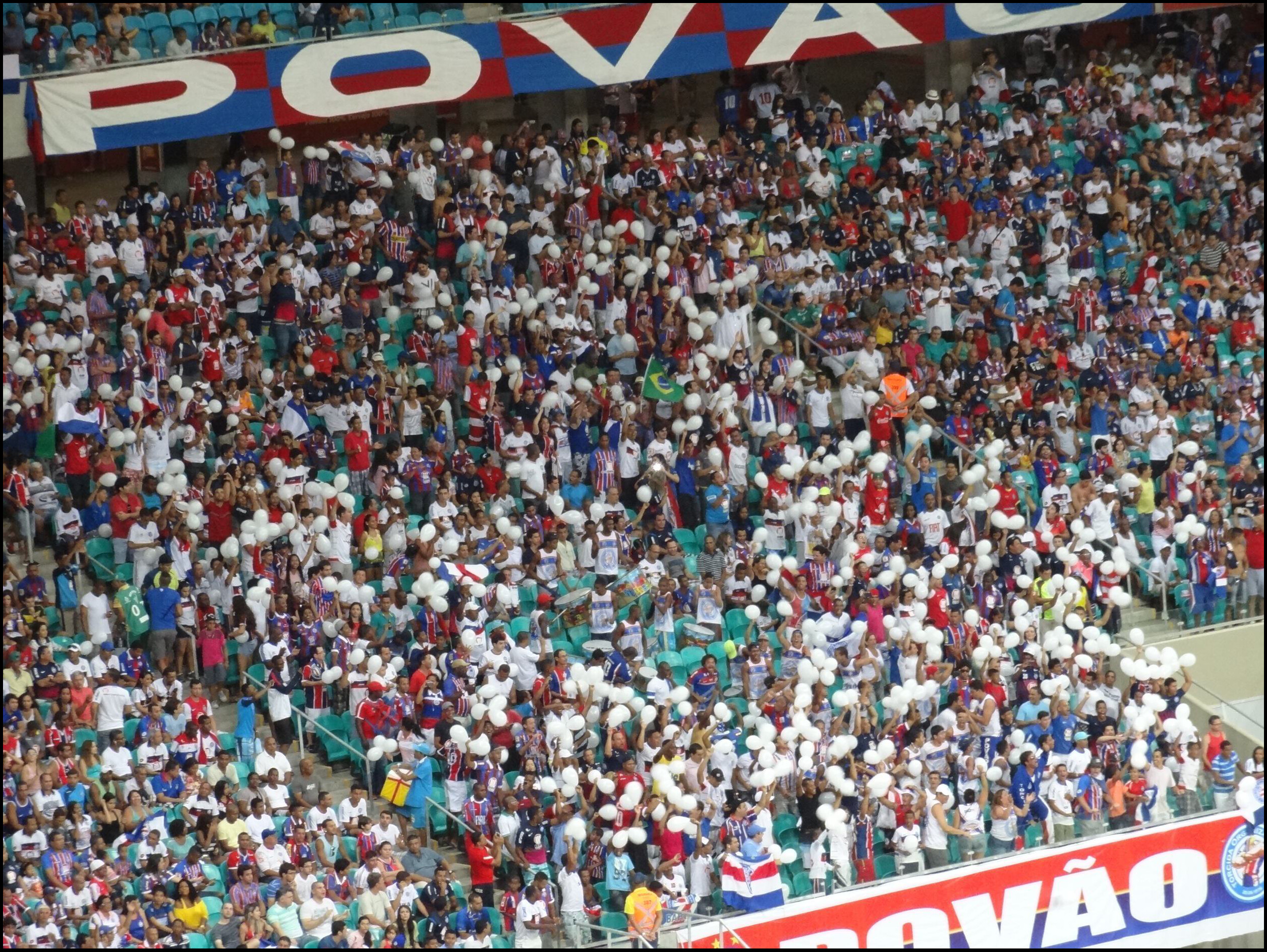}
    \end{subfigure}
    \begin{subfigure}{.24\linewidth}
        \centering
        \includegraphics[width=\linewidth]{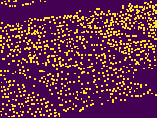}
    \end{subfigure}
    \begin{subfigure}{.24\linewidth}
        \centering
        \includegraphics[width=\linewidth]{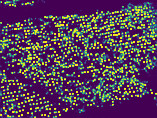}
    \end{subfigure}
    \begin{subfigure}{.24\linewidth}
        \centering
        \includegraphics[width=\linewidth]{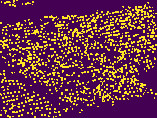}
    \end{subfigure}
    \begin{subfigure}{.24\linewidth}
        \centering
        \includegraphics[width=\linewidth]{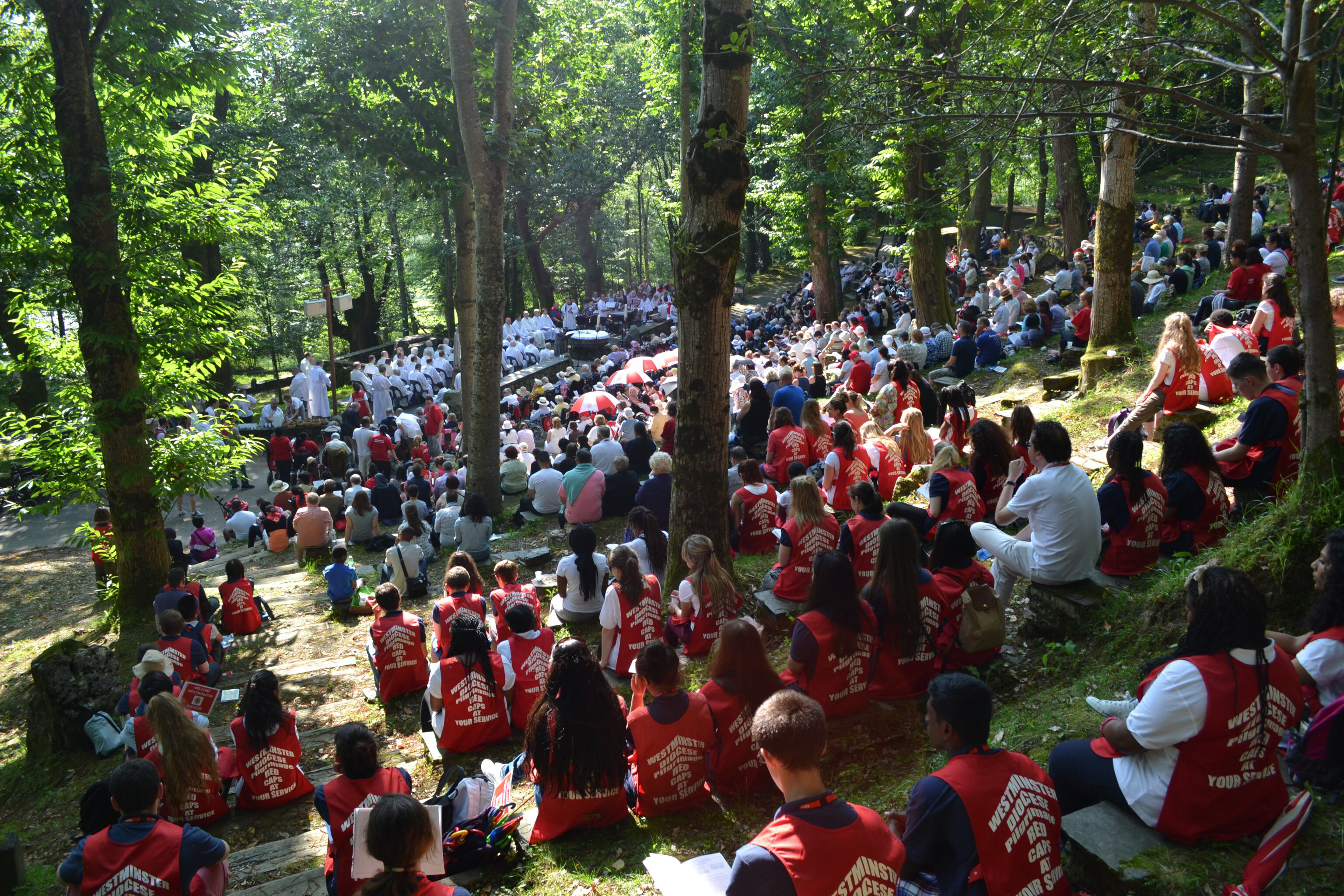}
    \end{subfigure}
    \begin{subfigure}{.24\linewidth}
        \centering
        \includegraphics[width=\linewidth]{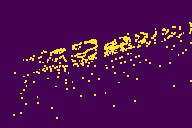}
    \end{subfigure}
    \begin{subfigure}{.24\linewidth}
        \centering
        \includegraphics[width=\linewidth]{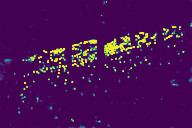}
    \end{subfigure}
    \begin{subfigure}{.24\linewidth}
        \centering
        \includegraphics[width=\linewidth]{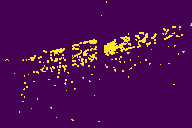}
    \end{subfigure}
    \begin{subfigure}{.24\linewidth}
        \centering
        \includegraphics[width=\linewidth]{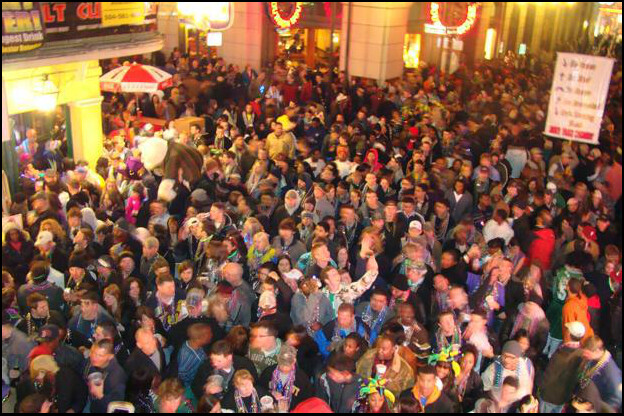}
    \end{subfigure}
    \begin{subfigure}{.24\linewidth}
        \centering
        \includegraphics[width=\linewidth]{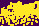}
    \end{subfigure}
    \begin{subfigure}{.24\linewidth}
        \centering
        \includegraphics[width=\linewidth]{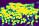}
    \end{subfigure}
    \begin{subfigure}{.24\linewidth}
        \centering
        \includegraphics[width=\linewidth]{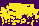}
    \end{subfigure}
    \begin{subfigure}{.24\linewidth}
        \centering
        \includegraphics[width=\linewidth]{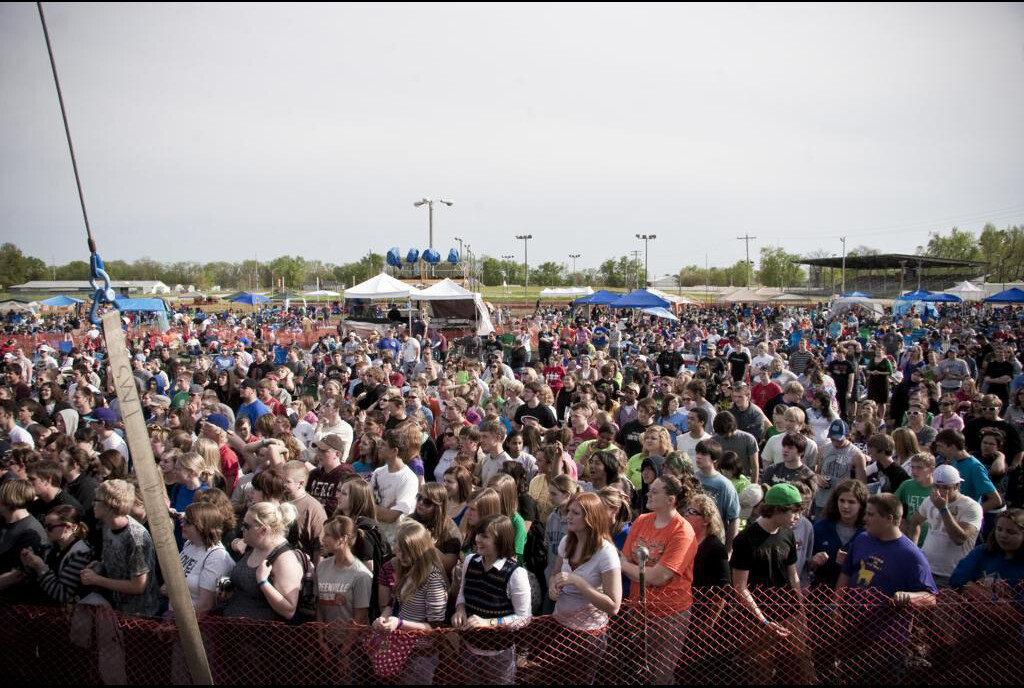}
        \caption{Image}
    \end{subfigure}
    \begin{subfigure}{.24\linewidth}
        \centering
        \includegraphics[width=\linewidth]{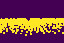}
        \caption{GT}
    \end{subfigure}
    \begin{subfigure}{.24\linewidth}
        \centering
        \includegraphics[width=\linewidth]{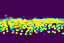}
        \caption{Predicted}
    \end{subfigure}
    \begin{subfigure}{.24\linewidth}
        \centering
        \includegraphics[width=\linewidth]{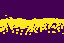}
        \caption{Binarized}
    \end{subfigure}
    \caption{Visualization results of PCMs under various settings}
    \label{fig:pcm}
\end{figure}

\cref{fig:pcm} displays the ground-truth PCMs and the predicted PCMs before and after binarization from different datasets. The results indicate that, the predicted PCMs can accurately capture the information of crowd location, and noisy predictions are successfully filtered out by binarization. 

\begin{figure}[h]
    \centering
    \begin{subfigure}{.24\linewidth}
        \centering
        \includegraphics[width=\linewidth]{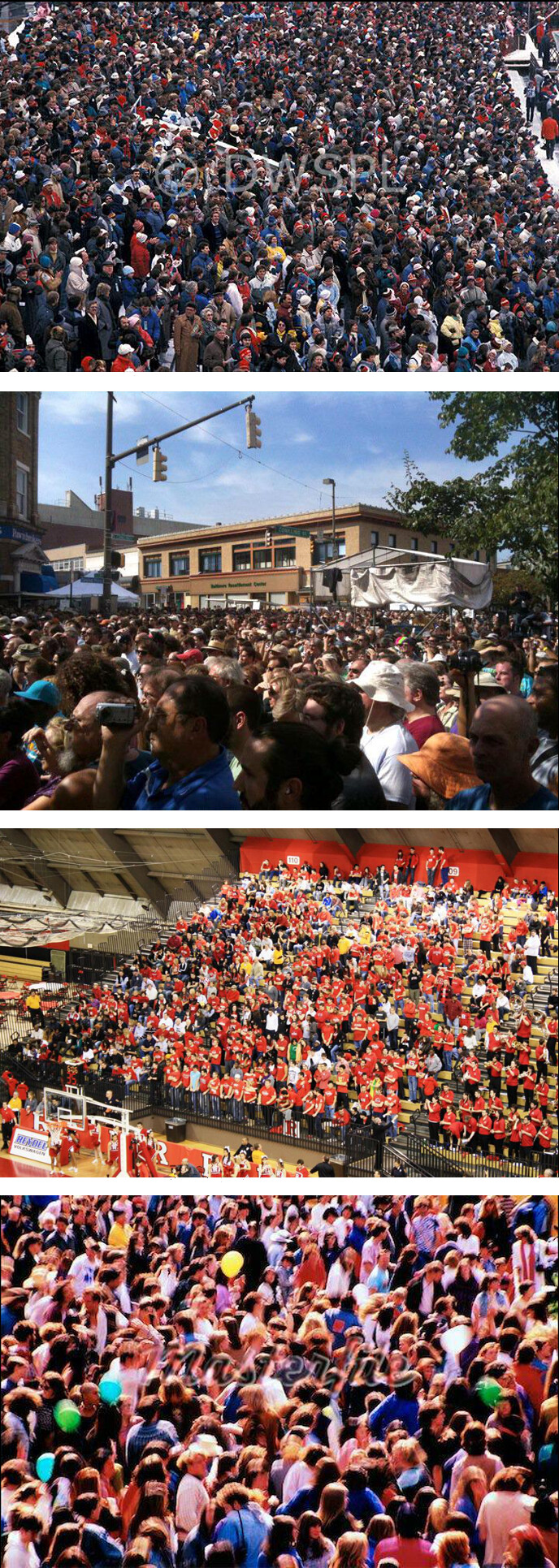}
        \caption{Image}
    \end{subfigure}
    \begin{subfigure}{.24\linewidth}
        \centering
        \includegraphics[width=\linewidth]{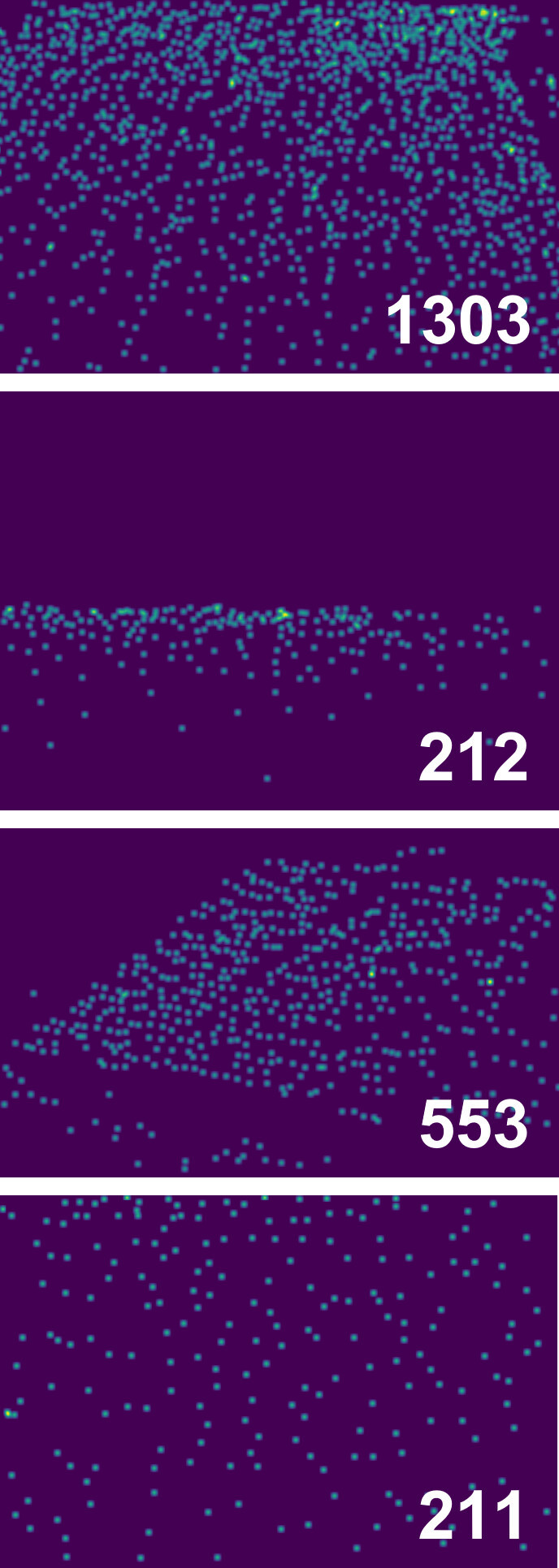}
        \caption{GT}
    \end{subfigure}
    \begin{subfigure}{.24\linewidth}
        \centering
        \includegraphics[width=\linewidth]{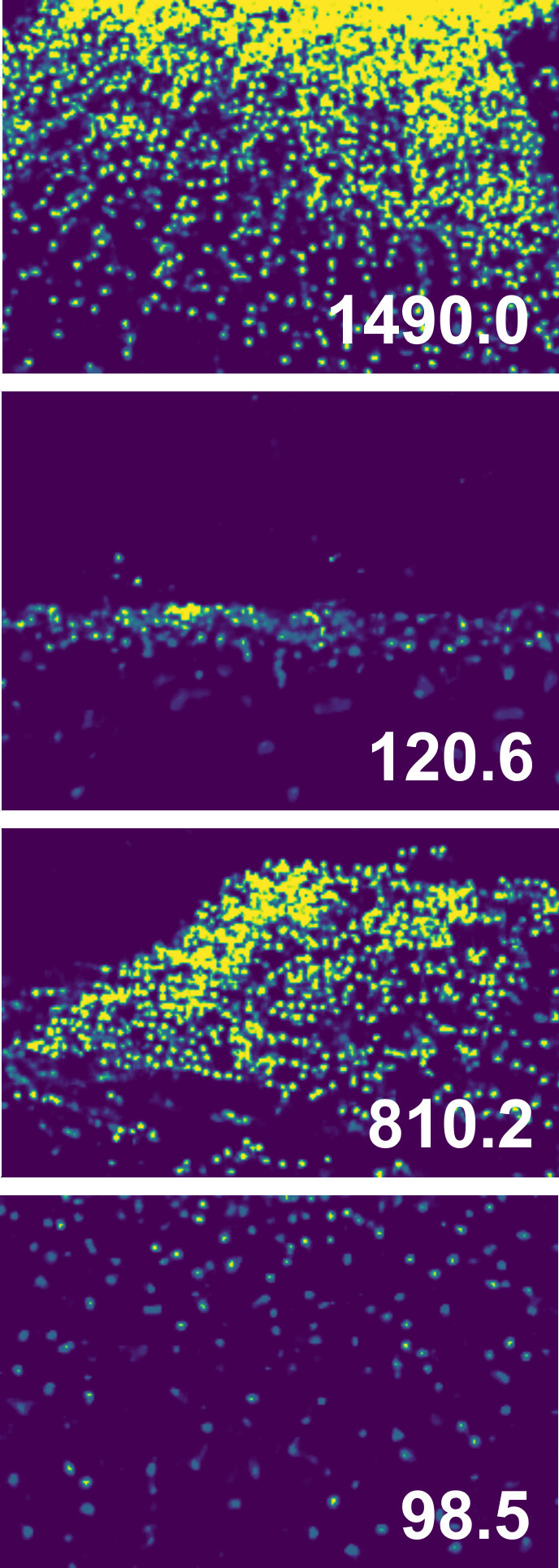}
        \caption{DCCUS}
    \end{subfigure}
    \begin{subfigure}{.24\linewidth}
        \centering
        \includegraphics[width=\linewidth]{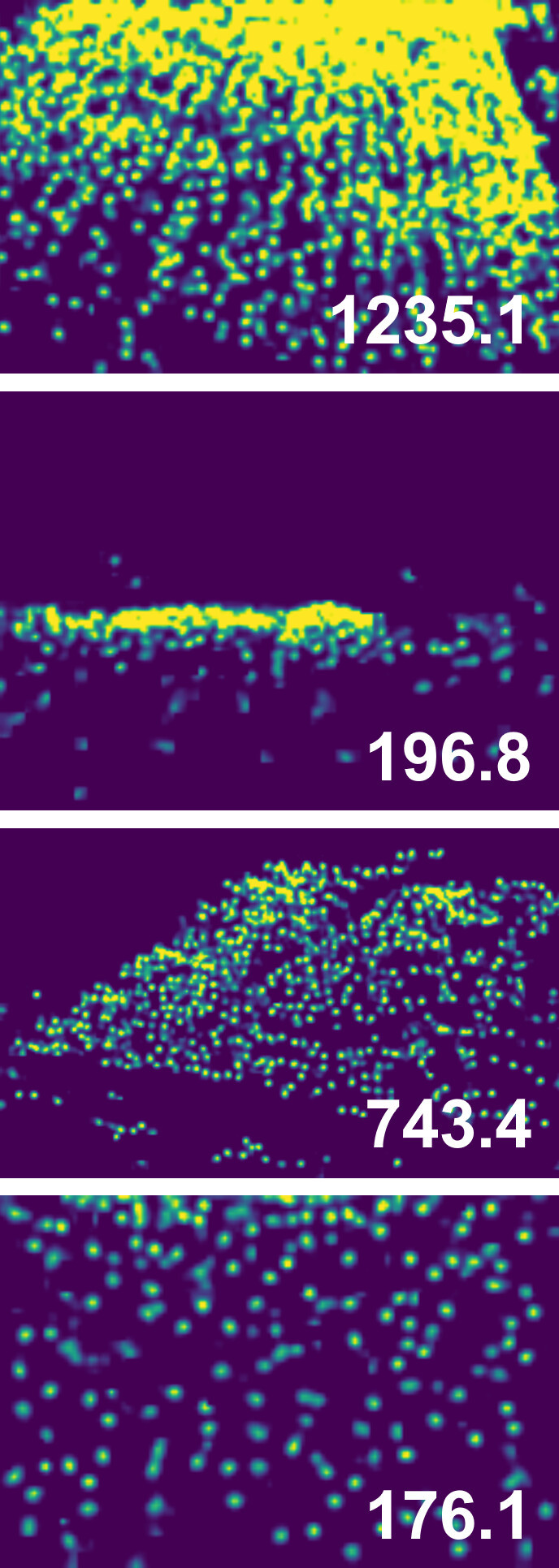}
        \caption{MPCount}
    \end{subfigure}
    \caption{Visualization results on B $\rightarrow$ A}
    \label{fig:b2a}
\end{figure}

\begin{figure}[h]
    \centering
    \begin{subfigure}{.24\linewidth}
        \centering
        \includegraphics[width=\linewidth]{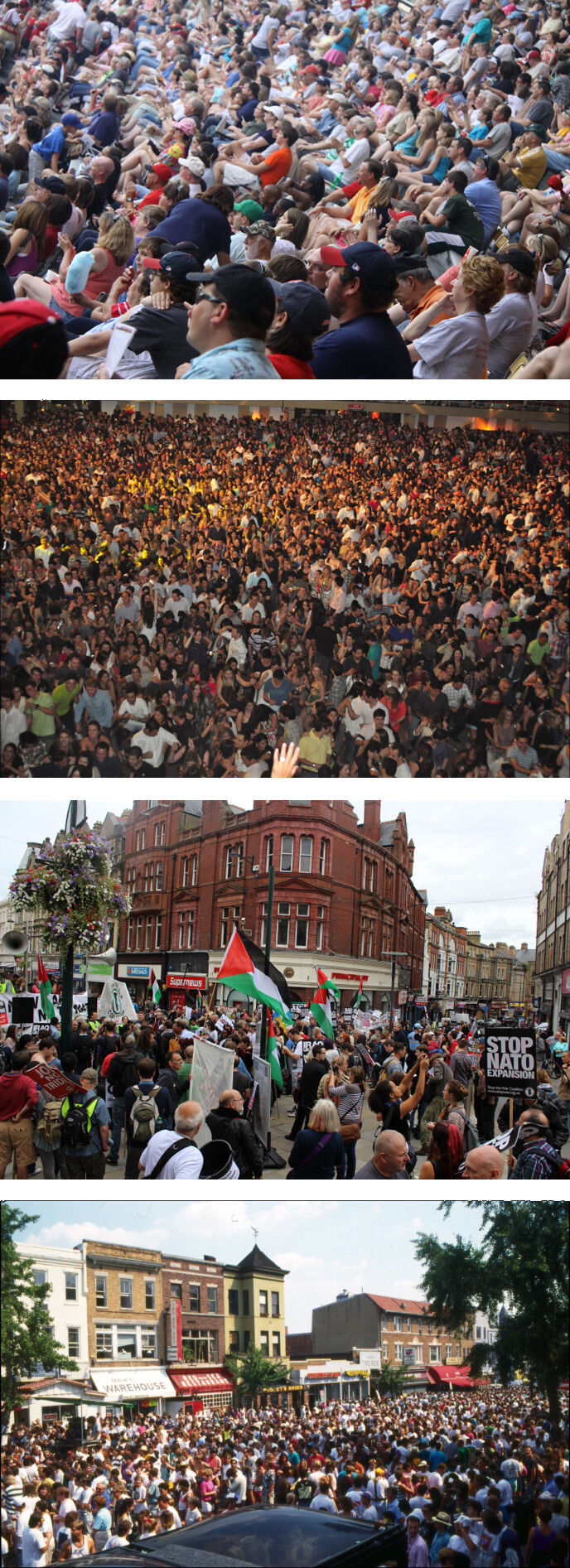}
        \caption{Image}
    \end{subfigure}
    \begin{subfigure}{.24\linewidth}
        \centering
        \includegraphics[width=\linewidth]{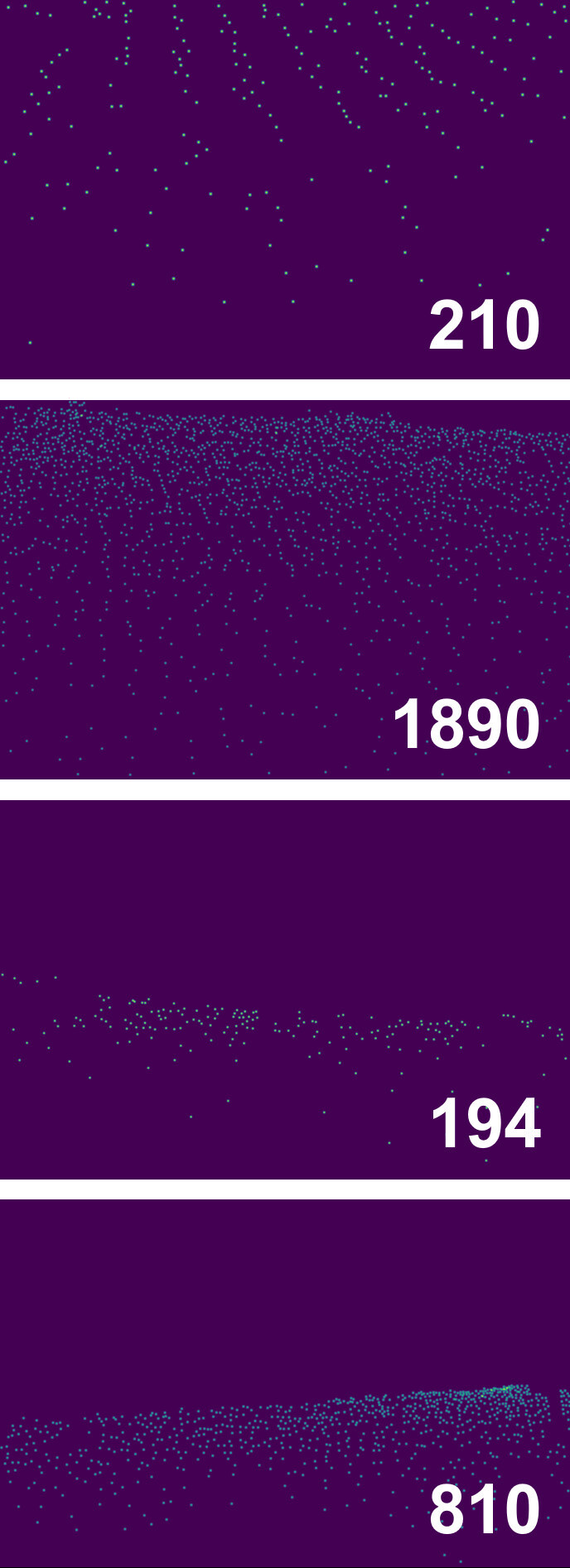}
        \caption{GT}
    \end{subfigure}
    \begin{subfigure}{.24\linewidth}
        \centering
        \includegraphics[width=\linewidth]{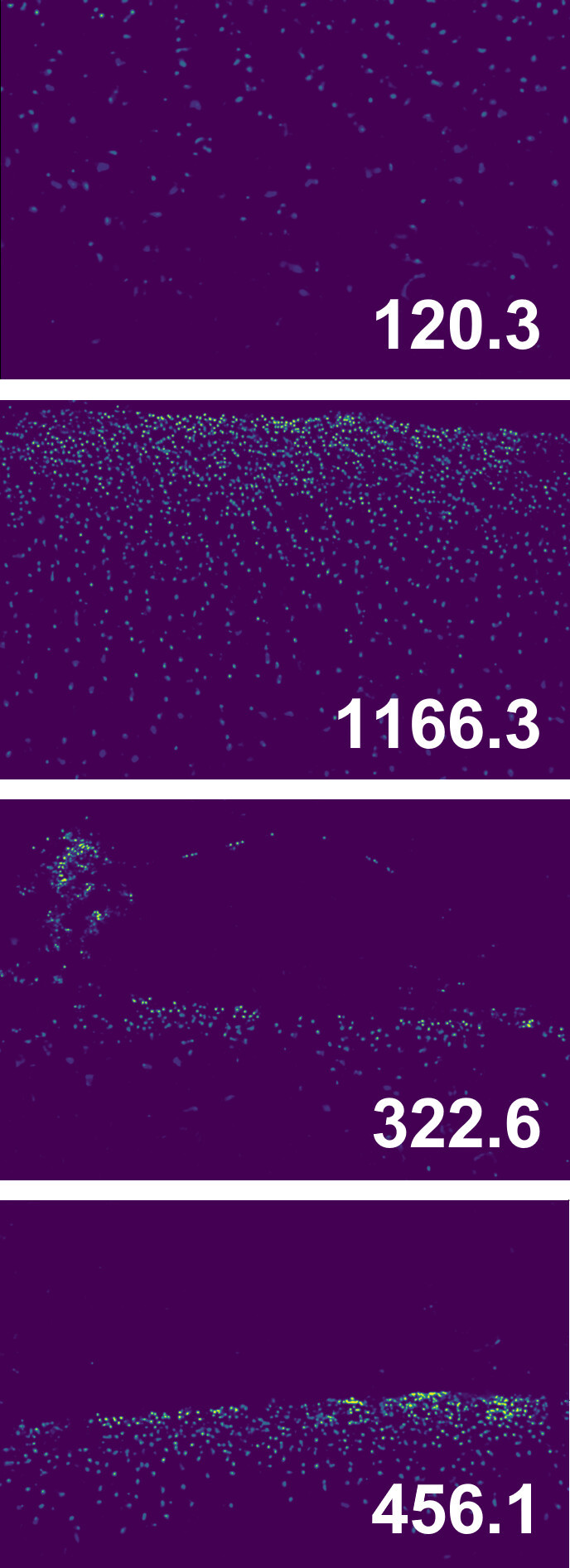}
        \caption{DCCUS}
    \end{subfigure}
    \begin{subfigure}{.24\linewidth}
        \centering
        \includegraphics[width=\linewidth]{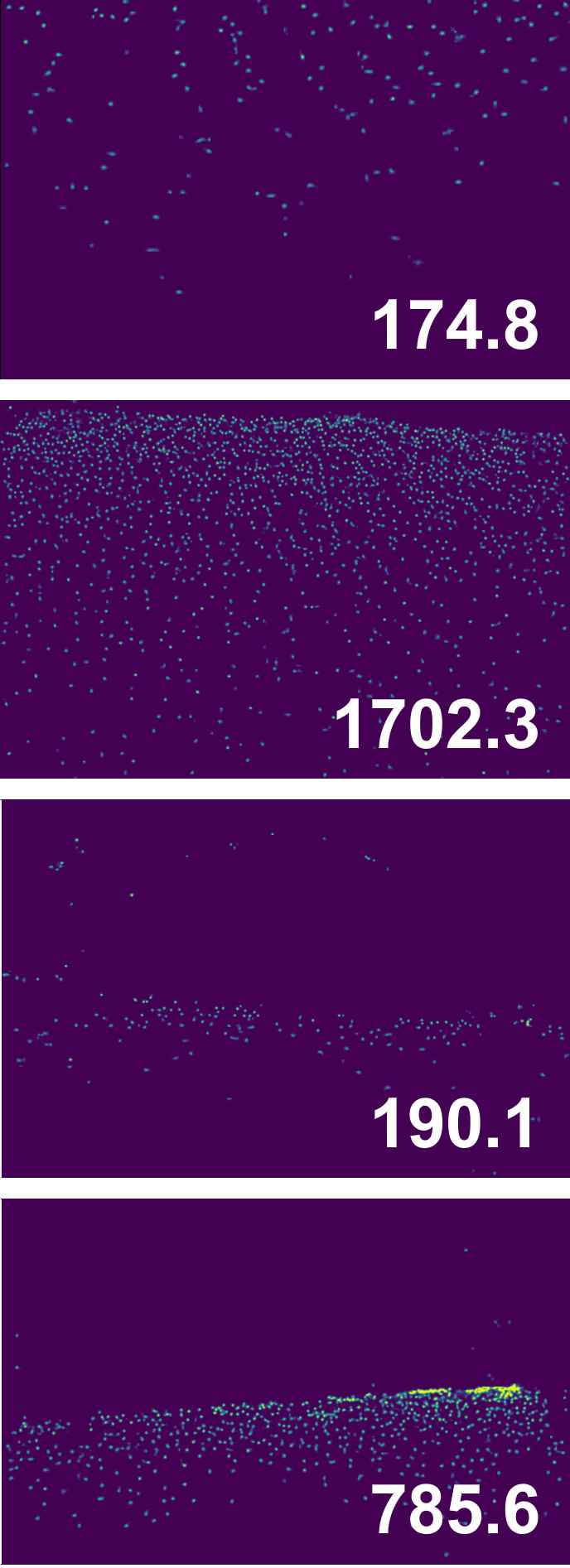}
        \caption{MPCount}
    \end{subfigure}
    \caption{Visualization results on B $\rightarrow$ Q}
    \label{fig:b2q}
\end{figure}

